\documentclass[11pt]{article}

\usepackage[final]{acl}
\usepackage{times}
\usepackage{latexsym}

\usepackage[T1]{fontenc}

\usepackage[utf8]{inputenc}

\usepackage{microtype}

\usepackage{inconsolata}

\usepackage{graphicx}
\usepackage{tikz}
\usepackage[ruled,vlined]{algorithm2e}
\usepackage{amsmath}
\usepackage{amssymb}
\usepackage{hyperref}       
\usepackage{url}            
\usepackage{booktabs}       
\usepackage{siunitx}        
\usepackage{multirow}       
\usepackage{caption}        
\usepackage{amsfonts}       
\usepackage{nicefrac}       
\usepackage[dvipsnames,table]{xcolor}   
\usepackage{adjustbox}

\usepackage{tcolorbox}
\tcbuselibrary{breakable}
\usepackage{capt-of} 
\usepackage{subcaption}
\tcbuselibrary{listings,breakable}
\newtcolorbox[auto counter, number within=section]{promptbox}[2][]{
    breakable,
    colback=gray!5,
    colframe=black!75,
    fonttitle=\bfseries,
    title=Prompt~\thetcbcounter: #2,
    sharp corners,
    boxrule=0.8pt,
    listing only,
    #1
}

\usepackage{tikz}
\usetikzlibrary{positioning, arrows.meta, shapes.geometric}

\title{LLM Explainability with Counterfactual Chains and Causal Graphs}

\author{
  \textbf{Nirit Nussbaum-Hoffer\textsuperscript{T}} \quad
  \textbf{Nitay Calderon\textsuperscript{T}} \quad
  \textbf{Liat Ein-Dor\textsuperscript{I}} \quad
  \textbf{Roi Reichart\textsuperscript{T}}
\\
\\
  \textsuperscript{T}Faculty of Data and Decision Sciences, Technion \quad
  \textsuperscript{I}IBM Research
\\
  \texttt{snnuss@campus.technion.ac.il} \quad \texttt{nitay@campus.technion.ac.il} \\
  \texttt{liate@il.ibm.com} \quad \texttt{roiri@technion.ac.il}
}

\begin{document}
\maketitle
\begin{abstract}
Causal graphs provide a high-level language for making mechanisms transparent. Recent work uses Large Language Models (LLMs) to recover causal graphs of external-world processes. Instead, in this paper, we use causal graphs to model LLM inference itself, providing stakeholders with a transparent view of how the model perceives and organizes high-level concepts to produce a prediction. We propose a four-phase method for constructing such graphs. Given a target LLM and a set of textual examples, our method discovers class-discriminative, human-interpretable concepts and maps each input to LLM-perceived concept states. We then introduce an MCMC-inspired counterfactual augmentation procedure that expands the sparse observational data through chains of counterfactuals. This enables stable causal discovery with $\sigma$-CG, yielding informative, interpretable graphs. 
We apply our method to three LLMs across disease diagnosis, sentiment analysis, and LLM-as-a-judge classification tasks. We evaluate the learned graphs for predictive fidelity and structural stability, and the MCMC-inspired augmentation for convergence and downstream utility. Our results show that the discovered causal graphs capture meaningful dependencies consistent with LLMs' reasoning. Together, this paper provides a foundation for concept-level explainability of LLMs.
\end{abstract}

\section{Introduction}
\label{sec:intro}

Large Language Models (LLMs) exhibit broad capabilities \citep{bubeck2023sparksartificialgeneralintelligence, DBLP:journals/corr/abs-2005-14165}, yet their inference remains opaque: decision factors are unobservable, and generated explanations often lack faithfulness \citep{FederKMPSWEGRRS22, turpin2023languagemodelsdontsay, GatCFCSR24, DBLP:conf/icml/ZhangPMLS24, DBLP:conf/ijcnlp/LyuHSZRWAC23, DBLP:journals/corr/abs-2307-13702}. This opacity hinders adoption in high-stakes domains, risking confident hallucinations and biases in medicine \citep{omiye2024large}, and fictitious precedents that undermine legal accountability \citep{Dahl_2024}.

\begin{figure}[t]
    \centering
    \includegraphics[width=\linewidth]{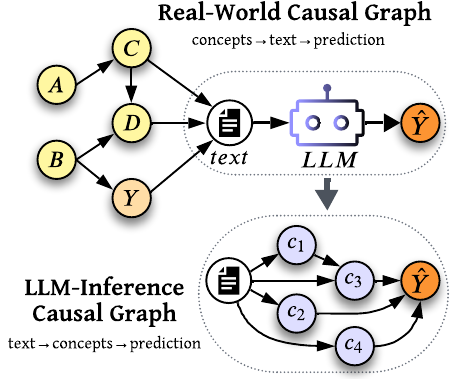}
\caption{Causal graphs provide a language for modeling mechanisms.
Prior work studies real-world structures, using text to infer relations among real-world concepts or estimating how interventions on them (e.g., changing an author's gender) affect model predictions. In contrast, we study the LLM-inference causal graph: a causal graph over LLM-perceived concepts that explains how the model maps text to prediction.}
    \label{fig:introfig}
\end{figure}

LLM opacity has motivated a broad literature on explainability and interpretability \citep{calderon-reichart-2025-behalf}. Attribution and attention-based methods are primarily correlational, whereas faithful explanations require causal evidence \citep{DBLP:journals/corr/abs-1902-10186,zhao2023explainabilitylargelanguagemodels, zevcevic2023causal}. Counterfactual methods provide stronger causal evidence, and recent approaches use causal graphs over textual variables \citep{DBLP:journals/corr/abs-2101-00288, GatCFCSR24, toker2026libertycausalframeworkbenchmarking}. However, they typically remain local and input-centered: they can quantify the causal effect of a given factor on the prediction but do not recover the model's high-level inference. Mechanistic interpretability targets internal components such as neurons and circuits. While this provides a valuable low-level science of neural computation, it is often misaligned with the explanations stakeholders need \citep{calderon-reichart-2025-behalf, 10.1145/3787104}. What remains missing is an applicable explainability method that provides a global, causal account of how a model organizes human-interpretable concepts and reduces them to a prediction.

In this paper, we take a step toward this ``holy grail'' by proposing a fully automated method that combines concept discovery with causal discovery to construct concept-level causal graphs of LLM inference for classification tasks (Figure~\ref{fig:introfig}). A causal graph is a directed graph whose edges encode direct cause-and-effect relationships among variables~\citep{Pearl_00}. In contrast to prior work that uses causal graphs to model relations among real-world variables, we use causal graphs as an interpretability object for the LLM's inference process. Specifically, we construct a graph over the input text, LLM-perceived concept states, and the prediction, which serves as the terminal node. Accordingly, the graph makes explicit how the model organizes concept-level information and how this structure leads to the final prediction.

Our proposed method is model-driven: the target LLM itself identifies latent concepts and generates the data required for causal discovery. Given a set of textual examples and a classification task, our framework implements a four-phase pipeline (Figure~\ref{fig:Full Process2}): (1)~Use the LLM to Generate class predictions for the textual examples;(2)~extracting concepts that differentiate between these predicted classes and representing each textual example with a concept vector; (3)~densely populating the concept space via MCMC-inspired counterfactual data expansion; and (4)~constructing a causal graph over the concepts and the model prediction. 

A central challenge in causal discovery is data coverage: reliable structure learning requires a representative sample that spans diverse combinations of concept values. 
In many domains, obtaining such coverage requires costly, impractical, or unethical interventions. Our setting makes this bottleneck more tractable: because the mechanism under study is an LLM, we can prompt the model to generate counterfactual texts in which a targeted concept is altered while the remaining context is kept as stable as possible. Our MCMC-inspired algorithm iteratively proposes counterfactual chains and rejects unfaithful candidates, producing a denser, higher-quality concept space for robust causal discovery.

We evaluate the framework with three LLMs on three classification tasks spanning natural and synthetic datasets: sentiment analysis, disease diagnosis, and LLM-as-a-judge classification. As no ground-truth graphs or baselines exist for this latent process, we evaluate predictive fidelity and expansion utility: we test whether discovered parents best predict each node, and if expansion improves graph accuracy and stability over the original data.

Our experiments reveal a dichotomy in LLM reasoning: on structured synthetic tasks, models converge on similar explanatory concepts, whereas on naturalistic data, each model develops distinct latent heuristics. We further show that the MCMC-inspired expansion reaches distributional and topological convergence, and that the discovered causal parents are stronger predictors of each node than alternative concept sets.

In summary, this work makes three primary contributions. First, we introduce a causal paradigm for global, concept-based explainability that maps LLM reasoning to causal graphs. Second, we propose an MCMC-inspired data augmentation framework that generates counterfactuals to better cover the concept space, enabling robust causal discovery. Third, we develop an evaluation protocol for assessing both the structural stability and predictive fidelity of the resulting causal graphs.

\section{Related Work}
\label{sec:related_work}

\paragraph{Causal Graphs and Interpretability.}
LLM interpretability methods include feature attribution \citep{DBLP:journals/corr/abs-2107-14000,DBLP:conf/emnlp/LanXHHL25}, attention analysis \citep{DBLP:conf/acl/0008CSLFP24,DBLP:journals/tvcg/YehCWCVW24}, probing \citep{DBLP:journals/corr/abs-2502-04789, DBLP:journals/corr/abs-2506-01042,DBLP:journals/corr/abs-2508-06030}, concept-based explanations \citep{DBLP:conf/icml/KimWGCWVS18, DBLP:conf/aaai/ZhangWLAH25}, and chain-of-thought rationales \citep{DBLP:journals/corr/abs-2501-18645}. These methods are fundamentally associative. Faithful explanations require causal evidence \citep{DBLP:journals/corr/abs-2402-04614}, motivating causal graphs as an explanatory formalism. 
Causal graphs appear in LLM interpretability in three distinct senses. 

\emph{First}, in mechanistic interpretability, causal graphs are defined over model-internal components, such as attention heads, neurons, residual-stream directions, or higher-level representations \citep{DBLP:conf/iclr/WangVCSS23}. These graphs abstract away from the complete computation and retain only the components hypothesized to causally mediate a targeted behavior \citep{DBLP:journals/corr/abs-2408-01416}.

\emph{Second}, causal graphs are used to enable estimation of the causal effects of high-level concepts on model predictions, typically via causal inference methods such as counterfactuals \citep{toker2026libertycausalframeworkbenchmarking}, matching \citep{GatCFCSR24}, or adjustment \citep{DBLP:journals/coling/FederOSR21}.
The graphs are usually assumed in advance (e.g., provided by a domain expert) and describe the data-generating process from concepts to text to model prediction, i.e., concepts $\rightarrow$ text $\rightarrow$ prediction ~\cite{DBLP:conf/emnlp/Paul0BF24}. 

Finally, we introduce a \emph{third} utility, causal graphs to describe the model's high-level reasoning: text $\rightarrow$ concepts $\rightarrow$ prediction. In this setting, the graph is itself the explanation, and both the relevant concepts and their causal relations must be inferred rather than assumed. Our work operates in this setting, and although this utility is promising, it has received no attention in the literature.

\paragraph{Causal Discovery and LLMs.}

Causal discovery aims to recover causal structure from data \citep{Pearl_00}. but is underdetermined from observational data alone: many graphs can induce the same distribution, requiring assumptions such as acyclicity, temporal ordering, or restricted functional forms \citep{10.3389/fgene.2019.00524}. Classical methods include constraint-based, score-based, and hybrid approaches \citep{spirtes2000causation,DBLP:journals/jmlr/Chickering02a,zanga2022survey}, with deep learning extensions for non-linear settings \citep{zheng2018dags, yu2019dag}. We employ the $\sigma$-CG algorithm \citep{forre2018constraint}, which handles discrete variables and potentially cyclic structures, fitting our setting where we cannot impose prior structural or parametric assumptions on how the LLM relates concepts during inference.

Recent work incorporates LLMs into the causal discovery process, but with a different goal from ours. Since causal graphs are fundamental to scientific modeling, these works use LLMs to help discover causal graphs of real-world mechanisms \citep{10.1145/3627673.3680042}. Typically, the term \emph{causal discovery with LLMs} refers to using LLMs to predict whether a causal edge exists between two real-world variables \citep{DBLP:conf/naacl/Ma25}. While promising, various studies have shown that LLMs often memorize high-frequency relations rather than exhibit genuine causal generalization \citep{feng-etal-2025-reliability}.

The closest work to ours is COAT \citep{NEURIPS2024_b99a0748}, which also uses an LLM to propose concepts and then applies causal discovery. However, COAT aims to recover real-world causal structures and, therefore, focuses on local Markov blankets for causal effect estimation~\citep{DBLP:journals/jmlr/AliferisSTMK10}. In contrast, our goal is to characterize the model's internal reasoning, which requires two key algorithmic departures. First, we recover the full graph from text to prediction, rather than only a local Markov blanket, to capture the complete reasoning process. Second, because observational data sparsely cover the model's internal decision space, we introduce an MCMC-inspired counterfactual augmentation procedure that actively expands the concept space for more robust graph discovery.

\section{Method}
\label{sec:method}

We propose a fully automated method that recovers a concept-level causal graph capturing how a target LLM derives its predictions on a classification task. The method is fundamentally model-driven: the same target LLM extracts latent concepts, represents examples by concept vectors, and generates counterfactuals; thus, the resulting graph reflects the model's internal reasoning rather than an external-world process.

\begin{figure*}[t]
    \centering
    \definecolor{phaseblue}{RGB}{205,224,242}
    \definecolor{phaseteal}{RGB}{203,232,227}
    \definecolor{phasegreen}{RGB}{217,232,196}
    \definecolor{phasepink}{RGB}{244,205,205}
    \resizebox{\textwidth}{!}{%
    \begin{tikzpicture}[
        x=1cm,
        y=1cm,
        >=Latex,
        panel/.style={draw=gray!70, rounded corners=8pt, fill=gray!3, line width=0.8pt},
        header/.style={rounded corners=6pt, font=\bfseries\large, align=center, minimum height=0.92cm, text width=4.25cm},
        stepbox/.style={draw=gray!65, rounded corners=5pt, fill=white, align=center, text width=3.95cm, inner sep=5pt, font=\small},
        outbox/.style={draw=gray!65, rounded corners=5pt, fill=blue!8, align=center, text width=3.95cm, inner sep=5pt, font=\small},
        llm/.style={draw, circle, fill=blue!18, minimum size=0.95cm, font=\bfseries\small},
        arrow/.style={->, thick, draw=blue!60!black},
        flow/.style={->, very thick, draw=blue!60!black}
    ]
    \draw[panel] (0,-0.35) rectangle (4.7,8.9);
    \node[header, fill=phaseblue] at (2.35,8.38) {1: Label Prediction};
    \node[stepbox, fill=gray!10] (p1in) at (2.35,6.95) {\textbf{Input text}\\$x=$ \emph{``bright orange, mushy papaya''}};
    \node[llm] (p1llm) at (2.35,5.3) {LLM};
    \node[outbox, fill=red!10] (p1out) at (2.35,3.65) {\textbf{Output}\\$\hat{y}=\text{\emph{not-tasty}}$\\replaces dataset label};
    \draw[arrow] (p1in) -- (p1llm);
    \draw[arrow] (p1llm) -- (p1out);

    \draw[panel] (5.25,-0.35) rectangle (9.95,8.9);
    \node[header, fill=phaseteal] at (7.6,8.38) {2: Concept Discovery\\and Annotation};
    \node[stepbox] (p2in) at (7.6,7.05) {\textbf{Input}\\texts $+$ predicted labels};
    \node[llm, fill=phaseteal] (p2llm) at (7.6,5.85) {LLM};
    \node[stepbox] (p2cand) at (7.6,4.35) {\textbf{Candidates}\\Softness, Color, Origin};
    \node[stepbox, fill=green!10] (p2filter) at (7.6,2.75) {\textbf{Filter concepts}\\relevant: not usually $\emptyset$\\discriminative: not usually $\mathcal{Y}$\\threshold: $1/|\mathcal{Y}|$};
    \node[outbox] (p2out) at (7.6,0.7) {\textbf{Output}\\$\mathcal{C}=\{\text{Softness},\text{Color}\}$\\Softness $\mapsto\{\text{\emph{not-tasty}}\}$\\Color $\mapsto\{\text{\emph{tasty}}\}$};
    \draw[arrow] (p2in.south) -- (p2llm.north);
    \draw[arrow] (p2llm.south) -- (p2cand.north);
    \draw[arrow] (p2cand.south) -- (p2filter.north);
    \draw[arrow] (p2filter.south) -- (p2out.north);

    \draw[panel] (10.5,-0.35) rectangle (15.2,8.9);
    \node[header, fill=phasegreen] at (12.85,8.38) {3: MCMC-Inspired\\Data Expansion};
    \node[stepbox] (p3in) at (12.85,6.95) {\textbf{Input}\\$x^{(0)}$, Softness,\\$y^*=\text{\emph{tasty}}$, $dx=\textsc{More}$};
    \node[llm, fill=phasegreen] (p3llm) at (12.85,5.55) {LLM};
    \node[stepbox] (p3prop) at (12.85,4.05) {\textbf{Proposal $\tilde{x}$}\\\emph{``bright orange, firm papaya''}};
    \node[stepbox, fill=green!10] (p3tests) at (12.85,2.35) {\textbf{Acceptance tests}\\\checkmark\ target alignment\\\checkmark\ minimal side effects};
    \node[outbox, fill=green!8] (p3out) at (12.85,0.85) {\textbf{Output}\\accepted counterfactuals};
    \draw[arrow] (p3in) -- (p3llm);
    \draw[arrow] (p3llm) -- (p3prop);
    \draw[arrow] (p3prop) -- (p3tests);
    \draw[arrow] (p3tests) -- (p3out);

    \draw[panel] (15.75,-0.35) rectangle (20.45,8.9);
    \node[header, fill=phasepink] at (18.1,8.38) {4: Causal Discovery\\via $\sigma$-CG};
    \node[stepbox, fill=green!10] (p4in) at (18.1,6.95) {\textbf{Input}\\expanded annotated data\\$\{(\phi(x),\hat{y})\}$};
    \node[draw, rounded corners=7pt, fill=phaseblue, minimum width=2.6cm, minimum height=0.95cm, font=\bfseries, align=center] (sig) at (18.1,5.45) {$\sigma$-CG};
    \node[draw, circle, fill=green!25, minimum size=0.75cm] (soft) at (17.15,3.55) {\scriptsize Softness};
    \node[draw, circle, fill=green!25, minimum size=0.75cm] (color) at (19.05,3.55) {\scriptsize Color};
    \node[draw, ellipse, fill=blue!15, minimum width=1.6cm, minimum height=0.62cm] (ypred) at (18.1,2.2) {\scriptsize Prediction};
    \draw[arrow] (soft) -- (color);
    \draw[arrow] (soft) -- (ypred);
    \draw[arrow] (color) -- (ypred);
    \node[outbox, fill=phasepink!35] (p4out) at (18.1,0.8) {\textbf{Output}\\concept-level causal graph};
    \draw[arrow] (p4in) -- (sig);
    \draw[arrow] (sig) -- (18.1,4.15);
    \draw[arrow] (ypred) -- (p4out);

    \draw[flow] (4.78,4.45) -- (5.17,4.45);
    \draw[flow] (10.03,4.45) -- (10.42,4.45);
    \draw[flow] (15.28,4.45) -- (15.67,4.45);
    \end{tikzpicture}%
    }
    \caption{Overview of our four-phase pipeline for constructing causal graphs. The ``papaya'' running example is drawn from the full toy walkthrough in Appendix~\ref{appendix:sec:app_running_example}.
    }
    \label{fig:Full Process2}
\end{figure*}
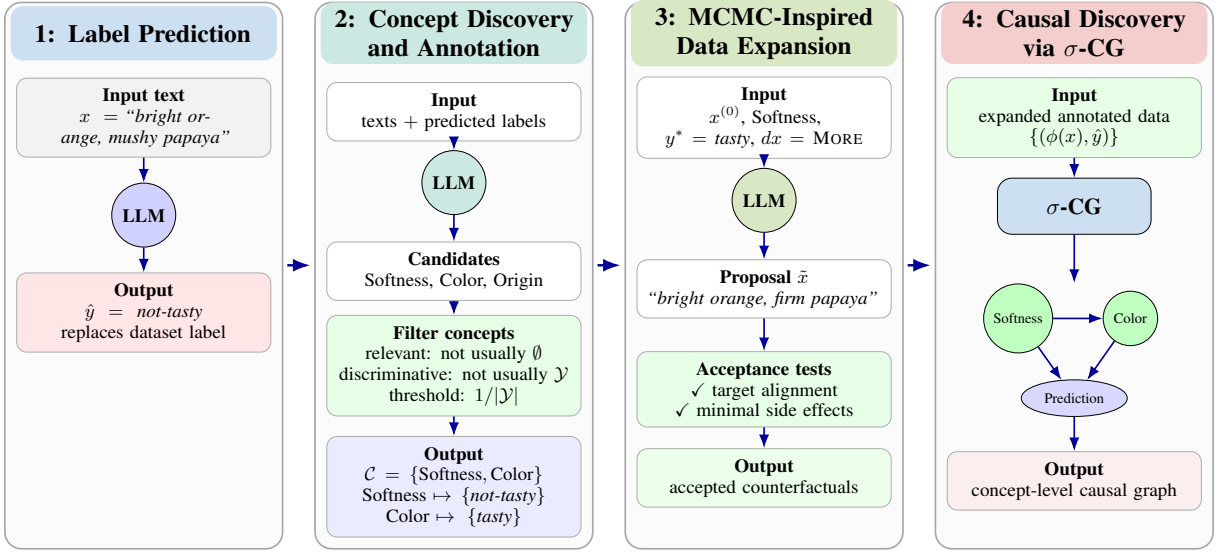

Our method, illustrated in Figure~\ref{fig:Full Process2}, employs a four-phase pipeline: 
\textbf{Label Prediction} substitutes ground-truth labels with the LLM’s own predictions. 
\textbf{Differentiative Concept Extraction} iteratively identifies a set of concepts that differentiate between predicted classes, representing each example as a concept vector. 
\textbf{MCMC-Inspired Data Expansion} populates the concept space with counterfactuals to ensure coverage. 
Finally, \textbf{Causal Discovery} constructs the final graph over these concepts and the model prediction.

\subsection{Problem Statement and Formulation}
\label{sub:formulation}

Given a target LLM $f$ acting as a classifier and a dataset $\mathcal{D}=\{(x^{j},y^{j})\}_{j=1}^{N}$ of textual examples, split into $\mathcal{D}_{\mathrm{train}}$ and $\mathcal{D}_{\mathrm{test}}$, where $x$ represents the input text and $y \in \mathcal{Y}$ represents the ground-truth label (with $\mathcal{Y}$ being the set of task classes), our goal is to recover a concept-level causal graph that explains how $f$ derives its prediction $\hat{y}=f(x)$.

We adopt the Causal Graph framework of \citet{Pearl_00}, relaxed to admit cyclic dependencies \citep{Bongers_2021}, acknowledging potential reciprocal influences among the concepts an LLM uses internally. Let $\mathcal{C} = \{c_1, \dots, c_n\}$ denote a set of human-interpretable concepts. We define a concept-extraction function $\phi$ that maps an input text $x$ to a \emph{concept vector} of length $n$: 
\[\phi(x) = [l_{c_1}, l_{c_2}, \dots, l_{c_n}]\] 
where each $l_{c_i}$ represents the LLM-perceived state of concept $c_i$ in the context of $x$. Specifically, $l_{c_i}$ indicates whether the concept is absent and, if present, which task classes the model perceives it to support. We represent each $l_{c_i}$ as a categorical variable taking values in $\mathcal{V} =\{0, 1, \dots, 2^{|\mathcal{Y}|}-1\}$,
where each value corresponds bijectively to a subset $S \subseteq \mathcal{Y}$. The subset $S$ denotes the classes with which concept $c_i$ is perceived to be aligned. The value corresponding to $S=\emptyset$ indicates that the concept is absent from $x$ or irrelevant, while the value corresponding to $S=\mathcal{Y}$ indicates that the concept is present and aligns with all possible classes without differentiation.

For example, in a disease diagnosis task with 
$\mathcal{Y}\!=\!\{\text{Migraine}, \text{Sinusitis}, \text{Influenza}\}$, suppose $c_1$ denotes the concept \emph{Headache}. If $\phi(x)[c_1]$ takes the value corresponding to $S\!=\!\{\text{Migraine}, \text{Sinusitis}\}$, then the model perceives this concept as present in $x$ and as evidence for both Migraine and Sinusitis. If it takes the value corresponding to $S=\emptyset$, the concept is perceived as absent; if it takes the value corresponding to $S=\mathcal{Y}$, the concept is perceived as present but non-discriminative among the candidate labels.

Finally, we represent the LLM's inference process as a directed graph over the input text, concept variables, and final prediction:
\[
\mathcal{G}=(V,E), \qquad V = \{X\}\cup\mathcal{C}\cup\{\hat{Y}\}.
\]
The graph captures a high-level causal flow from the input text \(X\), through LLM-perceived concepts \(\mathcal{C}\), to the model prediction \(\hat{Y}\), while allowing dependencies among concepts. We impose three structural constraints: \(X\) has only outgoing edges and serves as the root of the graph;\footnote{For simplicity, we visualize graphs without the text node.} \(\hat{Y}\) has only incoming edges, and the effect of the input text on the prediction is mediated by the concept variables. Accordingly, our task is to recover the concept set \(\mathcal{C}\) and the edge set \(E\).

\subsection{Concept Discovery and Annotation}
\label{sub:concept_discovery}

\paragraph{Label Prediction.}
We first replace the ground-truth labels in $\mathcal{D}$ with the LLM's predictions $\hat{y}=f(x)$. All downstream phases use these predicted labels, so the recovered graph reflects the model's perspective rather than the dataset labels (Box \ref{prompt:appendix:LLMasJudge}).

\paragraph{Discriminative Concept Discovery.}
We extract a set of concepts $\mathcal{C}$ that differentiate among the classes in $\mathcal{Y}$. The procedure processes $\mathcal{D}_{\mathrm{train}}$ in small, class-balanced batches, where each batch $B$ contains $|B|/|\mathcal{Y}|$ instances from each class. For each batch, the model is shown the examples and proposes new candidate concepts, which are added to $\mathcal{C}$ (prompts in Boxes~\ref{prompt:ExtractConcepts1stBatch} and~\ref{prompt:ExtractAddtionalConcepts}).
Every ten batches, we filter the accumulated concepts. To do so, we first annotate each example seen so far with a concept vector $\phi(x)$ (prompt in Box E.4), where each concept takes a scalar value from $\mathcal{V}$, corresponding to a subset  $S \subseteq \mathcal{Y}$. We then retain only concepts that are both relevant and discriminative. Specifically, a concept is retained if the fraction of examples for which its assigned value maps to a strictly partial subset of classes (i.e., neither $S = \emptyset$ nor $S = \mathcal{Y}$) exceeds the threshold $\tau = 1/|\mathcal{Y}|$.

Intuitively, this removes concepts that are rarely expressed in the text and therefore provide little evidence for a text-to-concept edge, as well as concepts that are broadly aligned with all classes and therefore provide little evidence for a concept-to-prediction edge. After processing all batches in $\mathcal{D}_{\text{train}}$, we annotate the examples in $\mathcal{D}_{\text{test}}$ with concept vectors and apply the same filtering criterion once more, yielding the final concept set $\mathcal{C}$. Algorithm~\ref{alg:concept_extraction} in the appendix provides the full procedure.

\subsection{Data Expansion and Causal Discovery}
\label{sub:expansion_and_discovery}

Causal discovery from observational data requires samples that adequately cover the joint variable space $(\phi(X), \hat{Y})$. Since the mechanism under study is an LLM, we can cheaply query it to generate textual counterfactuals. Our expansion phase, therefore, constructs chains of counterfactual texts, producing concept vectors that more densely populate the relevant regions of the concept space.


\paragraph{Markov Chain in Text Space.}
Markov Chain Monte Carlo (MCMC) methods provide a principled way to explore complex, high-dimensional spaces by constructing chains whose samples approximate a target distribution \citep{DBLP:books/sp/RobertC04, DBLP:journals/ml/AndrieuFDJ03}. Inspired by this idea, we introduce a targeted data-expansion algorithm that starts from a sparse initial dataset and generates a denser set of examples over the relevant domain. Rather than operating directly in the discrete concept space $\mathcal{V}^n$, the algorithm operates in the raw text space: at each step, the LLM proposes a counterfactual, i.e., a textual perturbation, which is then mapped to a concept vector by $\phi$ and retained only if it induces the intended conceptual shift. Operating in text space preserves linguistic coherence while allowing us to explore realizable regions of $\mathcal{V}^n$, i.e., concept configurations that can be expressed by natural text. 

Each original example $x \in \mathcal{D}_{\mathrm{train}}$ initiates an independent expansion process lasting $K=11$ steps. At each step, for a current example $x$, we iterate through all $c_i \in \mathcal{C}$ and sample a target class $y^* \in \mathcal{Y}$. Let $S \subseteq \mathcal{Y}$ be the subset of classes mapped to by the scalar concept value $\phi(x)[c_i]$. We then choose a directional shift $dx \in \{\textsc{More}, \textsc{Less}\}$ according to whether $y^*$ is included in this subset: $dx=\textsc{More}$ if $y^* \notin S$, and $dx=\textsc{Less}$ otherwise. Thus, the LLM-generated \emph{counterfactual proposal} $\tilde{x}$ either attempts to introduce an alignment between $c_i$ and $y^*$ or to remove an existing one.

For example, if $x$ is
\emph{``I feel a strong headache that gets worse in the light''},
the target concept is \emph{SensitivityToLight}, and $y^\star$ is \emph{Migraine}. If the LLM aligns this concept with \emph{Migraine}, the sampled direction is $dx=\textsc{Less}$. The LLM is then prompted to rewrite the text to reduce the alignment between \emph{SensitivityToLight} and \emph{Migraine}, while keeping other concepts fixed. A possible proposal $\tilde{x}$ is \emph{``I feel a strong headache since this morning.''}.

\paragraph{Acceptance Criteria.}
A proposal $\tilde{x}$ is appended to $\mathcal{D}_{\text{mcmc}}$ if it satisfies two conditions, akin to a Metropolis--Hastings test \citep{Hastings1970MonteCS}: \textbf{(i) Target alignment}: We compute $\phi(\tilde{x})$ (concurrently eliciting reasoning for later use). Let $\tilde{S} \subseteq \mathcal{Y}$ be the subset mapped from $\phi(\tilde{x})[c_i]$. We require $y^* \in \tilde{S}$ if $dx=\textsc{More}$, and $y^* \notin \tilde{S}$ otherwise. \textbf{(ii) Minimal side effects}: The count of drifted non-target concepts must be bounded by a tolerance $\epsilon \in \{1,2\}$, allowing for natural causal correlations: $\sum_{c_j\neq c_i}\mathbb{I}[\phi(\tilde{x})[c_j]\neq\phi(x)[c_j]] \le \epsilon$.

\begin{figure*}[t]
    \centering
    \includegraphics[width=0.98\textwidth]{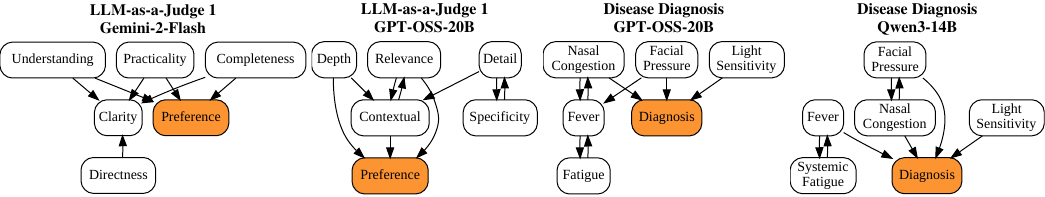}
    \caption{Illustrations of four causal graphs. All causal graphs are provided in Appendix\ref{appendix:sec:CG}.}
    \label{fig:cg1}
\vspace{-1em}
\end{figure*}

\paragraph{Recursive Refinement.}
If a proposal fails either acceptance condition, we re-prompt the LLM with feedback indicating which criterion was violated, together with the reasoning produced during the annotation of $\tilde{x}$. For each concept, the chain attempts up to $R=5$ regenerations. If all attempts fail, the chain proceeds without appending a sample for that concept. In summary, each original example can contribute up to $K|\mathcal{C}|$ counterfactual examples to $\mathcal{D}_{\mathrm{mcmc}}$, though the actual number may be smaller when proposals fail the acceptance criteria after all refinement attempts. The full MCMC procedure is described in Algorithm~\ref{alg:mcmc_expansion} in Appendix~\ref{appendix:sec:app_pseudo_algorithms}.

\paragraph{Rationale.} Our MCMC-inspired expansion densifies the latent concept space by exploring combinations beyond the original dataset. Starting from each example, the chain iteratively proposes local changes to concept assignments and prompts the LLM to generate corresponding counterfactuals. Proposals are accepted only if the generated example is valid and matches the intended configuration; otherwise, they are rejected to avoid inconsistent or unrealizable regions. Thus, the procedure stochastically explores the valid concept manifold, providing richer support for causal discovery.

\paragraph{Causal Discovery via $\sigma$-CG.}
We apply the $\sigma$-CG causal discovery algorithm \citep{forre2018constraint} to the expanded, annotated dataset $\mathcal{D}_{\mathrm{mcmc}}$, obtaining a directed graph $\mathcal{G}$ over the variables $V$. We choose $\sigma$-CG because it accommodates cyclic causal structures, which may arise in LLM inference and therefore cannot be ruled out a priori, and because it supports the discrete variables produced by our concept annotations. As background knowledge, we impose that $\hat{y}$ is the unique sink node, and enforce this constraint during edge orientation.

\section{Experimental Setup}
\label{sec:experimental}

\paragraph{Models.} We evaluate our framework and explain three LLMs: Gemini-2-Flash~\cite{geminiteam2025geminifamilyhighlycapable}, Qwen3-14B \citep{yang2025qwen3technicalreport}, and gpt-OSS-20b \citep{openai2025gptoss120bgptoss20bmodel} For the open-weights models (Qwen3-14B and gpt-OSS-20b) we use the vLLM framework~\cite{kwon2023efficientmemorymanagementlarge}. For each model, we use two temperature settings: $\mathcal{M}_{cls}$ with $\tau=0$ for deterministic inference and reproducible labeling, and $\mathcal{M}_{gen}$ with $\tau=0.5$ for concept extraction and MCMC-based counterfactual generation.

\paragraph{Datasets.}We evaluate on three classification tasks, scaling batch sizes by input length and class count. (1)~\emph{Disease Diagnosis (DD, $N\!=\!1448$, $B=9$)}: the LIBERTY dataset, a synthetic medical corpus in which patient descriptions are classified as Migraine, Sinusitis, or Influenza \citep{toker2026libertycausalframeworkbenchmarking}. (2)~\emph{Sentiment Analysis (SA, $N\!=\!2096$, $B=10$)}: the IMDB dataset, in which movie reviews are classified as Positive or Negative \citep{maas-EtAl:2011:ACL-HLT2011}. (3)~\emph{LLM-as-a-Judge (LAJ, $N\!=\!395$, $B=10$)}: a preference prediction task constructed from Reddit, where each input contains a question and two candidate responses and the model selects the preferred one ~\cite{CalderonER25}. To mitigate positional bias, each pair is presented twice with the response order swapped.
For Disease Diagnosis and Sentiment Analysis, we construct a single causal graph per task, capturing the reasoning structure that the model applies uniformly across all inputs in the domain. LAJ, however, poses a distinct challenge: queries span unrelated topics (e.g., nutrition, programming, travel advice), and the criteria governing the model's preference for one query bear little relation to those for another. A single graph cannot meaningfully capture such heterogeneous reasoning. We therefore adopt a \emph{query-level} approach: for each query, we generate a large set of diverse response pairs, forming a self-contained dataset that captures how the model adjudicates responses \emph{to that specific question}. We then construct a separate causal graph per query, explaining what drives the model's preference within a fixed topical context (see Appendix~\ref{appendix:subsec:datasets}).

\begin{table*}[t]
\centering
\normalsize
\begin{adjustbox}{width=0.85\textwidth}
\begin{tabular}{l|ccc|ccc|ccc}
\toprule
\cellcolor{Gray!15}
& \multicolumn{3}{c|}{\cellcolor{Gray!40}\textbf{Disease Diagnosis}}
& \multicolumn{3}{c|}{\cellcolor{Gray!40}\textbf{Sentiment Analysis}}
& \multicolumn{3}{c}{\cellcolor{Gray!40}\textbf{LLM-as-a-Judge}} \\
\cellcolor{Gray!15} \textbf{Target}
& \cellcolor{Gray!15} \shortstack{CG\\Acc.}
& \cellcolor{Gray!15} \shortstack{Others\\Acc.}
& \cellcolor{Gray!15} \shortstack{CG in\\Top-3}
& \cellcolor{Gray!15} \shortstack{CG\\Acc.}
& \cellcolor{Gray!15} \shortstack{Others\\Acc.}
& \cellcolor{Gray!15} \shortstack{CG in\\Top-3}
& \cellcolor{Gray!15} \shortstack{CG\\Acc.}
& \cellcolor{Gray!15} \shortstack{Others\\Acc.}
& \cellcolor{Gray!15} \shortstack{CG in\\Top-3} \\
\midrule
\multicolumn{10}{l}{\cellcolor{Gray!30}\textbf{\textit{Gemini-2-Flash}}} \\
Prediction $\hat{y}$
& \textbf{0.67} & 0.61 & \textcolor{ForestGreen}{\textbf{100.0\%}}
& \textbf{0.80} & 0.77 & \textcolor{ForestGreen}{\textbf{80.0\%}}
& \textbf{0.79} & 0.74 & \textcolor{ForestGreen}{\textbf{57.5\%}} \\
Concepts $\mathcal{C}$
& \textbf{0.54} & 0.51 & \textcolor{ForestGreen}{\textbf{96.7\%}}
& \textbf{0.61} & 0.58 & \textcolor{ForestGreen}{\textbf{95.0\%}}
& \textbf{0.83} & 0.72 & \textcolor{ForestGreen}{\textbf{80.8\%}} \\
\midrule
\multicolumn{10}{l}{\cellcolor{Gray!30}\textbf{\textit{GPT-OSS 20B}}} \\
Prediction $\hat{y}$
& \textbf{0.67} & 0.59 & \textcolor{ForestGreen}{\textbf{100.0\%}}
& \textbf{0.84} & 0.78 & \textcolor{ForestGreen}{\textbf{100.0\%}}
& \textbf{0.83} & 0.78 & \textcolor{ForestGreen}{\textbf{53.0\%}} \\
Concepts $\mathcal{C}$
& \textbf{0.53} & 0.48 & \textcolor{ForestGreen}{\textbf{76.7\%}}
& \textbf{0.60} & 0.55 & \textcolor{ForestGreen}{\textbf{78.0\%}}
& \textbf{0.77} & 0.67 & \textcolor{ForestGreen}{\textbf{86.2\%}} \\
\midrule
\multicolumn{10}{l}{\cellcolor{Gray!30}\textbf{\textit{Qwen3-14B}}} \\
Prediction $\hat{y}$
& \textbf{0.63} & 0.56 & 50.0\%
& \textbf{0.80} & 0.75 & \textcolor{ForestGreen}{\textbf{80.0\%}}
& \textbf{0.64} & 0.62 & \textcolor{ForestGreen}{\textbf{64.4\%}} \\
Concepts $\mathcal{C}$
& \textbf{0.65} & 0.54 & \textcolor{ForestGreen}{\textbf{92.5\%}}
& \textbf{0.61} & 0.60 & \textcolor{ForestGreen}{\textbf{97.5\%}}
& \textbf{0.88} & 0.80 & \textcolor{ForestGreen}{\textbf{82.9\%}} \\
\multicolumn{10}{l}{\cellcolor{Gray!30}\textbf{\textit{Random Baseline}}} \\
Prediction $\hat{y}$
& 0.33 & 0.33 & $\approx$12\%
& 0.50 & 0.50 & $\approx$12\%
& 0.50 & 0.50 & $\approx$12\% \\
Concepts $\mathcal{C}$ 
& 0.12 & 0.12 & $\approx$27\%
& 0.25 & 0.25 & $\approx$27\% 
& 0.25 & 0.25 & $\approx$27\% \\
\bottomrule
\end{tabular}
\end{adjustbox}
\caption{\textbf{Causal-Graph Prediction Results:} Average 10-fold cross-validation results. In ``Prediction $\hat{y}$'', logistic regression predicts the task label; in ``Concepts $\mathcal{C}$'', we train one logistic regression per concept to predict its value and report the mean accuracy across concepts. ``CG Acc.'' uses the causal-graph parent concepts as inputs, ``Others Acc.'' averages over concept combinations that do not contain the full parent set, and ``CG in Top-3'' is the percentage of cases in which CG Acc. ranks among the top three concept combinations.}
\label{tab:cg-prediction-results}
\end{table*}

\section{Results}
\label{sec:results}


\subsection{Predictive Fidelity Evaluation}
\label{subsec:fidelity}

Our goal is to recover a causal graph over the LLM’s latent inference process, and there is naturally no ground-truth graph for direct evaluation. We therefore evaluate the learned graphs through \emph{predictive fidelity}: If a graph captures meaningful dependencies in the model's inference, the direct causal parents should screen off indirect variables, making the parents of each node more predictive of that node than other concept subsets.

We test this using a post hoc prediction task. For each target node \(v \in \mathcal{C} \cup \{\hat{Y}\}\), we train a multinomial logistic regression model to predict \(v\) from a one-hot encoding of its graph-based parent set \(Pa(v)\). We compare it to models trained on all possible concept subsets \(Z \subseteq \mathcal{C}\) that do not fully contain the parent set, i.e., \(Pa(v) \not\subseteq Z\). We evaluate both on the final model output \(\hat{Y}\) and the concept nodes \(c_i \in \mathcal{C}\), using 10-fold cross-validation (CV).

As shown in Table~\ref{tab:cg-prediction-results}, graph-based predictors outperform the average alternative subset in every setting, across all models, datasets, and target types. This indicates that the discovered parent sets capture information relevant to both final predictions and concept-level dependencies. The Top-3 results further strengthen this conclusion: in every setting but one, the causal-graph parent set ranks among the best predictors in a majority of CV cases, despite being compared against tens of alternatives. 

\begin{figure}[t]
    \centering
    \includegraphics[width=\linewidth]
    {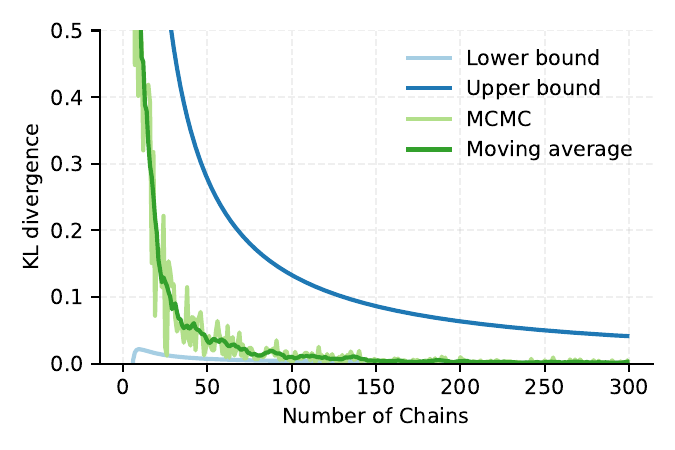}
    \caption{KL-divergence convergence during MCMC-inspired data expansion on Sentiment Analysis for GPT-OSS-20B. The empirical curve is compared against two bounds: perfect overlap (samples fall within the covered regions, lower bound) and orthogonal expansion (samples occupy unseen regions, upper bound).}
    \label{fig:KL con IMDB-OSS}
    \vspace{-1em}
\end{figure}

\subsection{Analysis of the Causal Graphs}
\label{subsec:causal_graphs_analysis}

We next discuss the learned concept sets and causal graphs. On the synthetic Disease Diagnosis dataset, the extracted concepts are consistent across LLMs. As shown in Table~\ref{tab:extracted_concepts}, they successfully recover the ground-truth variables (e.g., specific symptoms) defined in the causal graph used to generate the dataset \citep{toker2026libertycausalframeworkbenchmarking}. The learned graphs also reveal that different LLMs can follow different concept-level reasoning patterns. Even when models rely on similar concepts, their causal topologies differ (Figure \ref{fig:cg1}). This model-specific structure is even more pronounced in the naturalistic Sentiment Analysis and LAJ tasks, where the extracted concepts themselves vary across models (See Table~\ref{tab:extracted_concepts}), suggesting that models rely on different reasoning mechanisms. To summarize, these differences matter for deployment: choosing among models requires understanding not only their performance but also whether their underlying reasoning aligns with stakeholders' expectations and constraints.

\begin{figure*}[t]
    \centering
    \includegraphics[width=\textwidth]{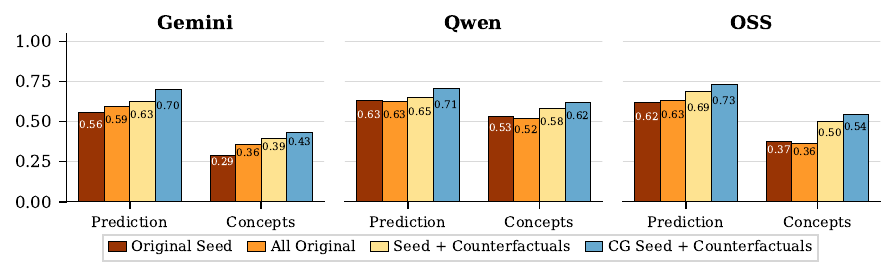}
        \caption{\textbf{Impact of MCMC:} Mean accuracy across DD and SA datasets and all concept combinations, under three training regimes: All Original (full original dataset), Original Seed (subset used as a seed for the MCMC), and Seed + Counterfactuals. For each target $v \in \mathcal{C} \cup \{\hat{y}\}$, a logistic regression is trained over all possible concept combinations. The blue bars show the accuracy when the input consists of the parents based on the causal graph.}
    \label{fig:validation_test_bar}
    \vspace{-0.8em}
\end{figure*}

\subsection{MCMC Convergence and Stability.}
\label{subsec:stablitiy}

Multi-chain diagnostics such as the Gelman-Rubin statistic are not directly applicable in our single-chain-per-instance setting. Instead, we monitor convergence by tracking the empirical probability distribution over the \(|\mathcal{V}|^{|\mathcal{C}|}\) combinatorial concept state space. Following each iteration, we calculate the Kullback-Leibler (KL) divergence between the updated global probability vector and the previous one.
A naive KL decrease is not sufficient evidence of convergence: as the sample pool grows, each new counterfactual carries diminishing marginal weight, mechanically reducing the KL regardless of whether the chain has genuinely stabilized. To separate genuine convergence from this artifact, we bound the empirical KL between two extremes (Equations~\ref{eq:appendix:convergance},~\ref{eq:appendix:non_convergeance}, in Appendix~\ref{appendix:subsec:app_mcmc_convergence}): an orthogonal expansion upper bound, where every new sample occupies a previously unseen concept state, and a perfect overlap lower bound, where every new sample duplicates an existing one. 

A desirable trajectory begins near the upper bound, indicating that the chain is actively discovering new valid concept states, and gradually approaches the lower bound, indicating that the accepted samples increasingly come from a stable, realizable region. As shown in Figure~\ref{fig:KL con IMDB-OSS}, our empirical KL follows precisely this trajectory.
Finally, we verify that this distributional stabilization is accompanied by structural stability: once the KL stabilizes, the Hamming distance between causal edge sets recovered across successive iterations drops to zero (Figures~\ref{fig:appendix kl_conv},~\ref{fig:appendix:SHD}), suggesting that additional augmentation no longer changes the learned graph.

\subsection{Impact of Counterfactual Augmentations}

We evaluate the contribution of MCMC expansion by comparing three data regimes: (i) \textit{All Original} (the full original dataset); (ii) \textit{Original Seed} (the subset seeding the MCMC chains); and (iii) \textit{Seed + Counterfactual}. For each regime and each node \(v \in \mathcal{C} \cup \{\hat{Y}\}\), we train a multinomial logistic regression predicting $v$ from every possible subset of the remaining concepts. Figure~\ref{fig:validation_test_bar} reports prediction accuracy averaged first over all concept subsets, then across SA and DD datasets; we also present the accuracy of the parent subset identified by the causal graph. Note that, unlike in \S\ref{subsec:fidelity}, this time we use a different held-out test set, and the average accuracy over combinations includes the parent subset itself and all of its supersets.

Two patterns emerge. First, the counterfactuals achieve the highest predictive accuracy for both concept nodes and prediction across all models. 
The MCMC-generated counterfactuals introduce concept-state combinations that are absent in the original data, enabling more accurate estimation of the dependencies between concepts. Second, consistent with \S\ref{subsec:fidelity}, the causal graph's parent subsets outperform the average over all concept subsets. Together, these results demonstrate the necessity of our counterfactual augmentation stage.

\section{Conclusion}
\label{sec:conclusion}

In this paper, we propose a novel causal framework for LLM explainability that models the inference process with a causal graph over human-interpretable concepts. We believe this representation can provide a more accessible and actionable form of interpretability for both model developers and domain experts using LLMs in practice.
A central technical contribution of this work is our MCMC-inspired data expansion algorithm, designed to address the sparsity of observational data in latent concept spaces. By iteratively generating counterfactuals, the method produces denser and more representative datasets that facilitate accurate and stable causal discovery. 
Our experiments further address the challenge of evaluation in the absence of gold-standard reasoning graphs by introducing predictive and structural validation protocols that consistently demonstrate both the quality of the discovered causal graphs and the contribution of the MCMC-inspired expansion process.

More broadly, we hope this work will contribute to a new generation of process-oriented explainability methods for LLMs, focused on modeling inference mechanisms rather than explaining isolated predictions. In future work, we plan to extend the framework to additional languages, open-ended generation settings, and multimodal domains, while further broadening the empirical evaluation to move toward more standardized and stakeholder-oriented explainability frameworks.

\section{Limitations}
\label{sec:limitations}

\paragraph{Sensitivity to Batch Pairing in Concept Extraction.}
Concept extraction is performed in small batches due to LLM output token limits, and we run the process only once due to inference costs. As a result, the candidate concepts may depend on the arbitrary grouping of examples within each batch. Different batch assignments could highlight distinct contrastive features and yield a different or more comprehensive set of concepts $\mathcal{C}$. Future work could mitigate this with multiple shuffled extraction passes when computational budget allows.

\paragraph{Restricted Evaluation to Local Parents.}
Our framework is designed to recover a global concept-level causal graph, but the current validation protocol primarily evaluates local predictive fidelity through discovered parent sets. This demonstrates that the immediate predictors of each node are informative and structurally stable, but it does not directly verify longer multi-hop causal chains or the broader hierarchy of the recovered graph. 

\paragraph{Dependence on Self-Annotation and Generation.}
During counterfactual generation, we rely on the target LLM to annotate concepts, generate counterfactuals, and judge whether proposed edits satisfy the intended constraints. Because LLM-generated reasoning and self-assessments are not guaranteed to be faithful, errors in these intermediate steps may propagate. Although our acceptance criteria reduce this risk by checking for target alignment and concept drift, future work should investigate external validation, human audits, or multi-model agreement to further assess the quality of annotations and counterfactuals.
\section{Acknowledgment}
\label{sec8:acknowlegment}
This research was funded by the IBM-Technion Grant Program on Natural Language Processing

\bibliography{custom}

\appendix

\appendix

\section{Supplementary Results and Extended Discussion}
\label{sec:app_supplementary_results}

\subsection{Datasets}
\label{appendix:subsec:datasets}

\paragraph{Dataset Scale and Subsampling.}
Our evaluation spans three core classification tasks: Disease Diagnosis (DD), Sentiment Analysis (SA), and LLM-as-a-Judge (LAJ). For the Sentiment Analysis task, we leverage the IMDB movie review corpus \citep{maas-EtAl:2011:ACL-HLT2011}. Given the computational and API cost constraints of running LLM-based pipelines over large corpora, we subsample the IMDB corpus to a representative subset of $N=2096$ instances.

\paragraph{Feature Isolation and Textual Extraction.}
We retain only the core textual content relevant to each classification task, discarding all metadata and structural fields. For Disease Diagnosis, we use raw patient descriptions from the LIBERTY benchmark's synthetic medical corpus \citep{toker2026libertycausalframeworkbenchmarking} ($N=1448$), predicting among Migraine, Sinusitis, and Influenza. For Sentiment Analysis, we use raw movie review texts from IMDB ($N=2096$), classified as Positive or Negative. For LLM-as-a-Judge, we retain the primary user query and two candidate responses sourced from Reddit \citep{CalderonER25} ($N=395$).

\paragraph{Handling Positional Bias in LAJ} The LAJ task is susceptible to positional bias: we observed that swapping the order of candidate responses altered the model's
classification outcome in over 30\% of instances. To mitigate this, we present each pair twice with the response order inverted during both the inference and labeling alignment phase and the data expansion phase. Concretely, for each dataset instance we run two independent
MCMC chains in parallel, one per response ordering, ensuring the expanded counterfactual dataset is balanced across both permutations.
\subsection{Concept Extraction}
\label{appendix:subsec:app_concept_extraction}
\begin{table*}[t]
  \centering
  \begin{tabular}{llp{10cm}}
    \hline
    \textbf{Dataset} & \textbf{Model} & \textbf{Extracted Concepts} \\
    \hline
    Sentiment Analysis & Qwen3-14B & Enjoyment, Audience Appeal, Expectation Management, Emotional Impact, Performance Quality \\
     & Gemini-2-Flash & Enjoyment, Fulfilment, Execution, Sincerity, Focus \\
     & \textit{gpt-OSS-20b} & Recommendation, Enjoyment, Tone, AudienceEngagement, OverallQuality \\
    \hline
    Disease Diagnosis & Qwen3-14B & Fever, Systemic Fatigue, Facial Pressure, Nasal Congestion, Light Sensitivity \\
     & Gemini-2-Flash & Fever, Facial Pressure, Nasal Congestion, Light Sensitivity \\
     & gpt-OSS-20b &  Fever, Fatigue, Facial Pressure, Nasal Congestion, Light Sensitivity \\
    \hline
    LAJ (Food) & Qwen3-14B & Usefulness and Practicality, Clarity and Specificity, Relevance and Focus, Tone and Professionalism, Feasibility and Realism \\
     & Gemini-2-Flash & Directness, Practicality, Completeness, Clarity, Understanding \\
     & \textit{gpt-OSS-20b} & Relevance, Depth, Specificity, Contextual, Detail \\
    \hline
  \end{tabular}
  \caption{Extracted concepts across different models and datasets. Each set of concepts represents the latent features identified by the specific model as most differentiative for the given task.}
  \label{tab:extracted_concepts}
\end{table*}

Table~\ref{tab:extracted_concepts} details the differentiative concepts extracted for every combination of model and dataset. Notably, our results indicate that within the synthetic DD benchmark, the extracted concepts are remarkably similar across different models and faithfully reconstruct the base graph utilized for data generation~\cite{toker2026libertycausalframeworkbenchmarking}. Conversely, the concepts derived from the natural datasets display a higher degree of model-specificity, highlighting how individual models leverage distinct internal representations for the same task.

\subsection{Data Expansion}
\label{subsec:app_data_expansion}
\label{appendix: naive expansion}

When evaluating the LAJ dataset, our framework constructs a distinct causal graph for each individual query. Consequently, the initial dataset $\mathcal{D}$ for each graph effectively consists of only a single seed example. In our evaluation, we analyze a subset of queries for each model, resulting in $Q=48$ unique graphs for Gemini, $Q=18$ for QWEN, and $Q=10$ for GPT-OSS. Under these conditions, the observed distribution of concepts is fundamentally insufficient for reliable causal discovery. To mitigate this cold-start problem and artificially inflate the data, we apply a ``coarse expansion'' phase as a mandatory first step. The objective of this phase is to enrich the dataset by guiding the target LLM $f$ to express the candidate concepts across a variety of classification targets.

Let $\mathcal{Y}$ denote the set of all possible class labels. For each initial data instance $x$ and each candidate concept $c_i \in \mathcal{C}$, we first evaluate its current alignment context, denoted as $\phi(x)[c_i] \in \mathcal{V}$. The specific value of this alignment dictates the direction of our textual perturbations. If the concept $c_i$ is currently aligned with a single, specific class $y \in \mathcal{Y}$ (i.e., $\phi(x)[c_i] \mapsto \{y\}$), we aim to observe how the concept behaves under different target labels. Thus, we prompt the model $f$ to rewrite the original text $x$ into new variations, specifically directing it to perturb the instance toward all remaining classes in the complement set $\mathcal{Y} \setminus \{y\}$. Conversely, if the concept is neutral and currently appears identically across all classes (i.e., it is aligned with the full set, $\phi(x)[c_i] \mapsto \mathcal{Y}$), we lack information about its discriminative boundaries. In this case, the expansion is executed across the entire label space, and the model is prompted to generate distinct variations targeted at every individual class in $Y$.

This perturbation strategy is deliberately exhaustive. At this preliminary stage, we merely instruct the LLM to shift the context toward alternative labels without enforcing any strict downstream constraints on the generation process (the exact prompt template utilized for this step is detailed in Box~\ref{prompt:dataexpensionStage1}). By generating these counterfactual examples, we create a denser joint distribution over $\phi$, which serve is the starting point of the MCMC expansion phase

\subsection{Validation Details}
\label{subsec:app_validation_details}

\paragraph{Fidelity and Predictive Validation.}
\label{par:app_fidelity_validation}
\begin{figure*}[t]
    \centering

    \begin{subfigure}{0.32\linewidth}
        \includegraphics[width=\linewidth]{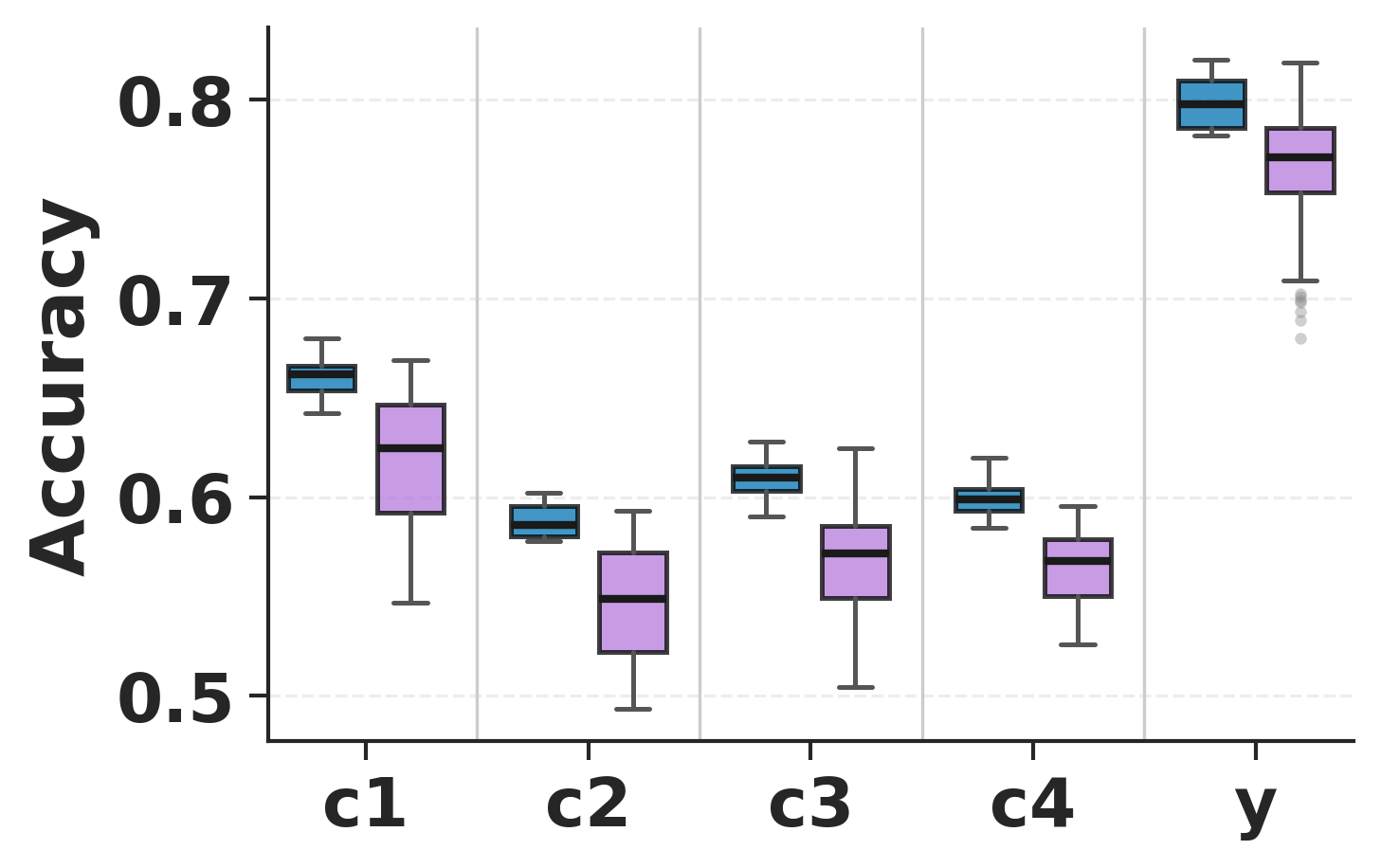}
        \caption{Dataset: SA, Model: Gemini}
    \end{subfigure}\hfill
    \begin{subfigure}{0.32\linewidth}
        \includegraphics[width=\linewidth]{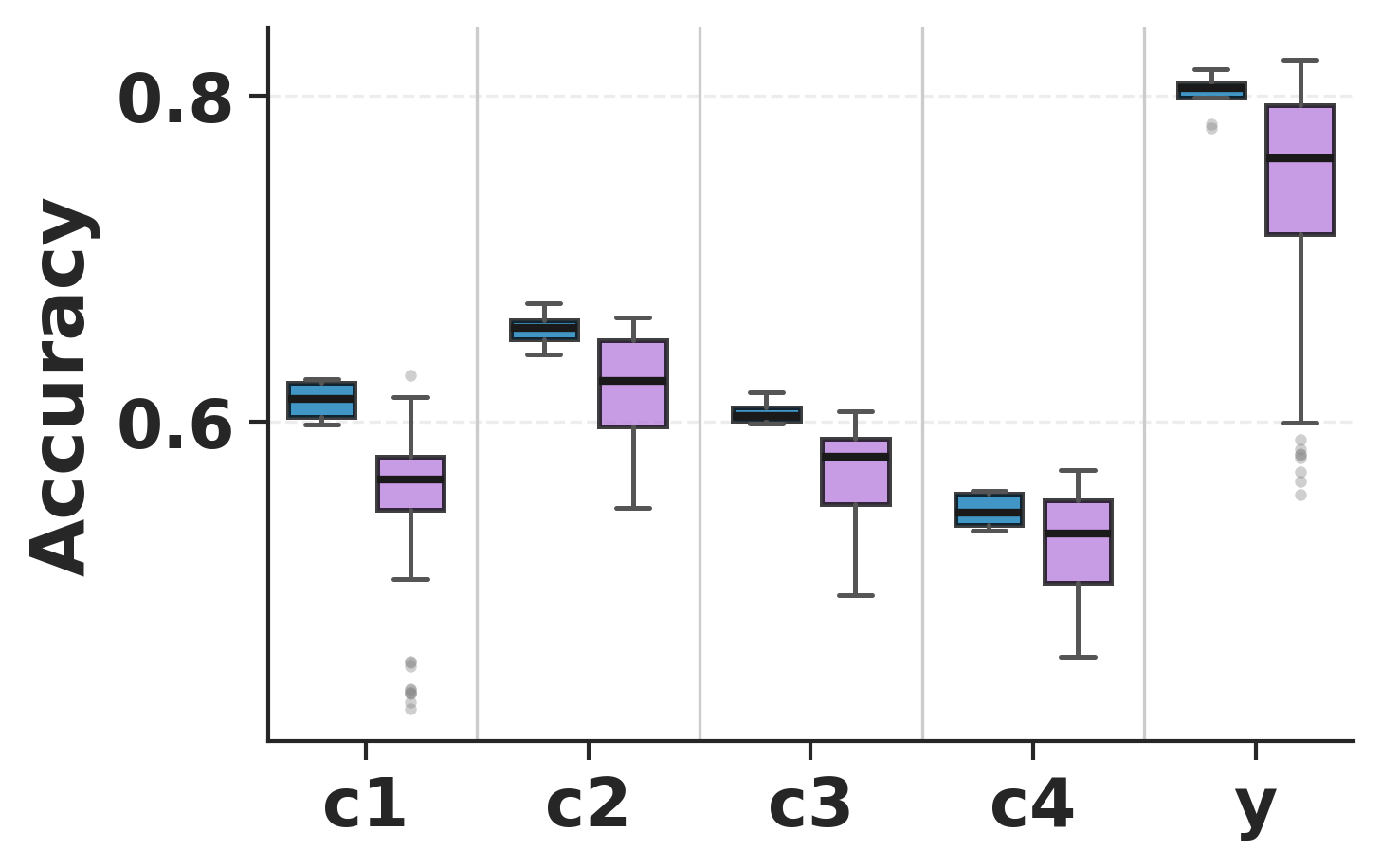}
        \caption{Dataset: SA, Model: QWEN}
    \end{subfigure}\hfill
    \begin{subfigure}{0.32\linewidth}
        \includegraphics[width=\linewidth]{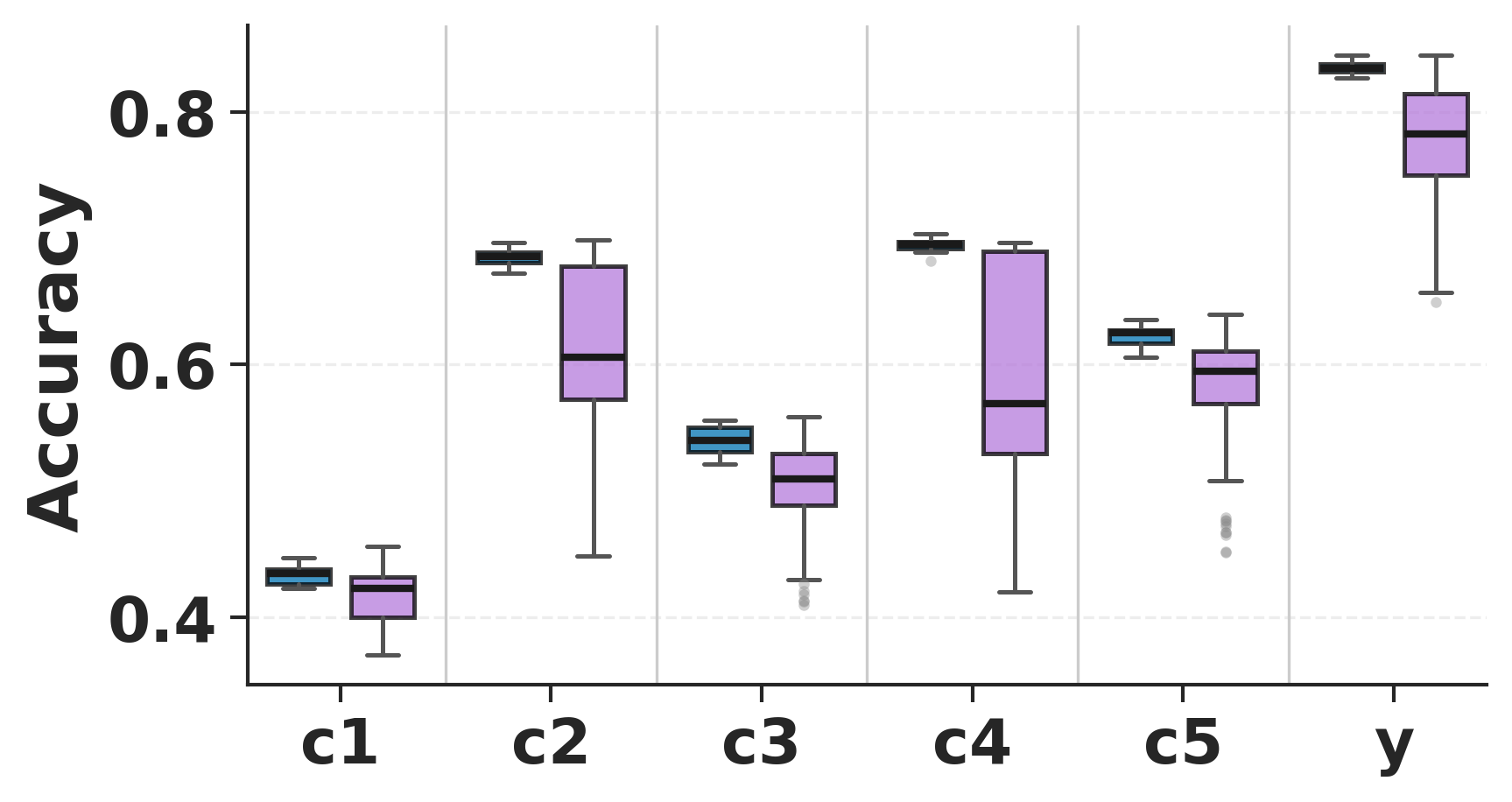}
        \caption{Dataset: SA, Model: GPT-OSS}
    \end{subfigure}

    \vspace{0.3cm} %

    \begin{subfigure}{0.32\linewidth}
        \includegraphics[width=\linewidth]{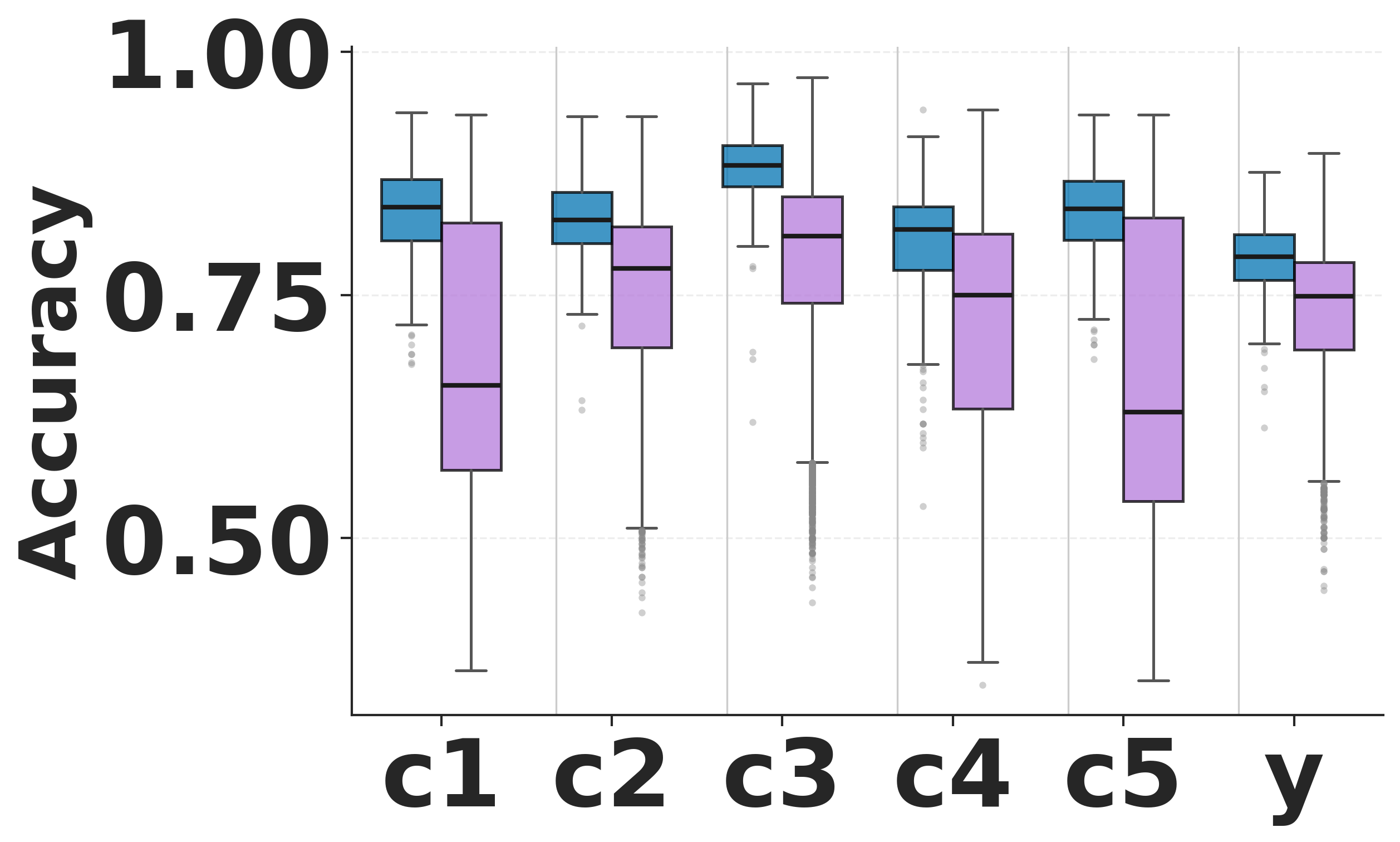}
        \caption{Dataset: LAJ, Model: Gemini}
    \end{subfigure}\hfill
    \begin{subfigure}{0.32\linewidth}
        \includegraphics[width=\linewidth]{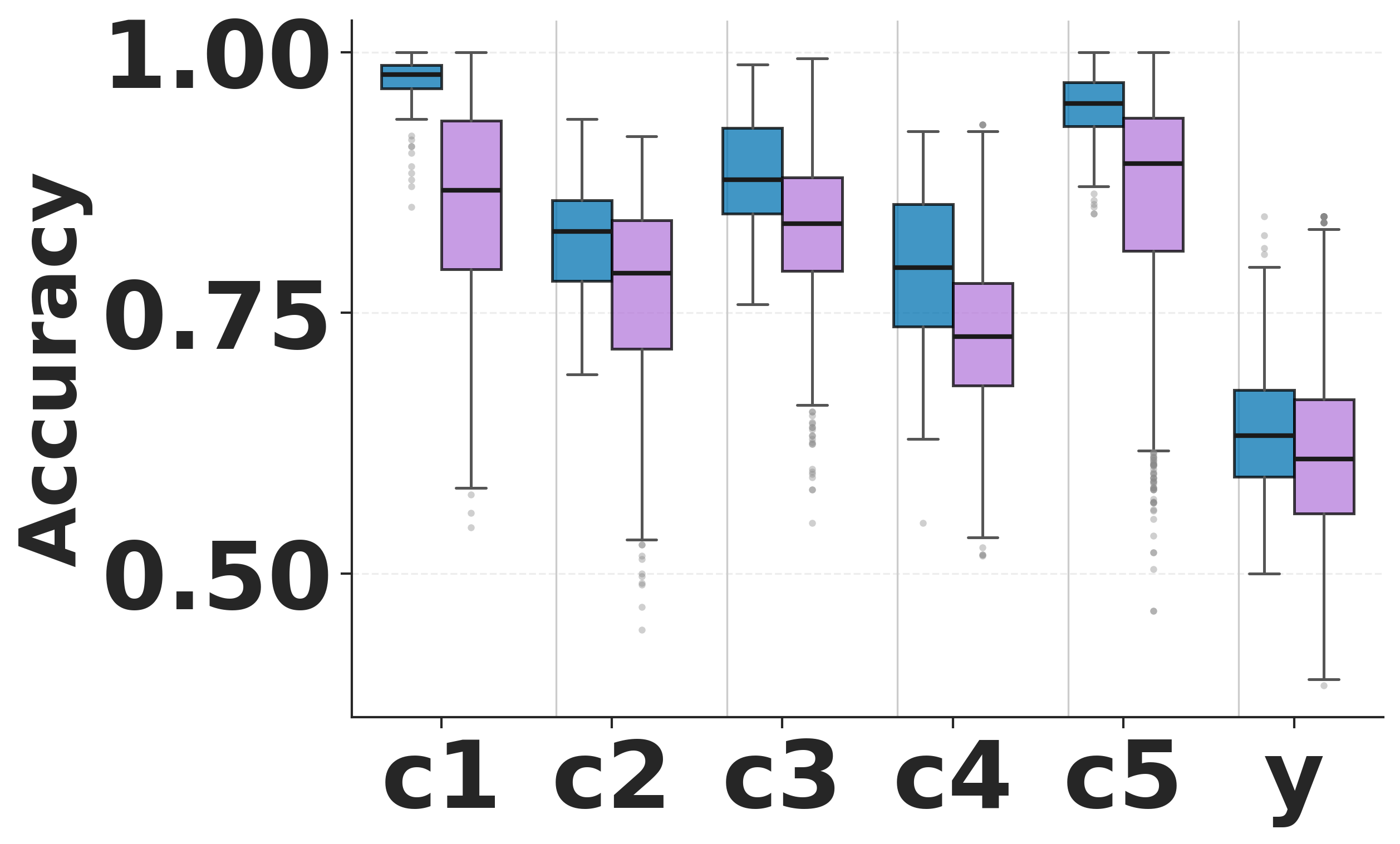}
        \caption{Dataset: LAJ, Model: QWEN}
    \end{subfigure}\hfill
    \begin{subfigure}{0.32\linewidth}
        \includegraphics[width=\linewidth]{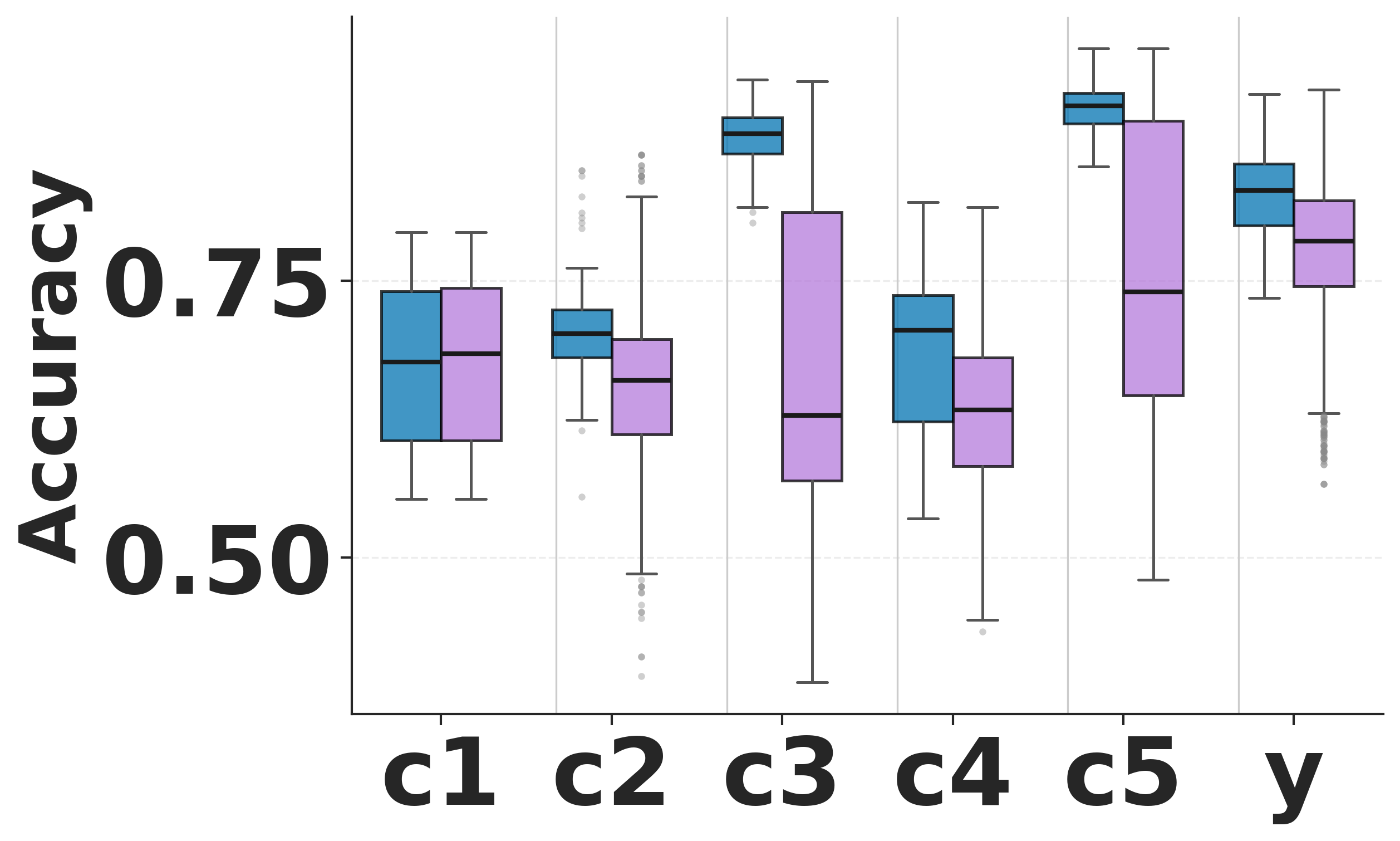}
        \caption{Dataset: LAJ, Model: GPT-OSS}
    \end{subfigure}

    \vspace{0.3cm} 

    \begin{subfigure}{0.32\linewidth}
        \includegraphics[width=\linewidth]{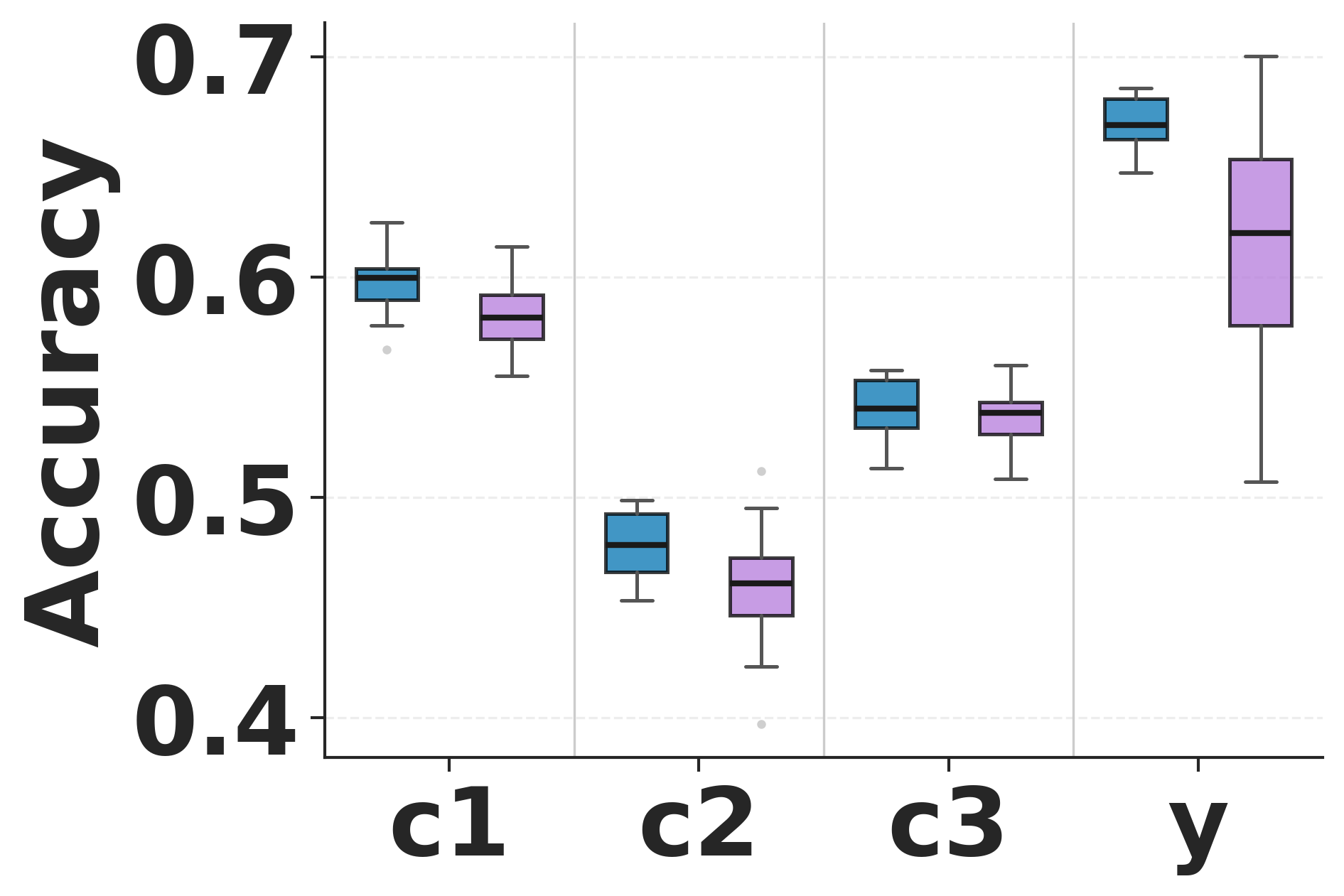}
        \caption{Dataset: DD, Model: Gemini}
    \end{subfigure}\hfill
    \begin{subfigure}{0.32\linewidth}
        \includegraphics[width=\linewidth]{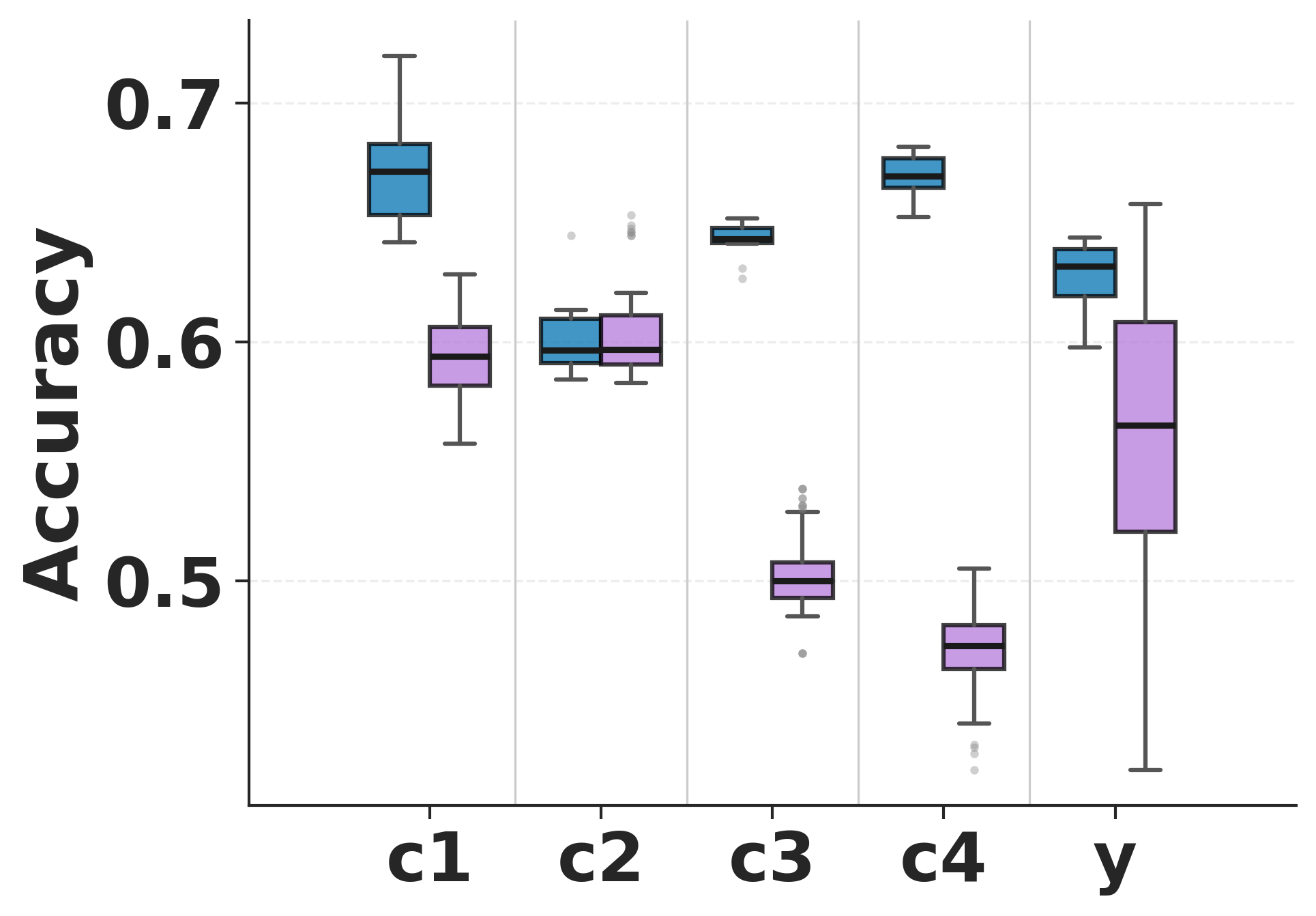}
        \caption{Dataset: DD, Model: QWEN}
    \end{subfigure}\hfill
    \begin{subfigure}{0.32\linewidth}
        \includegraphics[width=\linewidth]{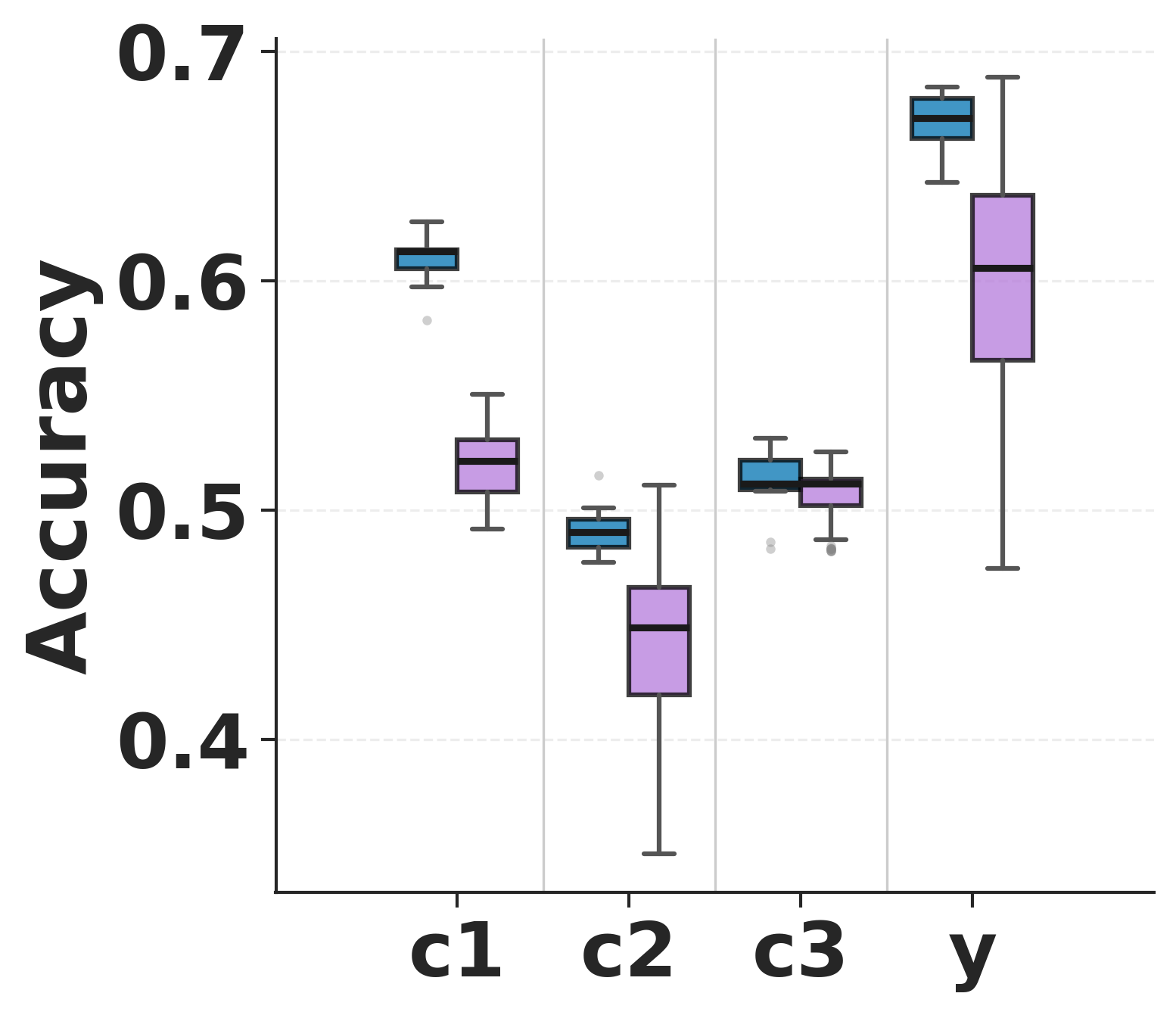}
        \caption{Dataset: DD, Model: GPT-OSS}
    \end{subfigure}

    \caption{Box plots detailing the classification accuracy across 10 rounds of cross-validation for each model and dataset combination. To optimize space, target variables are abbreviated as $c_i$ for intermediate latent concepts and $y$ for the final downstream classification task. The light blue boxes represent the predictive accuracy when conditioning exclusively on the graph-specified causal parents ($Pa(v)$) derived from the representative graph topology. Conversely, the purple boxes display the accuracy distributions of all alternative predictor subsets $Z$ that do not fully encapsulate the required parent set ($Pa(v) \not\subseteq Z$). Each data point in the underlying distribution represents a bootstrap accuracy score. Within each box plot, the central thick line denotes the median accuracy, while the bottom and top edges correspond to the 25th (Q1) and 75th (Q3) percentiles, respectively (representing the Interquartile Range, IQR, or the middle 50\% of the data). The whiskers extend to the furthest data points within 1.5 $\times$ IQR from the box edges. Individual grey markers denote outlier accuracy scores }
\label{fig:validation_boxplots}
    \label{fig:appendix_val_train}
\end{figure*}

To evaluate the predictive necessity and fidelity of the discovered causal structures, we implement an internal validation protocol strictly within the MCMC expended training set. 

For each benchmark, we perform 10 rounds of cross-validation on the dataset. In each validation cycle, the data is split into an 80\% internal training subset and a 20\% held-out validation slice. On the 80\% training subset, the $\sigma$-CG algorithm is executed to construct the representative causal graph. For each target node $v \in V$ (where $V = \mathcal{C} \cup \{\hat{y}\}$), once the topology is established on the 80\% split, we parameterize the functional mechanisms governing the SCM via multinomial logistic regression using its causal parent set, denoted as $Pa(v)$. After parameterization, the predictive capability of the isolated parent sets is evaluated on the remaining 20\% held-out validation set. To confirm that the discovered parents $Pa(v)$ constitute the optimal predictive features for each respective node, their classification accuracy is systematically benchmarked against alternative predictor combinations $S \subseteq V \setminus \{v\}$. To ensure a rigorous and fair comparison, we independently train a distinct logistic regression model for each alternative subset $Z$ using the same 80\% training split, enforcing the constraint that the full parent set is never encapsulated within the baseline ($Pa(v) \not\subseteq Z$).

Our empirical evaluation demonstrates that across the vast majority of datasets and model architectures, the predictors trained exclusively on the causal parent sets achieve the highest classification accuracy. In a few isolated cases, alternative feature combinations achieve parity (ties) with the causal parents, but never outperform them ( see figure ~\ref{fig:appendix_val_train} in appendix ~\ref{fig:appendix_val_train}. 

\paragraph{MCMC Importance.}
\label{appendix:par:app_mcmc_importance}
\begin{figure*}[t]
    \centering

    \begin{subfigure}{0.32\linewidth}
        \includegraphics[width=\linewidth]{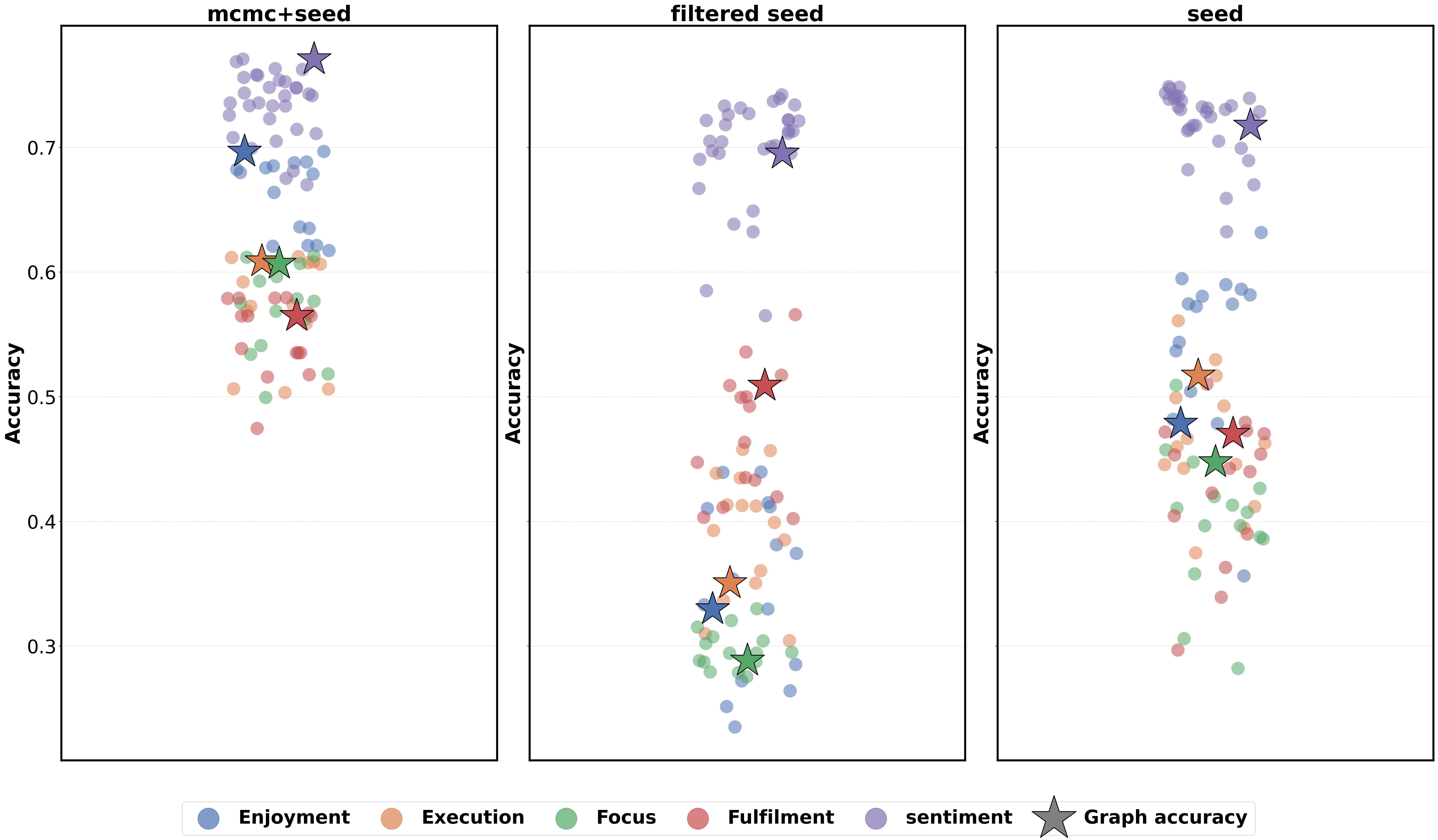}
        \caption{Dataset: IMDB, Model: Gemini}
    \end{subfigure}\hfill
    \begin{subfigure}{0.32\linewidth}
        \includegraphics[width=\linewidth]{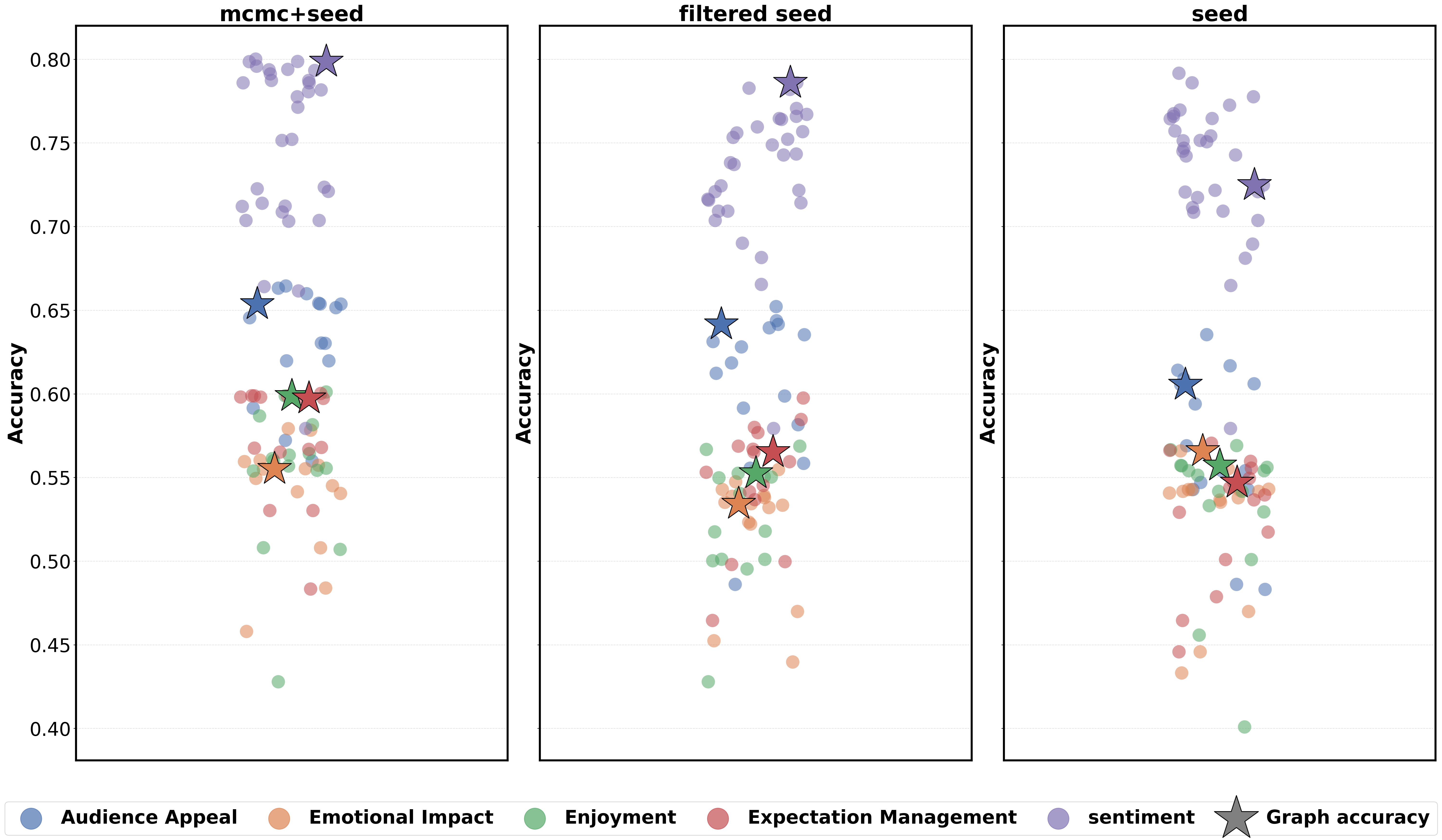}
        \caption{Dataset: IMDB, Model: QWEN}
    \end{subfigure}\hfill
    \begin{subfigure}{0.32\linewidth}
        \includegraphics[width=\linewidth]{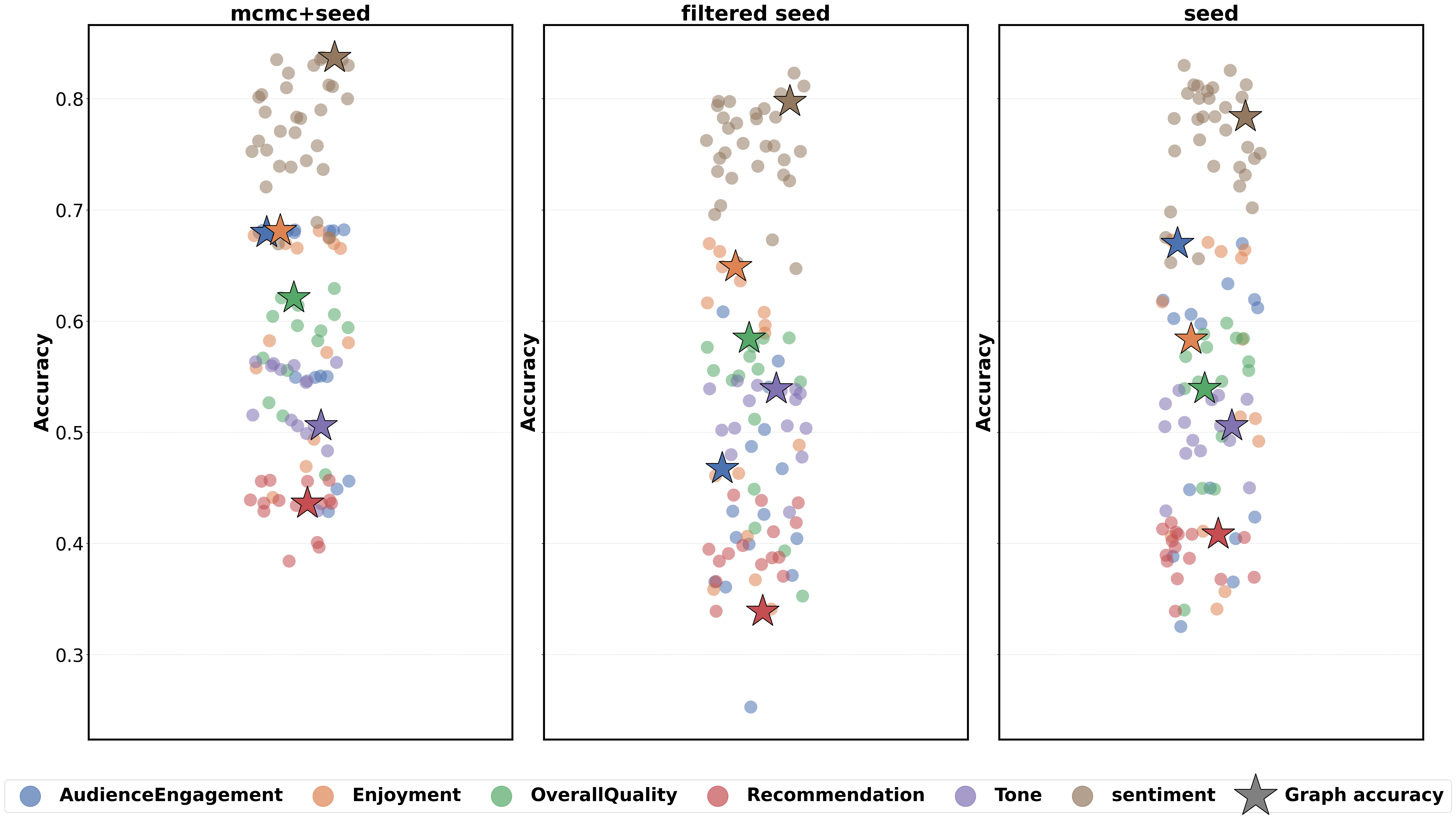}
        \caption{Dataset: IMDB, Model: GPT-OSS}
    \end{subfigure}

    \vspace{0.3cm}

    \begin{subfigure}{0.32\linewidth}
        \includegraphics[width=\linewidth]{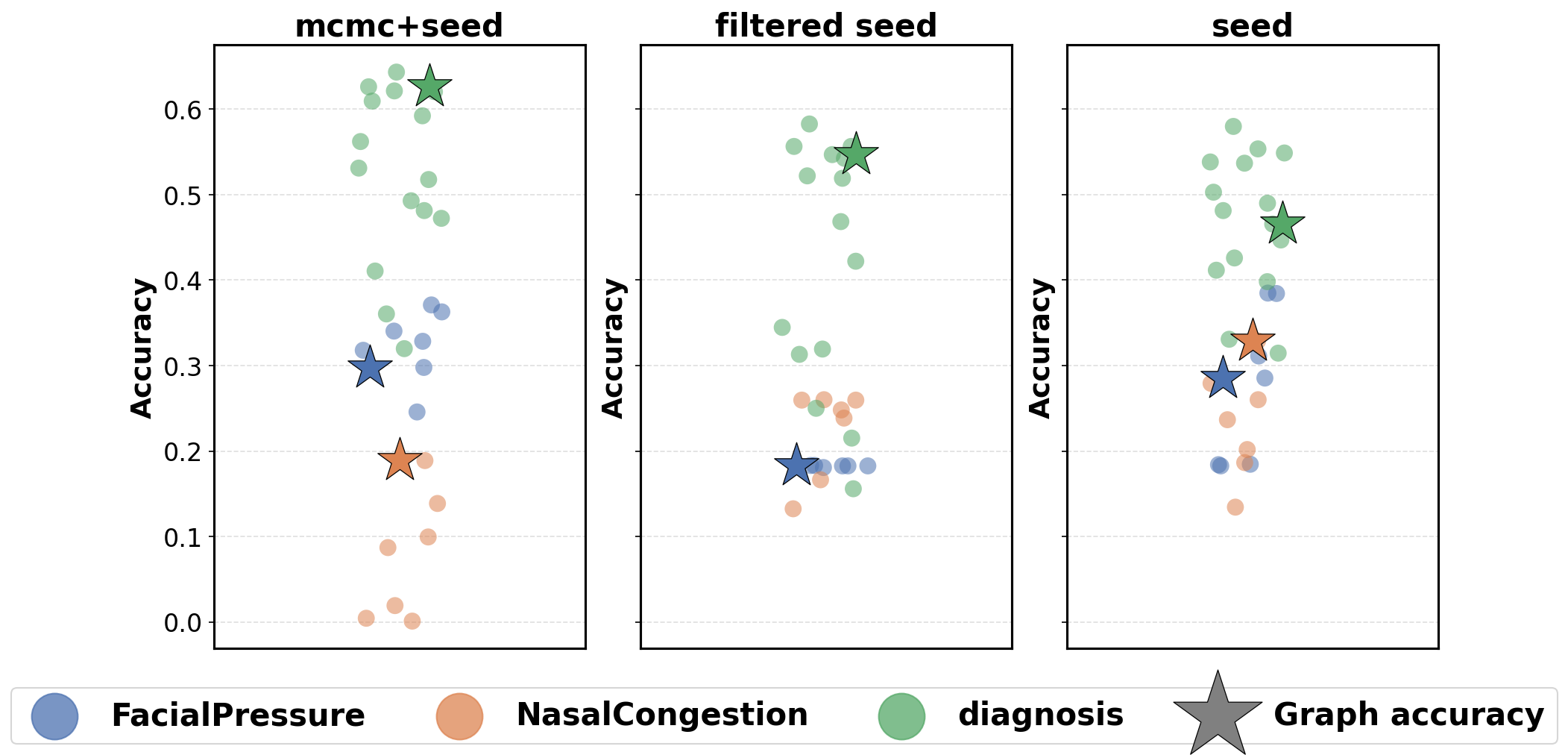}
        \caption{Dataset: LIBERTY, Model: Gemini}
    \end{subfigure}\hfill
    \begin{subfigure}{0.32\linewidth}
        \includegraphics[width=\linewidth]{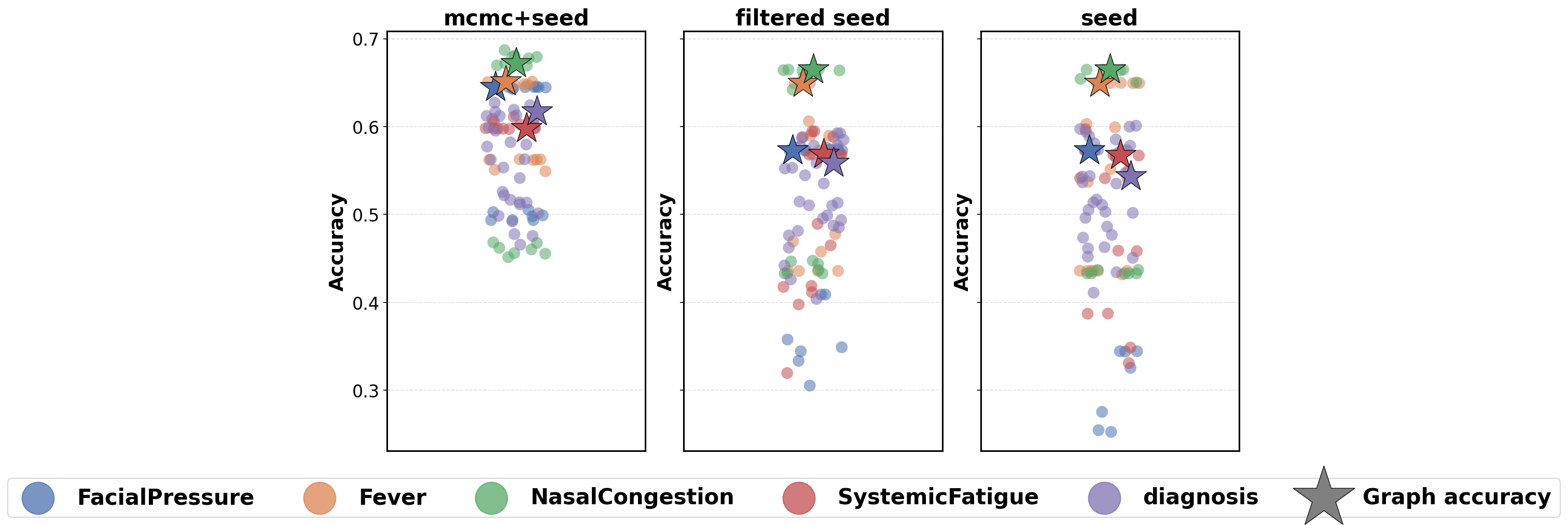}
        \caption{Dataset: LIBERTY, Model: QWEN}
    \end{subfigure}\hfill
    \begin{subfigure}{0.32\linewidth}
        \includegraphics[width=\linewidth]{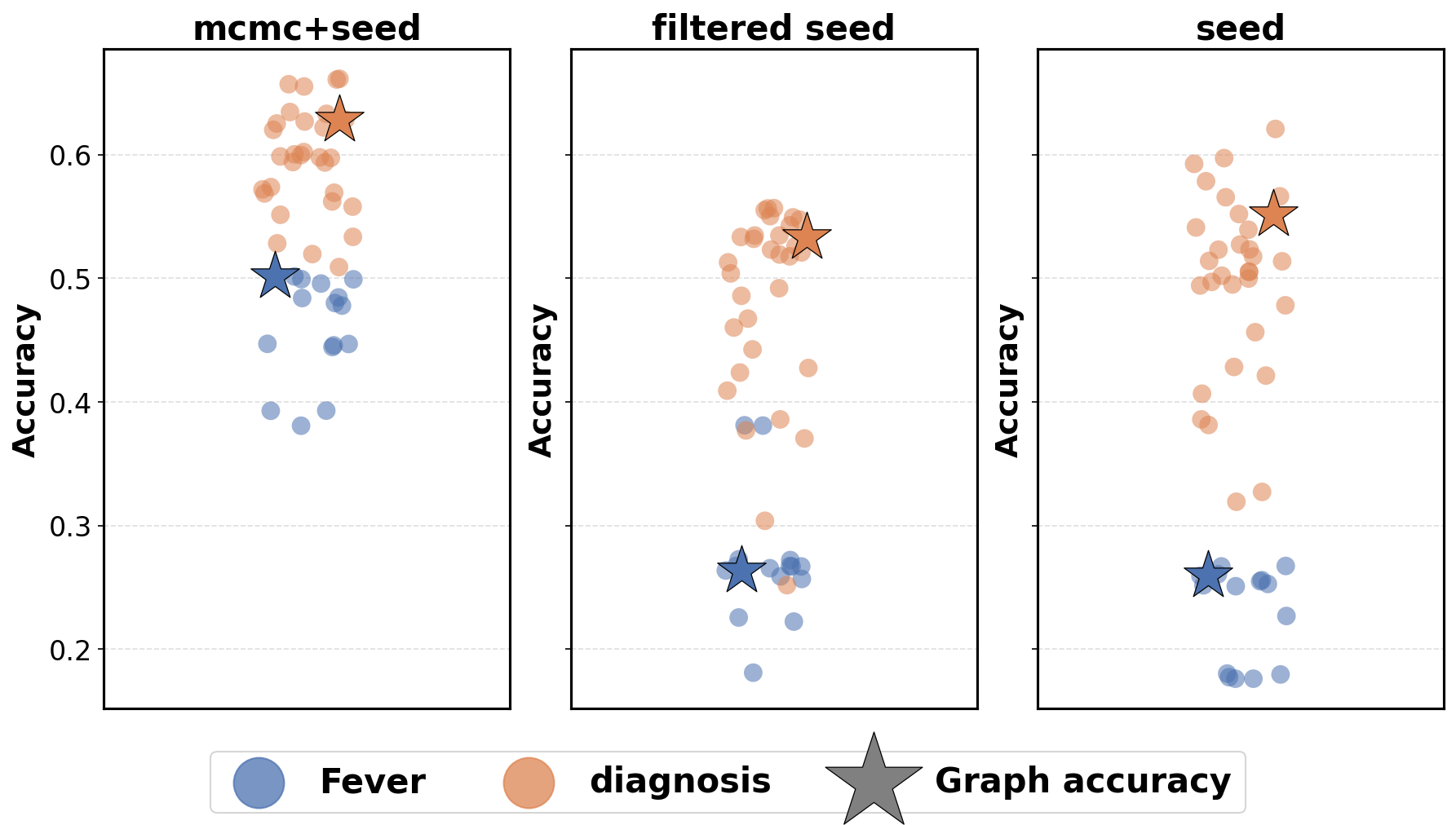}
        \caption{Dataset: LIBERTY, Model: GPT-OSS}
    \end{subfigure}

    \caption{Concept predictability accuracy comparison across training set variations (`mcmc+seed`, `filtered seed`, and `seed`), evaluated on the MCMC-expanded test set. Colors denote specific target concepts as defined in the legend Semi-transparent, colored scatter points represent accuracies across the full combinatorial distribution of evaluated parent-child configurations for each concept. Large, colored star markers explicitly highlight the optimal accuracy specified by the learned causal graph topology; the color of each star matches the concept it evaluates .}
\label{fig:concept_predictability_scatter}
    \label{fig:appendix_val_test}
\end{figure*}
To systematically validate the effectiveness of our MCMC-based data expansion process, we evaluate its impact on the downstream causal discovery across all presented datasets. Specifically, for each individual dataset, we execute the causal discovery process to extract and compare three distinct causal graphs, corresponding to three different training configurations. The first is the \emph{Seed} configuration, which serves as our baseline graph constructed utilizing the full original training set $\mathcal{D}_{\text{train}}$. The second is the \emph{MCMC+Filtered Seed} configuration; this represents our primary graph, extracted by combining the expanded dataset $\mathcal{D}_{\text{mcmc}}$ specifically with the subset of seed instances that successfully initiated valid counterfactual chains. The third is the \emph{Filtered Seed} configuration, which serves as a control baseline graph learned using only those exact same filtered seed instances, but without the addition of the MCMC-generated counterfactuals.

A key challenge in evaluating causal graphs over sparse observational data is that the initial evaluation set rarely spans the combinatorial concept space adequately. Evaluating model performance on a non-spanning test set can yield skewed or unrepresentative structural metrics. To address this limitation, we leverage our MCMC algorithm to expand the test set as well, forcing a dense coverage of alternative concept configurations. Due to computational constraints and design requirements specifically because the LAJ dataset constructs an isolated graph per unique query this test-set expansion procedure was applied exclusively to the LIBERTY and IMDB datasets.

Following the construction of these three graph configurations, we formalize their structural equations by estimating the potential functions via multiclass logistic regression. Specifically, for each target variable $v \in V$ (where $V = \mathcal{C} \cup \{\hat{y}\}$), we fit a separate logistic regression model to predict its state based exclusively on its set of direct parents in the recovered graph, denoted as $Pa(v)$. To comprehensively evaluate the predictive optimality of this causal structure, we independently train a distinct logistic regression model for every possible alternative subset of predictor concepts $Z \subseteq V \setminus \{v\}$. We then evaluate all learned models on the expanded test set, computing and comparing the predictability accuracy of each variable across the entire combinatorial space. Within the evaluation figures, the performance of the explicitly learned $Pa(v)$ configuration is highlighted with a star symbol.

The empirical results demonstrate the overall value of the MCMC expansion phase (see Figure~\ref{fig:appendix_val_test}). While the parent-conditioned accuracy derived from the \emph{MCMC+Filtered Seed} graph consistently outperforms both baseline configurations (\emph{Seed} and \emph{Filtered Seed}) for the majority of concepts, we do observe a few isolated exceptions across the evaluated settings. The most prominent deviation occurs with the Gemini model evaluated on the LIBERTY dataset, where the structural accuracy for certain intermediate concepts exhibits suboptimal performance under the expanded regime. Importantly, this degradation remains strictly localized; for the primary classification target (i.e., the \emph{Diagnosis} concept), the \emph{MCMC+Filtered Seed} regime actually improves predictive accuracy compared to the two baseline training configurations. While minor variations exist across nodes, the specifically learned graph-based parent configuration consistently achieves top-tier accuracy within each training regimes, decisively outperforming the vast majority of alternative concept combinations. Ultimately, in the majority of instances, the "center of mass" of the predictive accuracy distribution across the combinatorial space exhibits a clear upward shift under the \emph{MCMC+Filtered Seed} regime. This robust general tendency confirms that the MCMC algorithm successfully expands the observational data distribution while preserving the underlying structural causal signal.

\subsection{MCMC Convergence and Stability}
\label{appendix:subsec:app_mcmc_convergence}

\paragraph{Limitations of Classic Diagnostics.}
Classic MCMC convergence metrics, such as the Gelman-Rubin statistic \cite{10.1214/ss/1177011136}, require knowing the target probability distribution and computing variances across multiple independent chains. Our setting lacks both. First, the underlying probability distribution of our latent concepts is unknown. Second, although classic methods also initialize from multiple starting points, they require running several independent chains from each point to measure variance. In contrast, we run only a single chain per text example. Consequently, we cannot compute traditional variance-based diagnostics, necessitating our tailored KL-based convergence metric.

\paragraph{State Space and Probability Vector.}
To measure convergence, we define the state space of our extracted concepts. Given a set of concepts \(\mathcal{C}\), where each concept can take one of \(m\) possible labels, the total number of unique concept combinations is \(m^{|\mathcal{C}|}\). We flatten this combinatorial space into a single global probability vector of size \(m^{|\mathcal{C}|}\). Each index corresponds to a specific combination of concept labels, and the value at that index represents its empirical probability in our generated data. After each MCMC iteration, we update this vector and calculate the KL divergence from the previous state (see example in ~\ref{appendix:example:Causal Discovery}).

\paragraph{Theoretical Bounding Scenarios.}
Because the global probability vector is updated cumulatively, the relative impact of each new iteration naturally decreases, mathematically forcing the KL divergence to shrink over time. To determine whether the drop in KL divergence indicates genuine convergence rather than this structural artifact, we established two theoretical boundaries under the uniform distribution assumption. Let $h$ denote the accumulated number of instances evaluated up to the current stage, and $s$ denote the number of newly added counterfactual instances in the current iteration:
\begin{itemize}
    \item \textbf{Convergence Bound (Perfect Overlap):} Occurs when newly generated samples distribute proportionally across the already populated bins, perfectly mirroring the existing empirical distribution. the new samples act as a representative sub-distribution of the previous step. Since the finite number of samples added per iteration cannot cover the entire target space, this proportional overlap implies the algorithm has mapped the relevant support and stabilized. The closed-form KL divergence for this bound is given by:
    \begin{align}
    \label{eq:appendix:convergance}
    KL_{\text{overlap}} ={}& \frac{2s}{h+s} \log\left(\frac{2h}{h+s}\right) \\ \nonumber
    &+ \frac{h}{h+s} \log\left(\frac{h}{h+s}\right)
    \end{align}
    \item \textbf{Non-Convergence Bound (Orthogonal Expansion):} Occurs when new samples fall into completely empty bins. This demonstrates that the algorithm is still exploring entirely new regions, meaning the underlying distribution has not yet stabilized. To prevent undefined logarithmic evaluations for previously empty bins, we introduce a smoothing constant $\epsilon = 10^{-10}$. The closed-form KL divergence for this expansion is given by:
    \begin{align}
    \label{eq:appendix:non_convergeance}
    KL_{\text{orthogonal}} ={}& \frac{s}{h+s} \log\left(\frac{1}{\epsilon(h+s)}\right) \\ \nonumber
    &+ \frac{h}{h+s} \log\left(\frac{h}{h+s}\right)
    \end{align}
\end{itemize}
By plotting the empirical KL divergence against these closed-form boundary curves, we can accurately determine when the state space exploration has concluded.
\begin{figure*}[t]
    \centering

    \begin{subfigure}{0.32\linewidth}
        \includegraphics[width=\linewidth]{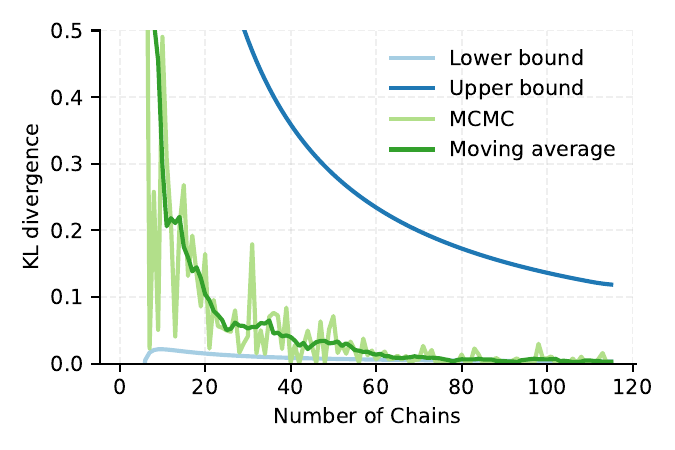}
        \caption{Dataset: IMDB, Model: Gemini}
    \end{subfigure}\hfill
    \begin{subfigure}{0.32\linewidth}
        \includegraphics[width=\linewidth]{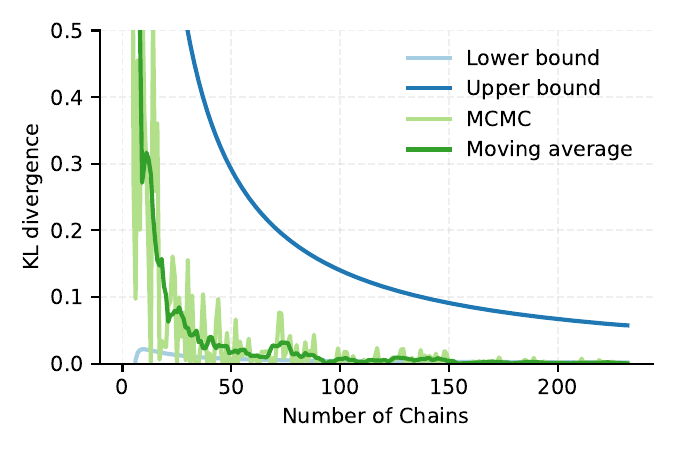}
        \caption{Dataset: IMDB, Model: QWEN}
    \end{subfigure}\hfill
    \begin{subfigure}{0.32\linewidth}
        \includegraphics[width=\linewidth]{images/IMDB/OSS/kl_divergence_four_lines_zoom.pdf}
        \caption{Dataset: IMDB, Model: GPT-OSS}
    \end{subfigure}

    \vspace{0.3cm}

    \begin{subfigure}{0.32\linewidth}
        \includegraphics[width=\linewidth]{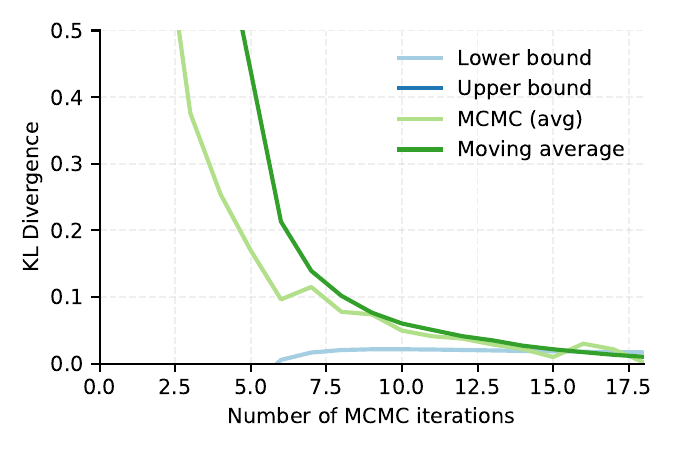}
        \caption{Dataset: LAJ, Model: Gemini}
    \end{subfigure}\hfill
    \begin{subfigure}{0.32\linewidth}
        \includegraphics[width=\linewidth]{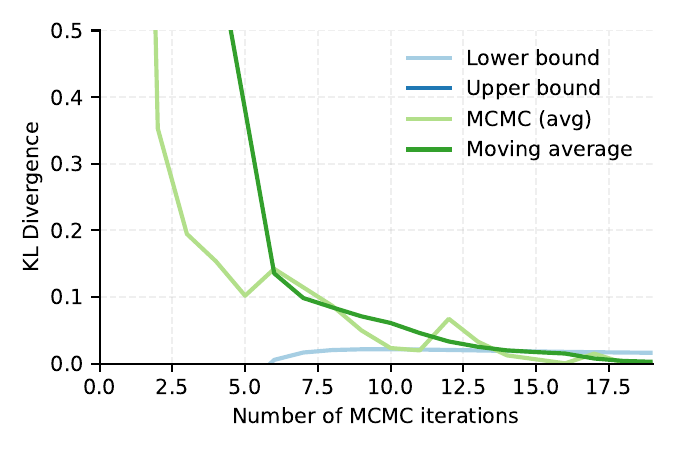}
        \caption{Dataset: LAJ, Model: QWEN}
    \end{subfigure}\hfill
    \begin{subfigure}{0.32\linewidth}
        \includegraphics[width=\linewidth]{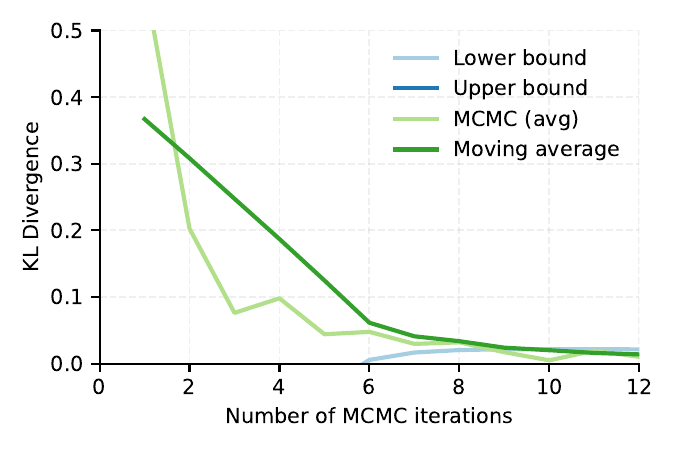}
        \caption{Dataset: LAJ, Model: GPT-OSS}
    \end{subfigure}

    \vspace{0.3cm} 

    \begin{subfigure}{0.32\linewidth}
        \includegraphics[width=\linewidth]{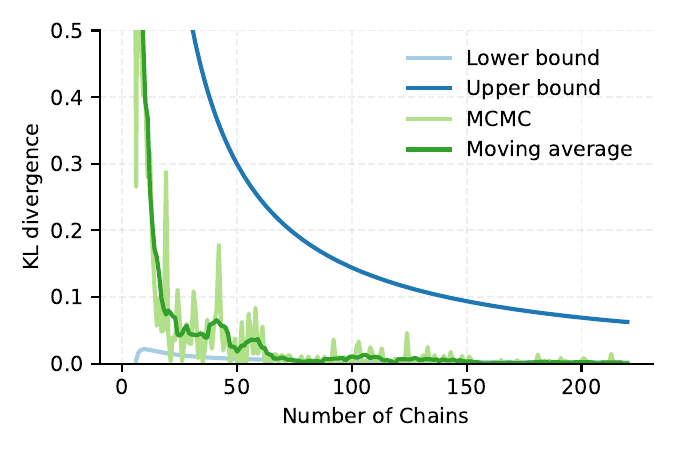}
        \caption{Dataset: LIBERTY, Model: Gemini}
    \end{subfigure}\hfill
    \begin{subfigure}{0.32\linewidth}
        \includegraphics[width=\linewidth]{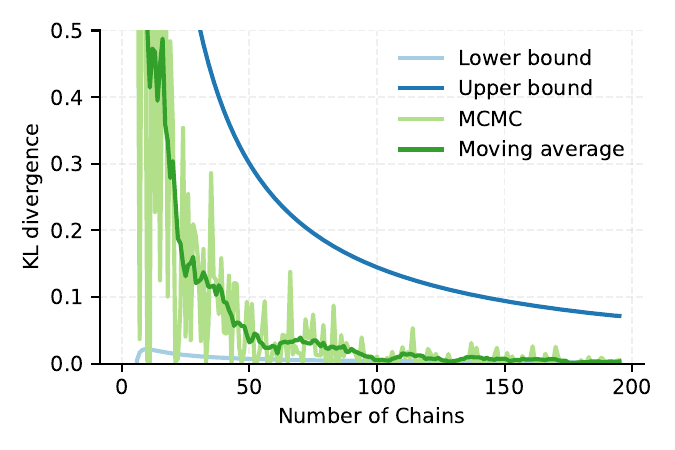}
        \caption{Dataset: LIBERTY, Model: QWEN}
    \end{subfigure}\hfill
    \begin{subfigure}{0.32\linewidth}
        \includegraphics[width=\linewidth]{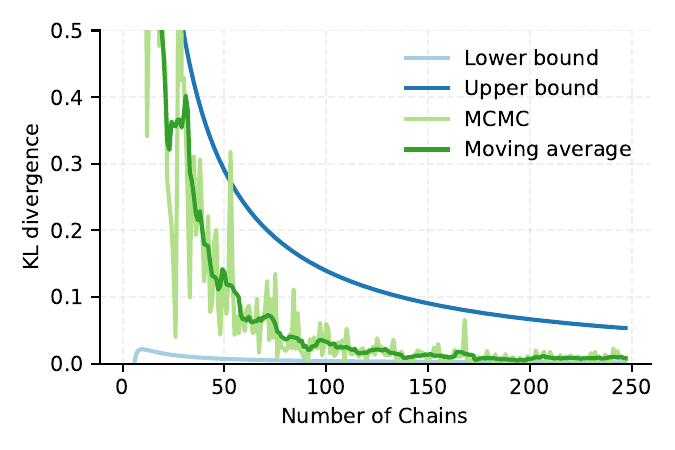}
        \caption{Dataset: LIBERTY, Model: GPT-OSS}
    \end{subfigure}

    \caption{The plots illustrates the convergence of our MCMC data expansion algorithm. The x-axis represents the number of MCMC iterations, where each iteration corresponds to the full expansion of a single original data instance into a new batch of samples in the IMDB and LIBERTY dataset, and the expension of on example form previous naive expension step for LAJ dataset. The y-axis denotes the Kullback-Leibler (KL) divergence between the updated cumulative probability distribution and the distribution from the previous iteration. To contextualize the empirical convergence against the natural mathematical artifact of accumulating samples, we plot two theoretical boundaries. The blue curves indicate the converged-case boundary of perfect overlap, where newly generated samples perfectly align with the previously established distribution. Conversely, the red curves signify the non converged-case boundary of orthogonal expansion, representing a scenario where each iteration introduces completely novel samples in previously unexplored regions of the concept space. The green curves represent the empirical KL divergence calculated from our generated data. Notably, the expansion and convergence processes differ across datasets. For the LIBERTY and IMDB datasets, the process is evaluated globally across all data instances, yielding a single empirical green curve per plot. In contrast, for the LAJ dataset, we construct an independent causal graph for each distinct query. Consequently, the LAJ plots display multiple green curves, each reflecting the isolated convergence of a single data item. Furthermore, because each LAJ process is restricted to the context of a single query, its latent concept distribution space is inherently smaller, which accounts for the noticeably faster convergence rate observed in these specific plots. True convergence is achieved as the empirical green curves stabilize and approach the blue boundary.}
    \label{fig:appendix kl_conv}
\end{figure*}

\paragraph{Convergence Result}
In this section, we provide the complete empirical evaluation of our MCMC convergence diagnostics across all combinations of target Large Language Models and datasets, extending the representative examples illustrated in the main text. As detailed in Section~\ref{subsec:stablitiy}, our framework tracks the empirical probability distribution over the combinatorial concept space $\mathcal{C}$ across iterations, evaluating the Kullback-Leibler (KL) divergence against theoretical best-case (perfect overlap) and worst-case (orthogonal exploration) boundaries. 

Across all examined models and datasets, the empirical KL divergence curves consistently demonstrate robust and stable convergence ( see figure ~\ref{fig:appendix kl_conv} in appendix), systematically shifting away from the upper exploration bound and adhering closely to the lower optimal boundary. This universal trend confirms that the proposed MCMC data expansion successfully convergence prior to the termination of the chains.

However, a distinct structural variation emerges when comparing the architectural paradigms of the datasets. For the IMDB and LIBERTY benchmarks, the global probability distribution is computed across a singular, unified concept space mapped over the entire dataset, resulting in a single empirical convergence curve per model. Conversely, the LAJ framework operates under a distinct constraint, constructing a localized causal graph tailored to each individual query. 

Consequently, the combinatorial state space for any single query in LAJ is inherently more constrained and tightly bounded than the global space evaluated in the other datasets. This severe reduction in space complexity manifests empirically as an accelerated rate of convergence, where the chains stabilize significantly faster. 

This architectural difference directly dictates the visual representation of the convergence plots. While the IMDB and LIBERTY datasets yield a single empirical convergence curve corresponding to their singular global causal graph, the LAJ dataset evaluation displays multiple empirical curves (depicted as individual green trajectories) within the same plot. Each green trajectory corresponds to the MCMC convergence process of a distinct, query-specific causal graph. These visualizations confirm that despite the independent nature of each query's causal mechanism, every isolated chain strictly satisfies our theoretical convergence criteria.

\begin{figure*}[t]
    \centering

    \begin{subfigure}{0.32\linewidth}
        \includegraphics[width=\linewidth]{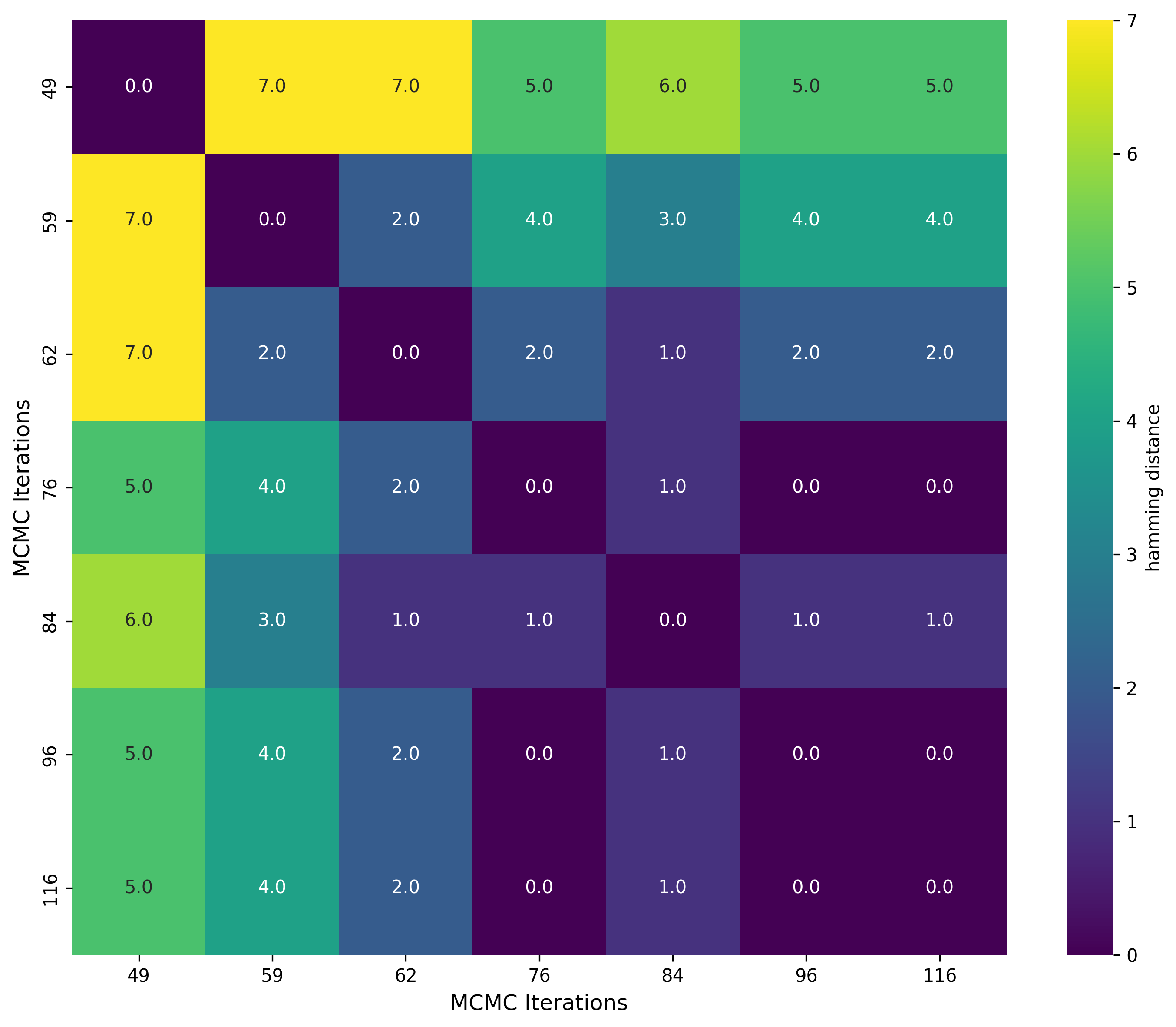}
        \caption{Dataset: IMDB, Model: Gemini}
    \end{subfigure}\hfill
    \begin{subfigure}{0.32\linewidth}
        \includegraphics[width=\linewidth]{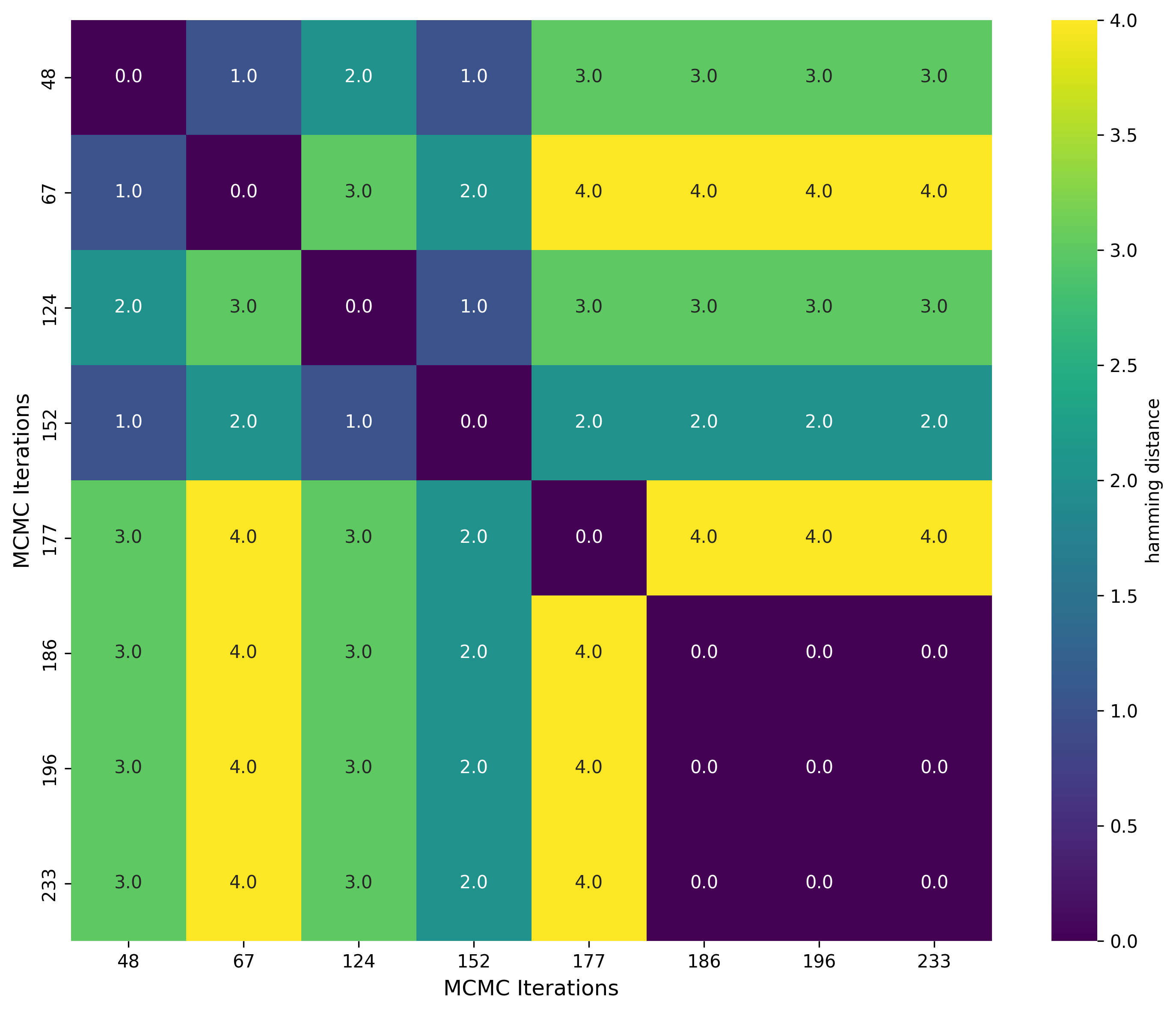}
        \caption{Dataset: IMDB, Model: QWEN}
    \end{subfigure}\hfill
    \begin{subfigure}{0.32\linewidth}
        \includegraphics[width=\linewidth]{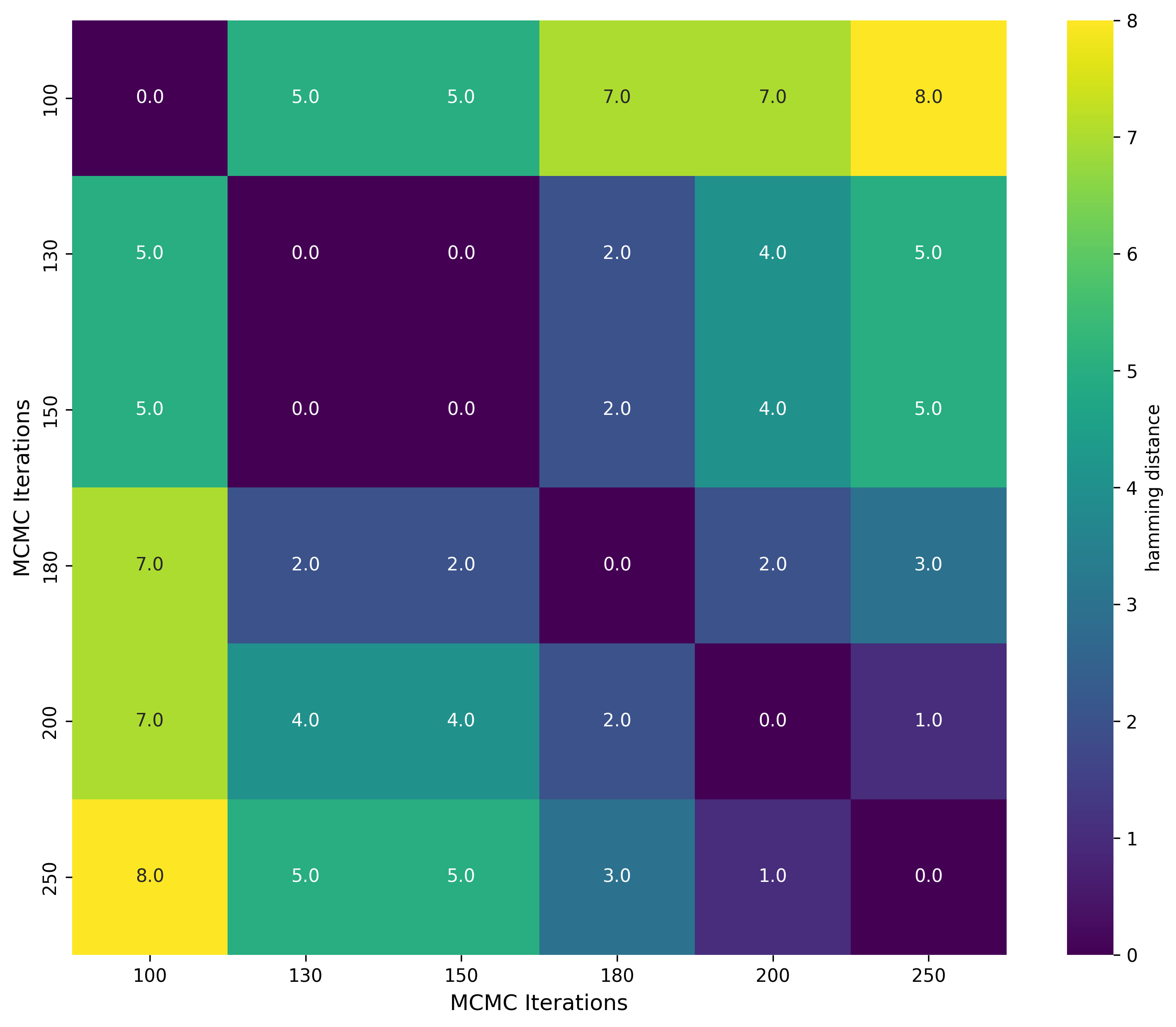}
        \caption{Dataset: IMDB, Model: GPT-OSS}
    \end{subfigure}

    \vspace{0.3cm} 

    \begin{subfigure}{0.32\linewidth}
        \includegraphics[width=\linewidth]{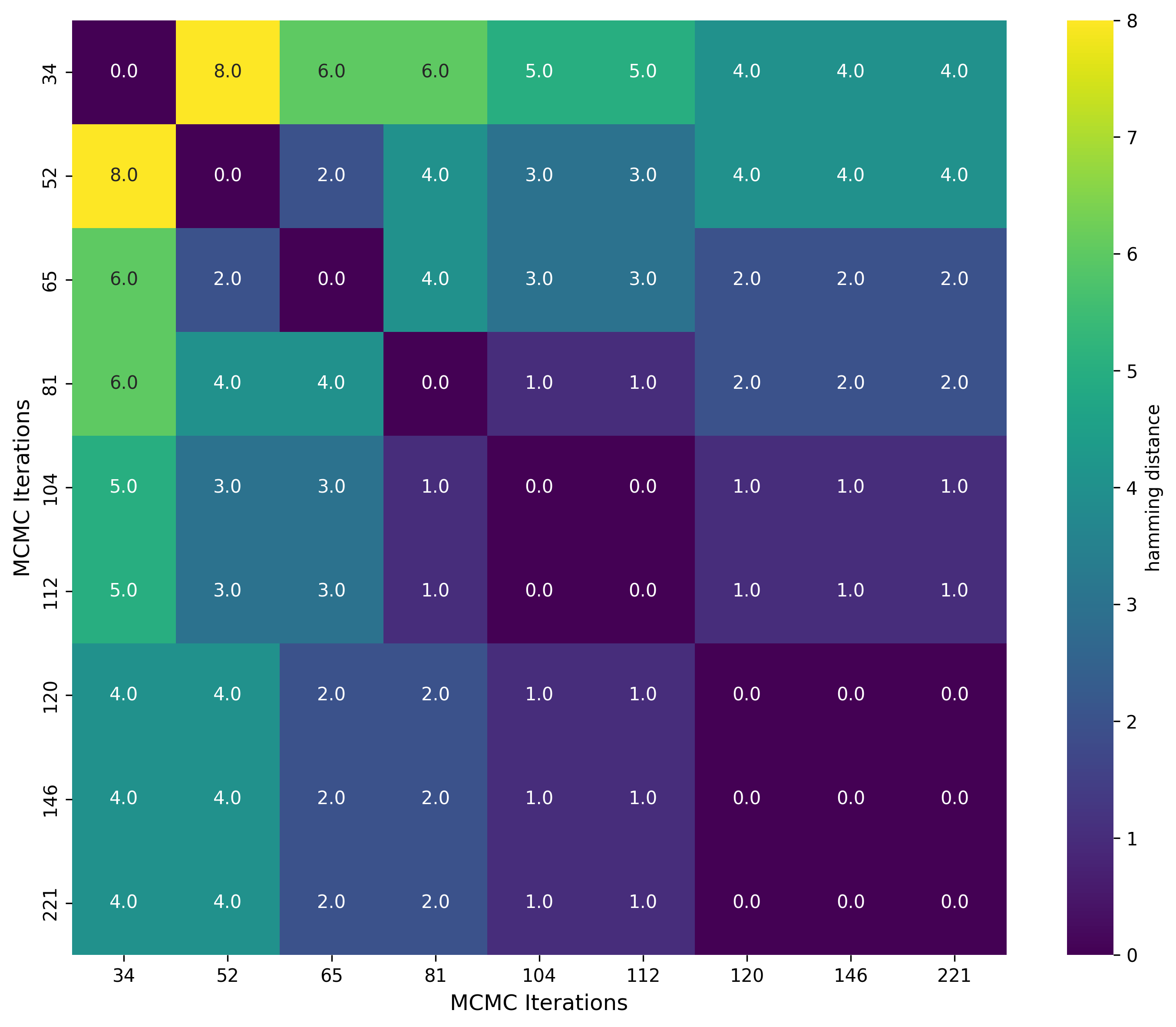}
        \caption{Dataset: LIBERTY, Model: Gemini}
    \end{subfigure}\hfill
    \begin{subfigure}{0.32\linewidth}
        \includegraphics[width=\linewidth]{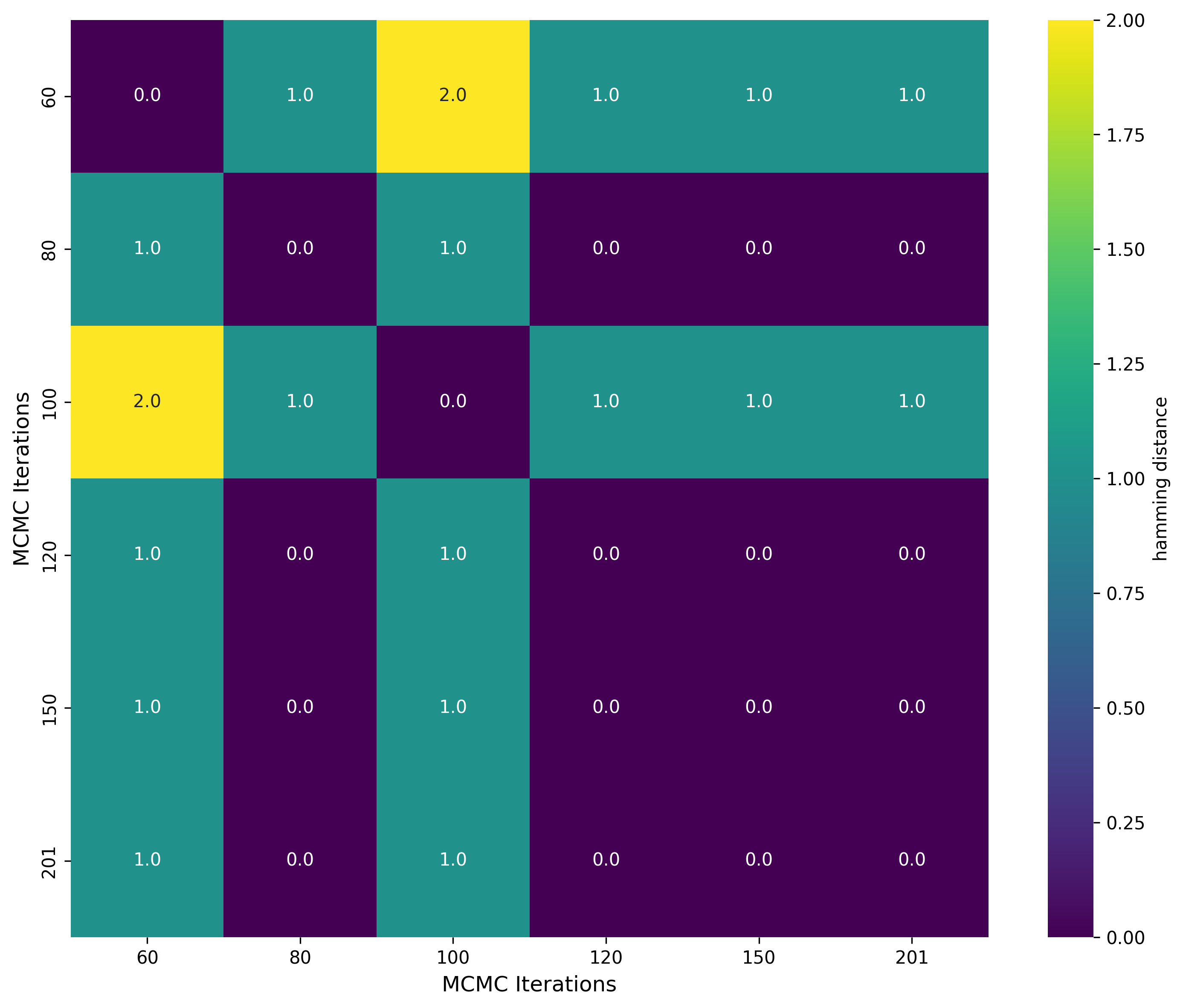}
        \caption{Dataset: LIBERTY, Model: QWEN}
    \end{subfigure}\hfill
    \begin{subfigure}{0.32\linewidth}
        \includegraphics[width=\linewidth]{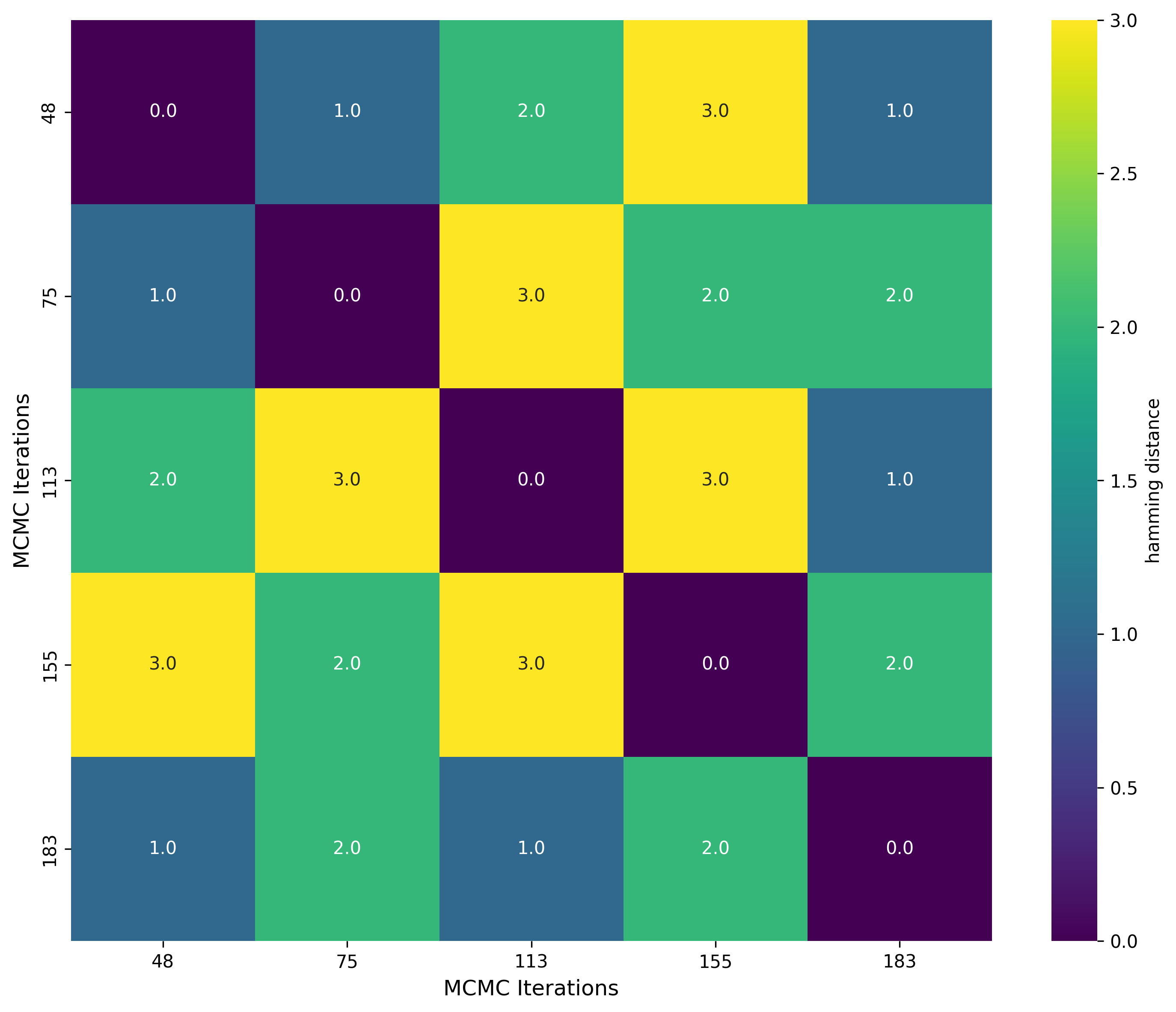}
        \caption{Dataset: LIBERTY, Model: GPT-OSS}
    \end{subfigure}

    \caption{Structural consistency analysis of the learned causal models across different stages of the MCMC expansion. The figure presents six subplots, each displaying a distance matrix that compares the causal graphs generated after varying numbers of MCMC iterations. The values within the matrices represent the Hamming distance, which quantifies the exact number of differing edges between any two given graphs. This evaluation serves as a supplementary empirical validation for our MCMC convergence criterion. As the number of MCMC iterations increases, the Hamming distances between subsequently generated graphs diminish and eventually stabilize. This indicates that the topology of the causal graph achieves structural stability and ceases to change, thereby confirming that the MCMC data expansion process has successfully converged.}
    \label{fig:appendix:SHD}
\end{figure*}

\paragraph{Stability Results}

To provide empirical validation for the convergence of our MCMC data expansion algorithm, we conducted a structural consistency check across progressive stages of the expansion process. Since the LAJ dataset yields an independent causal graph for each individual query, this analysis is performed exclusively on the IMDB and LIBERTY benchmarks. For each dataset, we learn the underlying causal graphs using an increasing number of MCMC iterations, corresponding to a growing volume of expanded counterfactual instances. Specifically, at each evaluated milestone of MCMC iterations, we implement a 10-run cross-validation protocol; in each run, the causal graph is independently learned using a random 80\% subset of the data generated up to that stage. The final consensus graph topology for that specific milestone is then determined via a majority vote, such that an edge is included in the final graph if and only if it appeared in 5 or more of the 10 generated graphs. We then quantify the structural distance between the resulting consensus graphs across different expansion stages using the Structural Hamming Distance (SHD), which measures the exact number of edge discrepancies between two graphs (where a distance of 0 denotes topological identity). Our empirical results reveal nuanced convergence patterns across the evaluated models: for Gemini and Qwen, the SHD between subsequent graph configurations consistently diminishes and eventually reaches exactly zero, demonstrating absolute structural invariance. For GPT-OSS, although the structural distances significantly decrease, they stabilize at a low, non-zero baseline rather than completely vanishing (see Figure ~\ref{fig:appendix:SHD} in the appendix).

\section{Running Example: Full Pipeline Walkthrough}
\label{appendix:sec:app_running_example}
To provide a concrete understanding of our methodology, this section walks through the four-phase pipeline using a synthetic toy task: classifying the tastiness of a papaya from a short textual description. Let the set of task classes be $\mathcal{Y}=\{\text{\emph{tasty}},\text{\emph{not-tasty}}\}$. Each concept takes values in $\mathcal{V}=2^{\mathcal{Y}}$: $\emptyset$ means that the concept is absent, $\mathcal{Y}$ means that it is aligned with both classes, and singleton sets indicate class-specific alignment.

\paragraph{Phase 1: Label Prediction.}
\label{appendix:example:Inference and Label Alignment}
Suppose the initial dataset $\mathcal{D}$ contains a seed text $x^{(0)}$: \emph{``Today I ate a bright orange, mushy papaya.''} The ground-truth label in the dataset may be \emph{tasty}, but the target LLM predicts $\hat{y}=f(x^{(0)})=\text{\emph{not-tasty}}$, likely because it associates ``mushy'' with being overripe. We therefore replace the ground-truth label with the LLM prediction, so all downstream stages reflect the model's perspective.

\paragraph{Phase 2: Discriminative Concept Discovery and Representation.}
\label{appendix:example:Differentiative Concept Extraction}
We process $\mathcal{D}_{\text{train}}$ in small, class-balanced batches. Given texts and the predicted labels from Phase~1, the LLM proposes candidate concepts that distinguish between the predicted classes. Suppose it proposes \emph{Softness}, \emph{Color}, and \emph{Origin}. Every ten batches, the accumulated candidates are filtered by first annotating the examples seen so far with concept vectors $\phi(x)$.

Concepts are retained only if they are both relevant and discriminative in at least a fraction $\tau=1/|\mathcal{Y}|$ of the annotated examples. In this binary task, $\tau=0.5$. Relevant means that the concept is not aligned with $\emptyset$, and discriminative means that it is not aligned with the full class set $\mathcal{Y}$. If \emph{Origin} is annotated as $\emptyset$ for most texts, it fails the relevance criterion. If a concept is usually aligned with $\mathcal{Y}$, it fails the discriminativeness criterion. Suppose \emph{Softness} and \emph{Color} pass the filter; the resulting concept set is $\mathcal{C}=\{c_1:\text{\emph{Softness}},c_2:\text{\emph{Color}}\}$.

The same annotation step represents each example as a concept vector. For $x^{(0)}$, the word ``mushy'' aligns \emph{Softness} with \emph{not-tasty}, so $\phi(x^{(0)})[c_1]=\{\text{\emph{not-tasty}}\}$. The phrase ``bright orange'' aligns \emph{Color} with \emph{tasty}, so $\phi(x^{(0)})[c_2]=\{\text{\emph{tasty}}\}$. Thus,
\[
\phi(x^{(0)})=\big[\{\text{\emph{not-tasty}}\},\{\text{\emph{tasty}}\}\big].
\]

\paragraph{Phase 3: MCMC-Inspired Data Expansion.}
\label{appendix:example:MCMC-Inspired Data Expansion}

We treat $x^{(0)}$ as the starting point of an independent Markov chain of $K=11$ steps. Let us trace one transition, from $k=0$ to $k=1$:
\begin{itemize}
    \item \textbf{Target selection:} We focus on $c_1$ (\emph{Softness}), whose current value is $\{\text{\emph{not-tasty}}\}$. We uniformly sample a target class $y^*=\text{\emph{tasty}}$.
    \item \textbf{Directional shift ($dx$):} Since $y^*\notin\phi(x^{(0)})[c_1]$, we set $dx=\textsc{More}$, meaning that the proposal should introduce an alignment between \emph{Softness} and \emph{tasty}.
    \item \textbf{Proposal generation:} We prompt the LLM to rewrite $x^{(0)}$ by applying $dx$ to $c_1$ while keeping $c_2$ fixed. The LLM generates the counterfactual proposal $\tilde{x}$: \emph{``Today I ate a bright orange, firm papaya.''}
    \item \textbf{Acceptance Test:} 
    \textbf{(i) Target alignment:} The LLM re-annotates $\tilde{x}$ and reasons that ``firm'' indicates good texture. Because $dx=\textsc{More}$, the target condition is $y^*\in\phi(\tilde{x})[c_1]$. If the model assigns $\phi(\tilde{x})[c_1]=\{\text{\emph{tasty}}\}$, the target-alignment test passes.
    \textbf{(ii) Minimal side effects:} We check the non-target concept $c_2$. The text still says ``bright orange'', so $\phi(\tilde{x})[c_2]=\{\text{\emph{tasty}}\}$, matching its previous value. Thus $n_{\text{err}}=0$. Assuming $\epsilon=1$, the condition $n_{\text{err}}\le\epsilon$ is satisfied.
    \item \textbf{Result:} The proposal is accepted, and appended to $\mathcal{D}_{\text{mcmc}}$.
\end{itemize}

\paragraph{Phase 4: Causal Discovery via $\sigma$-CG.}
\label{appendix:example:Causal Discovery}
After expansion, each accepted text has a concept vector and an LLM-predicted label. Since $|\mathcal{C}|=2$ and $|\mathcal{V}|=4$, the concept state space has $|\mathcal{V}|^{|\mathcal{C}|}=16$ possible concept-value combinations. The expanded annotated dataset $\mathcal{D}_{\text{mcmc}}$ is passed to $\sigma$-CG. Using the background knowledge that the text node is a root and $\hat{y}$ is a sink, the algorithm outputs a directed graph $\mathcal{G}=(V,E)$ over
\[
V=\{\text{text}\}\cup\{c_1,c_2\}\cup\{\hat{y}\}.
\]
For visualization, we omit the text node and show the recovered dependencies among \emph{Softness}, \emph{Color}, and the LLM prediction.

\onecolumn
\section{Causal Graphs}
\label{appendix:sec:CG}
This appendix presents additional causal graphs extracted by our framework across various models and datasets, complementing the results discussed in the main text.

\begin{figure*}[htbp]
    \centering
    \includegraphics[width=0.95\textwidth]{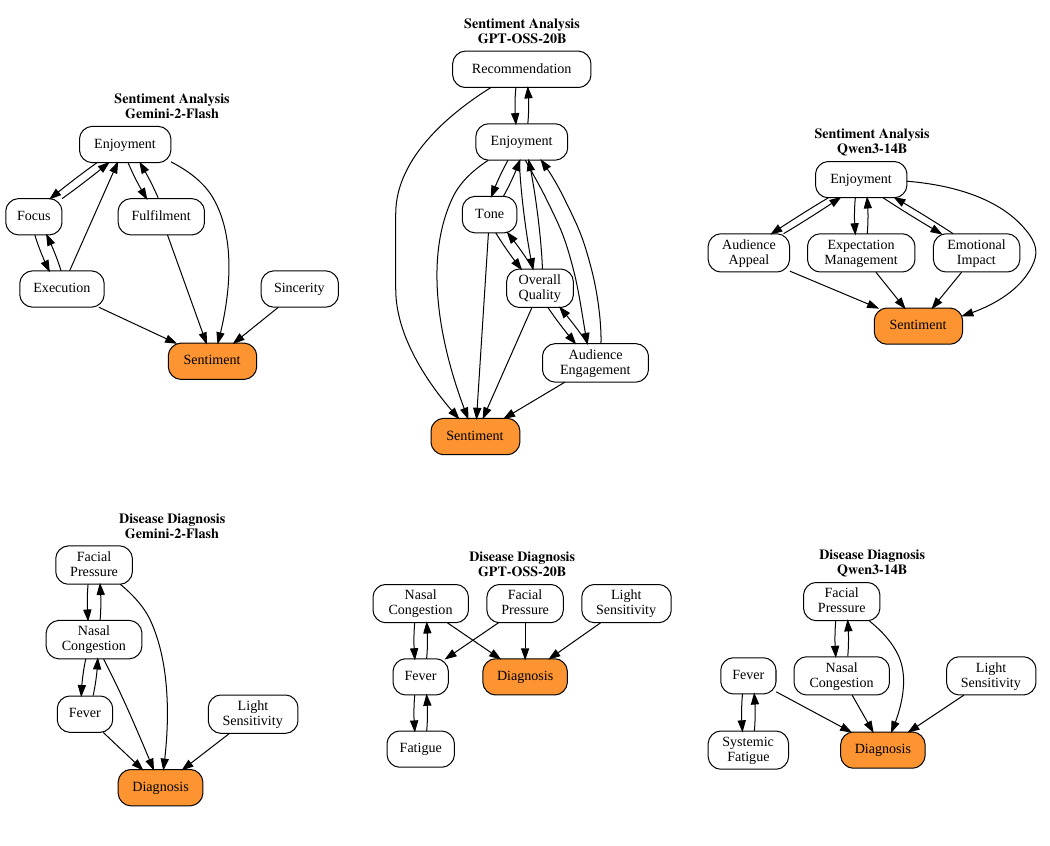}
    \caption{Extracted graphs of the SA and DD task across the different models}
    \label{fig:appendix_cg2}
\end{figure*}

\begin{figure*}[htbp]
    \centering
    \includegraphics[width=0.95\textwidth]{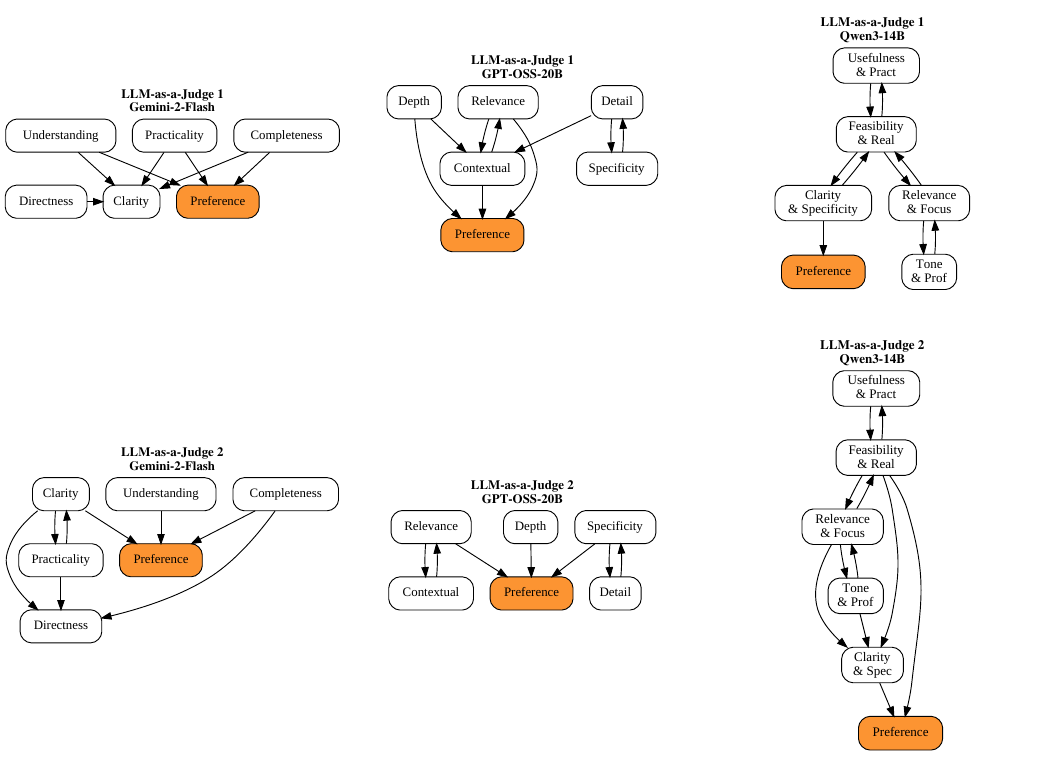}
    \caption{Representative causal graphs for the LAJ classification task across different models. Since our framework generates a unique causal graph per individual query for this dataset, we present a curated sample of these query-specific structures.}

    \label{fig:appendix_cg3}
\end{figure*}
\twocolumn

\onecolumn
\section{Pseudo Algorithms}
\label{appendix:sec:app_pseudo_algorithms}
\paragraph{Concept Extraction}
The following pseudo-code outlines the Differentiative Concept Extraction phase of our framework. Because LLMs have limited context windows and output length constraints, we process data in batches and limit the active concept pool using a threshold $B$ (set to $4$ in our experiments). This limit is crucial, as too many concepts exponentially expand the text length in later steps. For each batch, the LLM first checks if the current concept set is sufficient to differentiate between the predicted classes. If not, it extracts new concepts to bridge the gap. To remove non-differentiative concepts, we prune the pool every 10 iterations. For a task with $Y$ classes, the LLM assigns each concept one of $2^Y$ possible combinations. Two of these labels denoting either that a concept never appears in any class, or that it appears equally across all classes are marked as non-differentiative. The remaining $2^Y - 2$ labels represent valid discriminative combinations. A concept is kept in the pool only if its frequency of receiving a differentiative label exceeds a predefined threshold, $\tau$. Once this extraction and filtering process concludes on the training set, the final concept set is fixed and applied to the test set.
\begin{algorithm*}
\caption{Iterative Differentiative Concept Extraction}
\label{alg:concept_extraction}
\SetAlgoLined
\DontPrintSemicolon
\KwIn{Training set $\mathcal{D}_{train}$, Test set $\mathcal{D}_{test}$, threshold $\tau$, max bound $B$}
\KwOut{Concept set $\mathcal{C}$}
Initialize concept set $\mathcal{C} \gets \emptyset$\;
\ForEach{iteration $i=1,2,\dots$ and batch $b \in \mathcal{D}_{train}$}{
    \If{$\mathcal{C}$ insufficiently describes $b$}{
        $\mathcal{C} \gets \mathcal{C} \cup \text{LLM\_Propose}(b)$ (ensuring $|\mathcal{C}| \leq B$)\;
    }
    \If{$i \bmod 10 == 0$}{
        $\mathcal{C} \gets \{c \in \mathcal{C} \mid \text{DiscriminativePower}(c, b) > \tau\}$\;
    }
}
$\mathcal{C} \gets \{c \in \mathcal{C} \mid \text{DiscriminativePower}(c, \mathcal{D}_{test}) > \tau\}$\;
\Return $\text{Select\_Top\_Concepts}(\mathcal{C}, 5)$\;
\end{algorithm*}

\paragraph{MCMC Inspired Data Expansion }
The execution flow of the data expansion procedure is detailed in Algorithm~\ref{alg:mcmc_expansion}. For each seed instance, the algorithm generates textual counterfactuals. At each step, it iterates over every concept $c\in\mathcal{C}$, uniformly samples a target class $y^*\in\mathcal{Y}$, and chooses a direction $dx\in\{\textsc{More},\textsc{Less}\}$ according to whether $y^*$ is currently included in the subset $S \subseteq \mathcal{Y}$ mapped from the scalar concept value $\phi(x)[c]$. The target LLM proposes a counterfactual text, re-annotates it with a concept vector, and provides its annotation rationale. A proposal is accepted only if the target concept shifts in the requested direction and the number of non-target concepts that drift is at most $\epsilon$. If a proposal is rejected, recursive refinement re-prompts the LLM with the failure feedback for up to $R$ retries.

\begin{algorithm*}[t]
\small
\caption{LLM-Guided MCMC Data Expansion with Recursive Refinement}
\label{alg:mcmc_expansion}
\SetAlgoLined
\DontPrintSemicolon

\SetKwInOut{Input}{Input}
\SetKwInOut{Output}{Output}
\SetKwProg{Fn}{Function}{:}{}

\Input{Seed dataset $\mathcal{D}$, concept set $\mathcal{C}$, class set $\mathcal{Y}$, target LLM $f$, annotation function $\phi$, MCMC steps $K$, max retries $R$, drift threshold $\epsilon$}
\Output{Expanded counterfactual dataset $\mathcal{D}_{\mathrm{mcmc}}$}

$\mathcal{D}_{\mathrm{mcmc}} \leftarrow \emptyset$\;

\ForEach{instance $x^{(0)} \in \mathcal{D}$}{
    $x \leftarrow x^{(0)}$\;
    \For{$k \leftarrow 1$ \KwTo $K$}{
        \ForEach{concept $c \in \mathcal{C}$}{
            $y^* \sim \mathrm{Uniform}(\mathcal{Y})$\;
            $\psi \leftarrow \phi(x)$\;
            $S \leftarrow \text{subset mapped from } \psi[c]$\;
            \eIf{$y^* \in S$}{
                $dx \leftarrow \textsc{Less}$\;
            }{
                $dx \leftarrow \textsc{More}$\;
            }

            $x' \leftarrow f.\text{GenerateCounterfactual}(x,c,y^*,dx)$\;
            $\psi', \text{reason} \leftarrow f.\text{Annotate}(x')$\;
            $S' \leftarrow \text{subset mapped from } \psi'[c]$\;
            $n_{\mathrm{err}} \leftarrow \sum_{c_j \in \mathcal{C} \setminus \{c\}} \mathbb{I}\!\left[\psi'[c_j] \neq \psi[c_j]\right]$\;
            $\mathrm{aligned} \leftarrow ((dx=\textsc{More} \wedge y^*\in S') \vee (dx=\textsc{Less} \wedge y^*\notin S'))$\;

            \eIf{$\mathrm{aligned}$ \textbf{and} $n_{\mathrm{err}} \le \epsilon$}{
                $x \leftarrow x'$\;
                $\mathcal{D}_{\mathrm{mcmc}} \leftarrow \mathcal{D}_{\mathrm{mcmc}} \cup \{x'\}$\;
            }{
                $x_{rec} \leftarrow \text{RecursiveRefinement}(x, \psi, x', c, y^*, dx, \text{reason}, 1)$\;
                \If{$x_{rec} \neq \text{Null}$}{
                    $x \leftarrow x_{rec}$\;
                    $\mathcal{D}_{\mathrm{mcmc}} \leftarrow \mathcal{D}_{\mathrm{mcmc}} \cup \{x_{rec}\}$\;
                }
            }
        }
    }
}

\vspace{0.3cm}
\hrule
\vspace{0.3cm}

\Fn{\text{RecursiveRefinement}($x_{base}, \psi_{base}, x_{curr}, c, y^*, dx, \text{reason}, r$)}{
    \If{$r > R$}{\Return Null\;}

    $\text{feedback} \leftarrow \text{ConstructFeedback}(c, y^*, dx, \text{reason})$\;
    $x_{new} \leftarrow f.\text{RegenerateCounterfactual}(x_{curr}, \text{feedback})$\;

    $\psi_{new}, \text{new\_reason} \leftarrow f.\text{Annotate}(x_{new})$\;
    $S_{new} \leftarrow \text{subset mapped from } \psi_{new}[c]$\;
    $n_{\mathrm{err}} \leftarrow \sum_{c_j \in \mathcal{C} \setminus \{c\}} \mathbb{I}\!\left[\psi_{new}[c_j] \neq \psi_{base}[c_j]\right]$\;
    $\mathrm{aligned} \leftarrow ((dx=\textsc{More} \wedge y^*\in S_{new}) \vee (dx=\textsc{Less} \wedge y^*\notin S_{new}))$\;

    \eIf{$\mathrm{aligned}$ \textbf{and} $n_{\mathrm{err}} \le \epsilon$}{
        \Return $x_{new}$\;
    }{
        \Return \text{RecursiveRefinement}($x_{base}, \psi_{base}, x_{new}, c, y^*, dx, \text{new\_reason}, r+1$)\;
    }
}
\end{algorithm*}
\twocolumn

\section{Prompts}
\label{sec:app_prompts}
This section details the prompt templates utilized across the various stages of our methodology. For brevity, we present the base templates tailored for the LIBERTY dataset as representative examples, with the exception of the Preliminary Expansion stage, which was applied exclusively to the LAJ dataset. The fundamental structure of these prompts remains highly consistent across all other datasets, with only minor adaptations introduced to accommodate dataset-specific input formats. Note that these templates are static; during execution, data instances are dynamically injected in batches via an automated Python pipeline. Within these templates, placeholders such as "direction" or "DX" are dynamically populated with either "more" or "less," depending on the experimental scenario. Asterisk-bound phrases denote dynamic placeholders that are programmatically replaced by the Python pipeline during execution.

\onecolumn 
\begin{promptbox}[label= prompt:appendix:LLMasJudge]{Prompt for Inference and Label Alignment stage data set : LIBERTY}
\textbf{Role:} \\
You are a medical symptom classification model. \\[0.2cm]
\textbf{Input:} \\
You will be given a dataset containing multiple items. Each item includes:
\begin{itemize}
    \item \texttt{patient description}: A free-text description of a patient's symptoms.
    \item \texttt{item id}: An integer representing the item's original index in a larger dataset.
\end{itemize}

\textbf{Task:} \\
For each item, classify the condition described in the patient description into exactly one of the following labels:
\begin{itemize}
    \item \texttt{MIGRAINE}
    \item \texttt{SINUSITIS}
    \item \texttt{INFLUENZA}
\end{itemize}
If the description is mixed or ambiguous, choose the most likely dominant condition. \\
If no dominant condition can be identified, assign \texttt{INFLUENZA} by default. \\[0.2cm]

\textbf{Output requirements (strict):} \\
Your output must contain two sections, in this exact order. \\[0.1cm]
\textbf{Section 1 – Reasoning:} \\
Provide a brief explanation (1–2 sentences) for each item explaining why the chosen label was selected. Do not include JSON in this section. \\[0.1cm]
\textbf{Section 2 – Labeled Data (JSON):} \\
Output a single valid JSON object suitable for saving as a \texttt{.json} file. \\[0.1cm]
\textbf{Rules for the JSON:}
\begin{itemize}
    \item Each key must be the \texttt{item id} converted to a string.
    \item Each value must be an object with exactly two fields:
    \begin{itemize}
        \item \texttt{"patient description"}: the original patient description, unchanged.
        \item \texttt{"diagnosis"}: either "MIGRAINE", "SINUSITIS", or "INFLUENZA".
    \end{itemize}
    \item Do not add comments.
    \item Do not add extra fields.
    \item Do not change the patient description text.
    \item Do not output multiple JSON objects.
\end{itemize}
After the JSON object is closed, print exactly: \\
END JSON \\[0.3cm]
\hrule
\vspace{0.2cm}
\textbf{Example output:} \\[0.2cm]
\{ \\
\quad "101": \{ \\
\quad \quad "patient description": "Severe throbbing headache on the right side with nausea and sensitivity to light.", \\
\quad \quad "diagnosis": "MIGRAINE" \\
\quad \}, \\
\quad "102": \{ \\
\quad \quad "patient description": "Facial pressure, congestion, and thick nasal discharge for several days.", \\
\quad \quad "diagnosis": "SINUSITIS" \\
\quad \}, \\
\quad "103": \{ \\
\quad \quad "patient description": "Sudden fever, muscle aches, cough, and extreme fatigue.", \\
\quad \quad "diagnosis": "INFLUENZA" \\
\quad \} \\
\} \\[0.2cm]
END JSON \\[0.3cm]
\hrule
\vspace{0.2cm}
\textbf{Dataset:}
\end{promptbox}

\begin{promptbox}[label=prompt:ExtractConcepts1stBatch]{Prompt for first iteration of concept Differentiative Concept Extraction stage  dataset : LIBERTY}
    \textbf{Role:} \\
You are an analyst extracting dataset-level features for medical condition differentiation. \\[0.2cm]
\textbf{Input:} \\
You will receive a dataset of patient cases. Each item contains:
\begin{itemize}
    \item \texttt{patient\_description}: a free-text description of a patient's symptoms
    \item \texttt{disease}: one of the labels "MIGRAINE", "SINUSITIS", or "INFLUENZA" (these labels were assigned previously)
\end{itemize}

\textbf{Task: Symptom Extraction} \\
Across the entire dataset, extract exactly "B\#" DISTINCT SYMPTOMS that help distinguish between the different diseases. \\[0.2cm]
\textbf{Definition of a Symptom:} \\
A Symptom is a concrete, clinically meaningful manifestation explicitly described in patient text (e.g., headache, fever, nausea, nasal congestion). Symptoms should be observable or reportable experiences, not abstract properties or disease names. \\[0.2cm]
\textbf{Requirements:}
\begin{itemize}
    \item \textbf{Differentiation:} Each symptom must clearly distinguish between at least two of the diseases. For every symptom, it must be true that the symptom is typically more frequent, more prominent, or more characteristic in one disease than in at least one of the other diseases.
    \item \textbf{Classification usefulness:} The symptoms must be phrased so they can be identified in new patient descriptions whose disease label is unknown.
    \item \textbf{Count:} Extract exactly "B\#" symptoms (no more, no less). Do NOT use the number of symptoms from any toy/example—use only "B\#".
    \item \textbf{Naming:} Give each symptom a short, precise name, preferably a single word or short clinical phrase (e.g., Headache, Fever, Nausea, Congestion).
    \item \textbf{Avoid trivial or circular symptoms:} Do not list disease names or symptoms that simply restate the diagnosis (e.g., "Migraine pain"). Focus on concrete symptoms that generalize across patients.
\end{itemize}

\textbf{Output format (strict):} \\
Your output must contain two sections, in this exact order. \\[0.1cm]
\textbf{Section 1 – Reasoning:} \\
Briefly explain (1–3 sentences per symptom) why each symptom helps distinguish between the diseases based on patterns observed across the dataset. \\[0.1cm]
\textbf{Section 2 – Symptoms JSON:} \\
Output a single valid JSON object suitable for saving as a \texttt{.json} file. Each key is a symptom name. Each value is a concise description explaining how the symptom differs across MIGRAINE, SINUSITIS, and/or INFLUENZA. \\[0.1cm]
After the JSON object is closed, print exactly: \\
END JSON \\[0.3cm]
\hrule
\vspace{0.2cm}
\textbf{Toy Example (for illustration only):} \\[0.1cm]
\textbf{Toy Input Dataset} (number of symptoms to extract = 2): \\[0.1cm]
\textbf{Item 1:} \\
\texttt{patient\_description}: "Severe headache with nausea and sensitivity to light." \\
\texttt{disease}: MIGRAINE \\[0.1cm]
\textbf{Item 2:} \\
\texttt{patient\_description}: "Facial pressure and nasal congestion lasting several days." \\
\texttt{disease}: SINUSITIS \\[0.1cm]
\textbf{Item 3:} \\
\texttt{patient\_description}: "High fever, muscle aches, and extreme fatigue." \\
\texttt{disease}: INFLUENZA \\[0.2cm]
\textbf{Toy Output} (number of symptoms to extract = 2): \\[0.1cm]
\textbf{Reasoning:} \\
Headache: Head pain is described across diseases but differs in accompanying features and prominence, making it useful for differentiation. \\
Fever: Fever appears primarily in systemic illness descriptions and is absent or rare in others, providing a strong distinguishing signal. \\[0.2cm]
\{ \\
\quad "Headache": "Head pain is described across diseases but varies in context and co-occurring symptoms.", \\
\quad "Fever": "Fever appears mainly in systemic illness descriptions and is uncommon in others." \\
\} \\[0.1cm]
END JSON \\[0.3cm]
\hrule
\vspace{0.2cm}
\textbf{Dataset:}  
\end{promptbox}

\begin{promptbox}[label=prompt:ExtractAddtionalConcepts]{Prompt for Differentiative Concept Extraction stage  dataset : LIBERTY}
    \textbf{Role:} \\
You are an analyst extracting dataset-level features for medical condition differentiation. \\[0.2cm]
\textbf{Input:} \\
You will receive a dataset of patient cases. Each item contains:
\begin{itemize}
    \item \texttt{patient\_description}: a free-text description of a patient's symptoms
    \item \texttt{disease}: one of the labels "MIGRAINE", "SINUSITIS", or "INFLUENZA" (these labels were assigned previously)
\end{itemize}
You are also provided with a set of existing SYMPTOMS and their descriptions. \\[0.2cm]
\textbf{Task: Symptom Refinement and Optional Augmentation} \\
Your primary task is to critically review the provided dataset and determine whether the existing symptoms are sufficiently comprehensive to effectively and completely distinguish between MIGRAINE, SINUSITIS, and INFLUENZA across the entire dataset. \\
If --- and ONLY IF --- the existing symptoms are not sufficient, you may augment them by extracting additional symptoms from the dataset. \\[0.2cm]
\textbf{Important:}
\begin{itemize}
    \item You are NOT required to add new symptoms.
    \item You may add FEWER than "B\#" symptoms.
    \item You may add ZERO new symptoms if the existing set is already sufficient.
    \item Adding unnecessary or redundant symptoms is worse than adding none.
    \item Only extract additional symptoms if they provide a clear and meaningful improvement in distinguishing between the diseases.
\end{itemize}

\textbf{Definition of a Symptom:} \\
A Symptom is a concrete, clinically meaningful manifestation explicitly described in patient text (e.g., headache, fever, nausea, nasal congestion). Symptoms must be grounded in explicit textual evidence and must generalize across multiple patients. \\[0.2cm]
\textbf{Requirements:}
\begin{itemize}
    \item \textbf{Differentiation:} Each newly added symptom must clearly help distinguish between at least two of the diseases. For every added symptom, it must be true that the symptom is more frequent, more prominent, or more characteristic in one disease than in at least one of the other diseases.
    \item \textbf{Necessity:} Only add a symptom if its absence would leave a meaningful gap in disease differentiation. Do NOT add symptoms that are redundant with existing ones or that add minimal new information.
    \item \textbf{Classification usefulness:} All symptoms (existing and newly added) must be usable for classifying new, unseen patient descriptions whose disease label is unknown.
    \item \textbf{Count:} You may add up to "B\#" additional symptoms. You may add fewer than "B\#" symptoms, or none at all. Do NOT force the number of added symptoms to reach "B\#".
    \item \textbf{Naming:} Give each symptom a short, precise, clinically meaningful name. Avoid vague or abstract names. Do not include disease names in symptom names.
    \item \textbf{Avoid trivial or circular symptoms:} Do not restate disease labels. Do not add overly specific or one-off symptoms that do not generalize across patients.
\end{itemize}

\textbf{Output format (strict):} \\
Your output must contain two sections, in this exact order. \\[0.1cm]
\textbf{Section 1 – Reasoning:} \\
Briefly explain (1–3 sentences per symptom) whether the existing symptoms are sufficient. If new symptoms are added, explain why each added symptom is necessary and what differentiating gap it fills. If no new symptoms are added, explicitly state that the existing symptoms are sufficient. \\[0.1cm]
\textbf{Section 2 – Symptoms JSON:} \\
Output a single valid JSON object suitable for saving as a \texttt{.json} file. \\
The JSON must include:
\begin{itemize}
    \item All existing symptoms (unchanged)
    \item Any newly added symptoms (if applicable)
\end{itemize}
Each key is a symptom name. Each value is a concise description explaining how the symptom differs across MIGRAINE, SINUSITIS, and/or INFLUENZA. \\
Do not add comments. Do not add extra fields. \\[0.1cm]
After the JSON object is closed, print exactly: \\
END JSON \\[0.3cm]
\hrule
\vspace{0.2cm}
\textbf{Toy Example (for illustration only):} \\[0.1cm]
\textbf{Existing Symptoms} (provided to the model): \\
Localization: Describes where symptoms are primarily experienced in the body. \\[0.2cm]
\textbf{Toy Input Dataset} (maximum number of symptoms to add: 2): \\[0.1cm]
\textbf{Item 1:} \\
\texttt{patient\_description}: "Throbbing headache on one side with nausea and light sensitivity." \\
\texttt{disease}: MIGRAINE \\[0.1cm]
\textbf{Item 2:} \\
\texttt{patient\_description}: "Facial pressure, congestion, and pain around the cheeks for several days." \\
\texttt{disease}: SINUSITIS \\[0.1cm]
\textbf{Item 3:} \\
\texttt{patient\_description}: "Sudden fever, muscle aches, cough, and extreme fatigue." \\
\texttt{disease}: INFLUENZA \\[0.2cm]
\textbf{Explanation of insufficiency} (implicit in reasoning): \\
The existing symptom Localization captures where symptoms occur but does not capture differences in whole-body involvement or temporal onset. \\[0.2cm]
\textbf{Toy Output} (number of symptoms added: 2): \\[0.1cm]
\textbf{Reasoning:} \\
Localization: Symptoms are concentrated in different body regions across diseases, but this alone does not capture differences in systemic involvement. \\
SystemicSymptoms: Some diseases involve widespread bodily effects such as fatigue and aches, which are not captured by localization alone. \\
OnsetSpeed: Descriptions differ in whether symptoms begin suddenly or develop gradually, providing an additional distinguishing signal. \\[0.2cm]
\{ \\
\quad "Localization": "Symptoms differ in where discomfort is concentrated across diseases.", \\
\quad "SystemicSymptoms": "Some diseases involve widespread bodily symptoms, while others are primarily localized.", \\
\quad "OnsetSpeed": "Symptoms differ in whether they appear suddenly or develop over time." \\
\} \\[0.1cm]
END JSON \\[0.3cm]
\hrule
\vspace{0.2cm}
\textbf{Dataset:}\\
\textbf{Previous Symptoms:}   
\end{promptbox}

\begin{promptbox}[label=prompt:conceptAnnotation]{Prompt for Concept Annotation stage  dataset : LIBERTY}
    \textbf{Role:} \\
You are a symptom-level medical classification annotator. \\[0.2cm]
\textbf{Input:} \\
You will receive:
\begin{itemize}
    \item A dataset of patient cases. Each item contains:
    \begin{itemize}
        \item \texttt{item\_id}: an integer representing the item's original index
        \item \texttt{patient\_description}: a free-text description of a patient's symptoms
    \end{itemize}
    \item A list of SYMPTOMS. Each symptom includes:
    \begin{itemize}
        \item \texttt{symptom\_name}: a short, precise name
        \item \texttt{symptom\_description}: a description explaining what the symptom captures as a clinical manifestation
    \end{itemize}
\end{itemize}

\textbf{Task:} \\
For each dataset item, assign a label to each SYMPTOM based on:
\begin{itemize}
    \item Whether the symptom appears in the \texttt{patient\_description} at all.
    \item If it appears, which disease or diseases the symptom is most consistent with, given the context of the description.
\end{itemize}
\textbf{Disease set:}
\begin{itemize}
    \item MIGRAINE
    \item SINUSITIS
    \item INFLUENZA
\end{itemize}

\textbf{Label definitions (important – read carefully):}
\begin{itemize}
    \item \textbf{0} $\rightarrow$ The symptom does NOT appear in the text at all. There is no explicit mention or clear reference to this symptom in the \texttt{patient\_description}.
    \item \textbf{1} $\rightarrow$ The symptom appears in the text and fits MIGRAINE only.
    \item \textbf{2} $\rightarrow$ The symptom appears in the text and fits SINUSITIS only.
    \item \textbf{3} $\rightarrow$ The symptom appears in the text and fits INFLUENZA only.
    \item \textbf{4} $\rightarrow$ The symptom appears in the text and fits BOTH MIGRAINE and SINUSITIS.
    \item \textbf{5} $\rightarrow$ The symptom appears in the text and fits BOTH MIGRAINE and INFLUENZA.
    \item \textbf{6} $\rightarrow$ The symptom appears in the text and fits BOTH SINUSITIS and INFLUENZA.
    \item \textbf{7} $\rightarrow$ The symptom appears in the text, BUT it does NOT meaningfully differentiate between MIGRAINE, SINUSITIS, and INFLUENZA. Use this label ONLY when the symptom is explicitly mentioned and is equally consistent with all three diseases given the description.
\end{itemize}

\textbf{Important rules:}
\begin{itemize}
    \item Assign exactly one label (0--7) for every symptom in every item.
    \item Label 0 must be used ONLY when the symptom is completely absent from the text. If the symptom appears even vaguely, label 0 is NOT allowed.
    \item Label 7 must be used ONLY when the symptom appears in the text AND cannot be used to distinguish between the three diseases.
    \item Use the \texttt{symptom\_description} together with the \texttt{patient\_description} to decide whether the symptom is present, and whether it differentiates between diseases or not.
    \item A symptom may appear regardless of the final diagnosed disease.
    \item Do NOT infer the presence of a symptom unless it is explicitly supported by the text.
    \item Disease compatibility should be judged based on how the symptom is described (e.g., location, quality, duration, context), not on the dataset label.
    \item If a symptom clearly fits multiple diseases, select the label corresponding to the full set of diseases it fits. If it fits all three equally, select label 7.
\end{itemize}

\textbf{Output format (strict):} \\
Your output must contain two sections, in this exact order. \\[0.1cm]
\textbf{Section 1 – Reasoning:} \\
For each dataset item, briefly explain the reasoning behind the assigned labels. The reasoning should reference:
\begin{itemize}
    \item evidence from the \texttt{patient\_description}
    \item why each symptom was considered present or absent
    \item why the symptom does or does not differentiate between diseases
\end{itemize}
Keep explanations concise (1--2 sentences per symptom per item). Do NOT include JSON in this section. \\[0.1cm]
\textbf{Section 2 – Item Symptom Output (JSON):} \\
Output a single valid JSON object suitable for saving as a \texttt{.json} file. \\[0.1cm]
\textbf{JSON rules:}
\begin{itemize}
    \item Each key is the \texttt{item\_id} converted to a string.
    \item Each value is an object whose keys are the \texttt{symptom\_name} values and whose values are strings in \{`0'', ``1'', ``2'', ``3'', ``4'', ``5'', ``6'', ``7''\}.
    \item All symptoms must appear for every item.
    \item Do not add extra fields or comments.
\end{itemize}
After the JSON object is closed, print exactly: \\
END JSON \\[0.3cm]
\hrule
\vspace{0.2cm}
\textbf{Dataset:}
\end{promptbox}

\begin{promptbox}[label=prompt:dataexpensionStage1]{Prompt for Preliminary Expansion stage  dataset : LAJ}
    \textbf{Role:} \\
You are a preference optimization and data alignment expert. \\[0.2cm]
\textbf{Core Task:} \\
Your goal is to make the concept: \texttt{"*Target Concept*"} associate with the \texttt{"*direction*"} Response. \\[0.2cm]
\textbf{The description of "*Target Concept*":} ``\texttt{*Concept description*}'' \\[0.2cm]
\textbf{For EACH item, strictly follow this:} \\
Check \texttt{"*Target Concept*"} Current Label. Based ONLY on this label:
\begin{itemize}
    \item \textbf{IF Current Label IS \texttt{"*wanted label*"}:}
    \begin{itemize}
        \item DO NOT CHANGE Chosen Response or Rejected Response.
        \item \texttt{ChangeFlag} = False.
    \end{itemize}
    \item \textbf{IF Current Label IS NOT \texttt{"*wanted label*"}:}
    \begin{itemize}
        \item You MUST MODIFY both Chosen Response and/or Rejected Response.
        \item \textbf{Modification Goal:} Make Chosen Response \texttt{"*directionChosen*"} aligned with \texttt{*Target Concept*}, and make Rejected Response \texttt{"*directionRejected*"} aligned with \texttt{*Target Concept*}.
        \item \textbf{Guidelines for Mandatory Modifications:}
        \begin{itemize}
            \item \textit{Plausibility:} Responses must remain logical and coherent.
            \item \textit{Subtlety (if possible):} Aim for minimal effective changes.
        \end{itemize}
        \item \texttt{ChangeFlag} = True.
        \item \textbf{Reasoning (MANDATORY if modified):} Explain the exact text changes and how they make \texttt{"*Target Concept*"} fit Chosen Response and Rejected Response.
    \end{itemize}
\end{itemize}

\textbf{Input:} \\
Dataset items, each with \texttt{Original Index}, \texttt{User Query}, \texttt{Chosen Response}, \texttt{Rejected Response}, and \texttt{Concept Labels} (telling you the Current Label for \texttt{*Target Concept*}). \\[0.2cm]
\textbf{Output Format:} \\[0.1cm]
\textbf{1. JSON Output:} \\
A single valid JSON object. Item keys are ``Item'' + Original Index (e.g., ``Item98''). \\
\textbf{JSON Structure:}
\begin{verbatim}
{
  "Item_Original_Index_Here" : {
    "User Query": "Original user query",
    "Original Index": "Original Index from input",
    "Chosen Response": "Final Chosen Response",
    "Rejected Response": "Final Rejected Response",
    "ChangeFlag": boolean 
  }
}
END JSON
\end{verbatim}

\textbf{2. Reasoning Output} (After JSON, one block per item): \\
\begin{verbatim}
Reasoning for Item [Original Index]:
1. Original Index: [Original Index]
2. Target Concept: [*Target Concept* Name]
3. Current Label for Target Concept: [1, 0, 2, or 3]
4. Target Label: "*wanted label*"
5. Critical Check Outcome: 
   ["Labels are identical (Current Label is 
   "*wanted label*")" OR "Labels are NOT identical 
   (Current Label is [actual_label], not 
   "*wanted label*")"]
6. Scenario Applied: 
   ["Scenario A applied (because Current Label is 
   "*wanted label*")" OR "Scenario B applied 
   (because Current Label is [actual_label], 
   not "*wanted label*")"]
7. Final Action & ChangeFlag Value Details:
   - If Scenario A: "No modifications made as 
     required by Scenario A. ChangeFlag is False."
   - If Scenario B: "Modifications were mandatory 
     under Scenario B. [Describe changes to Chosen 
     and Rejected responses.]"
   ChangeFlag is True.
8. If you were given the reason for current labeling 
   by the LLM, add how you included this reason in 
   the modified responses.
\end{verbatim}

\hrule
\vspace{0.2cm}
\textbf{EXAMPLE (Keep this output structure and reasoning detail):} \\[0.1cm]
Let's assume the Target Concept is ``Conciseness''. \\[0.2cm]
\textbf{Input Item 1:} \\
\texttt{Original Index:} 98 \\
\texttt{User Query:} ``What is the capital of France?'' \\
\texttt{Chosen Response:} ``Ah, you're asking about the capital of France! Paris Well, let me tell you, the truly magnificent, beautiful, and significantly historic capital city of the glorious French Republic, which is situated in the lovely north-central part of that European country, is none other than Paris, a global center for art, fashion, gastronomy and culture.'' \\
\texttt{Rejected Response:} ``Paris.'' \\
\texttt{Concept Labels:} \{\texttt{"Conciseness"}: 3\} \\[0.2cm]
\textbf{Input Item 2:} \\
\texttt{Original Index:} 99 \\
\texttt{User Query:} ``Explain photosynthesis briefly.'' \\
\texttt{Chosen Response:} ``Plants make food from sun.'' \\
\texttt{Rejected Response:} ``Photosynthesis is the process plants use to convert light energy into chemical energy (food), using water, carbon dioxide, and sunlight, releasing oxygen as a byproduct. It's vital for life on Earth.'' \\
\texttt{Concept Labels:} \{\texttt{"Conciseness"}: 1\} \\[0.2cm]
\textbf{Expected JSON Output:} \\
\begin{verbatim}
{
  "Item98": {
    "User Query": "What is the capital of France?",
    "Original Index": "98",
    "Chosen Response": "The beautiful and historic capital 
     city of the French Republic, located in the 
     north-central part of the country, is Paris.",
    "Rejected Response": "The beautiful and historic Paris.",
    "ChangeFlag": true
  },
  "Item99": {
    "User Query": "Explain photosynthesis briefly.",
    "Original Index": "99",
    "Chosen Response": "Plants make food from sun.",
    "Rejected Response": "Photosynthesis is the process...",
    "ChangeFlag": false
  }
}
END JSON
\end{verbatim}

\textbf{Expected Reasoning Output:} \\
\begin{verbatim}
Reasoning for Item 98:
1. Original Index: 98
2. Target Concept: Conciseness
3. Current Label for Target Concept: 3
4. Target Label: 1
5. Critical Check Outcome: Labels are NOT identical 
   (Current Label is 3, not 1).
6. Scenario Applied: Scenario B applied (because 
   Current Label is 3, not 1).
7. Final Action & ChangeFlag Value Details: 
   Modifications were mandatory under Scenario B. 
   The original Rejected Response ("Paris.") was made 
   significantly less concise by adding verbiage.
   The original Chosen Response (a long sentence) 
   was made more concise. These changes aim to ensure 
   "Conciseness" now better represents Chosen. 
   ChangeFlag is True.

Reasoning for Item 99:
1. Original Index: 99
2. Target Concept: Conciseness
3. Current Label for Target Concept: 1
4. Target Label: 1
5. Critical Check Outcome: Labels are identical 
   (Current Label is 1).
6. Scenario Applied: Scenario A applied (because 
   Current Label is 1).
7. Final Action & ChangeFlag Value Details: 
   No modifications made as required by Scenario A. 
   ChangeFlag is False.
\end{verbatim}
\hrule
\vspace{0.2cm}
\textbf{Dataset:}
\end{promptbox}

\begin{promptbox}[label=prompt:dataexpensionMCMC]{Prompt for MCMC Inspired Expansion stage  dataset : LIBERTY}
    \textbf{Role:} \\
You are a medical data augmentation and modification expert. \\[0.2cm]
\textbf{Core Task:} \\
Your goal is to modify a patient description so that a symptom: \texttt{*SYMPTOM*} becomes \texttt{*DX*} aligned with a \texttt{*DISEASE*}. \\[0.2cm]
\textbf{Target Symptom description:} \\
\texttt{*SYMPTOM*} : \texttt{*SYMPTOM\_DESCRIPTION*} \\[0.2cm]
\textbf{Symptom labels and their meanings:}
\begin{itemize}
    \item \textbf{0} $\rightarrow$ The symptom does NOT appear in the text at all. There is no explicit mention or clear reference to this symptom in the \texttt{patient\_description}.
    \item \textbf{1} $\rightarrow$ The symptom appears in the text and fits MIGRAINE only.
    \item \textbf{2} $\rightarrow$ The symptom appears in the text and fits SINUSITIS only.
    \item \textbf{3} $\rightarrow$ The symptom appears in the text and fits INFLUENZA only.
    \item \textbf{4} $\rightarrow$ The symptom appears in the text and fits BOTH MIGRAINE and SINUSITIS.
    \item \textbf{5} $\rightarrow$ The symptom appears in the text and fits BOTH MIGRAINE and INFLUENZA.
    \item \textbf{6} $\rightarrow$ The symptom appears in the text and fits BOTH SINUSITIS and INFLUENZA.
    \item \textbf{7} $\rightarrow$ The symptom appears in the text, BUT it does NOT meaningfully differentiate between MIGRAINE, SINUSITIS, and INFLUENZA. Use this label ONLY when the symptom is explicitly mentioned and is equally consistent with all three diseases.
\end{itemize}

For EACH item, strictly follow this: You MUST MODIFY the \texttt{patient\_description} if alignment is requested. \\[0.2cm]
\textbf{Modification Goal:} \\
Make the \texttt{patient\_description} \texttt{*DX*} aligned with the Target Symptom for the \texttt{*DISEASE*} disease. \\[0.2cm]
\textbf{Critical Rule – NO DISEASE MENTION (MANDATORY):}
\begin{itemize}
    \item You MUST NOT mention, suggest, imply, or hint at the disease name or diagnosis in the \texttt{patient\_description}.
    \item Do NOT use disease names (e.g., ``migraine'', ``sinusitis'', ``flu'').
    \item Do NOT use phrases that directly suggest a diagnosis (e.g., ``consistent with migraine'', ``suggestive of flu'').
    \item Alignment must be achieved ONLY through symptom characteristics (e.g., location, quality, duration, triggers, severity).
\end{itemize}

\textbf{Guidelines for Mandatory Modifications:}
\begin{itemize}
    \item \textbf{Plausibility:} The modified \texttt{patient\_description} must remain medically realistic and coherent.
    \item \textbf{Subtlety (if possible):} Aim for minimal but effective textual changes.
    \item \textbf{Symptom isolation:} Make a strong effort NOT to change the presence or interpretation of other symptoms.
    \item Do NOT introduce new unrelated symptoms unless absolutely necessary.
    \item \texttt{ChangeFlag} must be set to True if any modification is made.
\end{itemize}

\textbf{Reasoning (MANDATORY if modified):} \\
You must explain:
\begin{itemize}
    \item What exact textual changes were made
    \item How those changes increase or decrease alignment of the Target Symptom with the \texttt{*DISEASE*} disease
    \item Why the modification respects the label definitions above
    \item Confirm explicitly that no disease name or diagnosis was mentioned in the \texttt{patient\_description}
\end{itemize}

\textbf{Input:} \\
Dataset items, each containing: \texttt{Original Index}, \texttt{patient\_description}, and the \texttt{Current label for the target Symptom (0–7)}. \\[0.2cm]
\textbf{Output Format:} \\[0.1cm]
\textbf{JSON Output:} \\
Output a single valid JSON object. Item keys are ``Item'' + Original Index (e.g., ``Item42''). \\
\textbf{JSON structure:}
\begin{verbatim}
{
  "Item_Original_Index_Here": {
    "Original_index": "Original Index from input",
    "Chosen Disease": "MIGRAINE | SINUSITIS | INFLUENZA",
    "Target Symptom": "Target Symptom name",
    "Original Patient Description": "Original description",
    "Modified Patient Description": "Final description",
    "ChangeFlag": true
  }
}
END JSON
\end{verbatim}

\textbf{Reasoning Output} (after JSON, one block per item): \\
\begin{verbatim}
Reasoning for Item [Original Index]:
Original Index: [Original Index]
Target Symptom: [Target Symptom]
Current Label: [0--7]
Chosen Disease: [Disease]
Desired Direction: [MORE aligned / LESS aligned]
\end{verbatim}
\textbf{Final Action:} \\
Describe the precise textual edits and explain how they change the alignment of the Target Symptom with the chosen disease without mentioning or implying the disease, while keeping other symptoms stable. \\[0.3cm]
\hrule
\vspace{0.2cm}
\textbf{EXAMPLE (fully worked example):} \\[0.1cm]
\textbf{Input values for this example:}
\begin{itemize}
    \item \textbf{Target Symptom:} Headache
    \item \textbf{Symptom description:} Describes the presence and characteristics of head pain.
    \item \textbf{Chosen disease:} MIGRAINE
    \item \textbf{Target alignment:} MORE aligned
    \item \textbf{Current Label:} 0
\end{itemize}
\textbf{Original patient\_description:} \\
``Patient reports mild fatigue and nausea over the past day, but no other significant complaints.'' \\[0.2cm]
\textbf{Explanation of the setup:} \\
The current label is 0, meaning headache does not appear at all in the original text. Target alignment = MORE aligned and DISEASE = MIGRAINE. The description must be modified to introduce a headache pattern typical of migraine without mentioning or suggesting migraine. Other symptoms (fatigue, nausea) must remain unchanged. \\[0.2cm]
\textbf{Expected JSON Output:} \\
\begin{verbatim}
{
  "42": {
    "Original_index": "42",
    "Chosen Disease": "MIGRAINE",
    "Target Symptom": "Headache",
    "Original Patient Description": "Patient reports mild 
     fatigue and nausea over the past day, but no other 
     significant complaints.",
    "Modified Patient Description": "Patient reports mild 
     fatigue and nausea over the past day, along with 
     a throbbing headache on one side of the head that 
     worsens with light exposure.",
    "ChangeFlag": true
  }
}
END JSON
\end{verbatim}

\textbf{Expected Reasoning Output:} \\
\begin{verbatim}
Reasoning for Item 42:
Original Index: 42
Target Symptom: Headache
Current Label: 0
Chosen Disease: MIGRAINE
Desired Direction: MORE aligned

Final Action:
A unilateral, throbbing headache with light sensitivity 
was added to the patient_description, which increases 
alignment of the headache symptom with migraine-related 
patterns. No disease names or diagnostic language were 
introduced, and existing symptoms were preserved 
unchanged, fully respecting the label definitions.
\end{verbatim}
\hrule
\vspace{0.2cm}
\textbf{Dataset:}
\end{promptbox}

\vspace{0.5cm}
\twocolumn

\section{Artifact Licenses and Usage}
\label{sec:appendix_licenses}
In compliance with reproducibility and ethical guidelines regarding artifact licenses, we detail the licensing, terms of use, and intended usage of the models and datasets utilized in this study. All computational artifacts were employed strictly for academic research purposes, consistent with their respective licenses.

\paragraph{Models}
\begin{itemize}
    \item \textbf{Gemini 2 Flash}~\cite{geminiteam2025geminifamilyhighlycapable}: Accessed via the official API. Usage complies with Google's Terms of Service and generative AI usage guidelines for research and evaluation purposes.
    \item \textbf{Qwen3-14B} ~\cite{yang2025qwen3technicalreport}: The model weights and inference code are utilized under the Apache 2.0 License, which explicitly permits academic research applications.
    \item \textbf{gpt-OSS-20b} ~\cite{openai2025gptoss120bgptoss20bmodel}: Utilized under its respective open-source license Apache 2.0, adhering to all research usage constraints.
\end{itemize}

\paragraph{Datasets}
\begin{itemize}
    \item \textbf{LIBERTY (Disease Diagnosis)} \citep{toker2026libertycausalframeworkbenchmarking}: A synthetic medical corpus. The dataset is publicly released by the authors for research purposes; while no explicit license is designated in the original publication, it is utilized here strictly for non-commercial academic research.
    \item \textbf{IMDB (Sentiment Analysis)} \citep{maas-EtAl:2011:ACL-HLT2011}: The dataset is publicly available for research purposes and is utilized in accordance with its standard academic distribution terms.
    \item \textbf{LAJ (LLM-as-a-Judge)} \citep{CalderonER25}: Derived from Reddit data, the dataset is publicly released by the authors for research purposes. While no explicit open-source license is provided in the original publication, our usage strictly complies with the Reddit API terms of service and is restricted entirely to non-commercial academic research.
\end{itemize}

\paragraph{Code and Framework}
The code for our concept extraction, MCMC-based counterfactual generation, and causal graph construction is open-sourced and released under the MIT License to facilitate unrestricted use and reproducibility by the research community.

\section{Computational Resources and Compute Cost}
\label{sec:appendix_compute}

\paragraph{Hardware Infrastructure}
Experiments evaluating the open-weight models (Qwen3-14B and gpt-OSS-20b) were conducted on an institutional remote compute cluster equipped with NVIDIA H200 and RTX6K GPUs. To optimize inference throughput during the intensive data expansion phase, these models were served utilizing the vLLM ~\cite{kwon2023efficientmemorymanagementlarge} framework. Gemini-2-Flash was accessed externally via its official API.

\paragraph{Cost Estimation}
The primary computational bottleneck in our proposed framework is the MCMC-inspired counterfactual data expansion, which necessitates repeated, iterative LLM sampling.
\begin{itemize}
    \item \textbf{API Costs (Gemini):} Executing the complete concept extraction and MCMC expansion pipeline via the Gemini API incurred a total cost of approximately \$540 USD.
    \item \textbf{Local GPU Compute (Qwen3 \& gpt-OSS):} Generating the counterfactuals for the open-weight models on our institutional cluster incurred negligible marginal compute costs (under \$5 USD in equivalent compute time). 
\end{itemize}
Importantly, once the counterfactual datasets are generated, the downstream causal discovery phase training the multinomial logistic regression models to extract the graph topologies is computationally lightweight and can be executed on standard CPU infrastructure within minutes.

\section{AI Assistance Statement}
In accordance with the ACL guidelines regarding the use of AI, we acknowledge the use of AI assistants during the preparation of this work. Specifically, GitHub Copilot was utilized to assist in writing data visualization scripts and refactoring portions of the codebase for asynchronous execution. Additionally, AI language models (Gemini, ChatGPT, and Claude) were used as writing assistants to refine phrasing and improve readability. The authors take full responsibility for all content, ideas, and code presented in this paper.

\end{document}